\documentclass[sigconf, nonacm]{acmart}

\usepackage{color}

\usepackage{xspace}
\usepackage{makecell}
\usepackage{subfigure}
\usepackage{multirow}
\usepackage{threeparttable}
\usepackage{enumitem,kantlipsum}
\usepackage[ruled,vlined]{algorithm2e}

\newcommand{\framework}{Picket\xspace}
\newcommand{\ModelName}{Picket\xspace}
\newcommand{\NetworkName}{PicketNet\xspace}
\newcommand{\newparagraph}[1]{\vspace{2pt}\noindent\textbf{#1\xspace~~}}
\newcommand{\blue}[1]{\textcolor{black}{#1}}

\begin{document}
% \title{Attention-based Mixed-type Data Validation for Machine learning Systems}
\title{\framework: Guarding Against Corrupted Data in Tabular Data during Learning and Inference}

%%
%% The "author" command and its associated commands are used to define the authors and their affiliations.
\author{Zifan Liu}
\email{zliu676@wisc.edu}
\affiliation{%
  \institution{UW-Madison}
%   \streetaddress{1210 W. Dayton Street}
%   \city{Madison}
%   \state{Wisconsin}
%   \postcode{53706-1613}
}

\author{Zhechun Zhou}
\email{zhechunz@Andrew.cmu.edu}
\affiliation{%
  \institution{Carnegie Mellon University in Australia}
%   \streetaddress{City Road, Camperdown/Darlington NSW 2006}
%   \city{Sydney}
%   \country{Australia}
%   \postcode{53706-1613}
}

\author{Theodoros Rekatsinas}
\email{thodrek@cs.wisc.edu}
\affiliation{%
  \institution{UW-Madison}
%   \streetaddress{1210 W. Dayton Street}
%   \city{Madison}
%   \state{Wisconsin}
%   \postcode{53706-1613}
}

\begin{abstract}
Data corruption is an impediment to modern machine learning deployments. Corrupted data can severely bias the learned model and can also lead to invalid inferences.
We present, \ModelName, a simple framework to safeguard against data corruptions during both training and deployment of machine learning models over tabular data. 
For the training stage, \ModelName identifies and removes corrupted data points from the training data to avoid obtaining a biased model. For the deployment stage, \ModelName flags, in an online manner, corrupted query points to a trained machine learning model that due to noise will result in incorrect predictions.
To detect corrupted data, \ModelName uses a self-supervised deep learning model for mixed-type tabular data, which we call \NetworkName. To minimize the burden of deployment, learning a \NetworkName model does not require any human-labeled data.
\ModelName is designed as a plugin that can increase the robustness of any machine learning pipeline.
We evaluate \ModelName on a diverse array of real-world data considering different corruption models that include systematic and adversarial noise during both training and testing. 
We show that \ModelName consistently safeguards against corrupted data during both training and deployment of various models ranging from SVMs to neural networks, beating a diverse array of competing methods that span from data quality validation models to robust outlier-detection models.
\end{abstract}

\maketitle

\section{Introduction}\label{sec:intro}
% motivate our study
Data quality assessment is critical in all phases of the machine learning (ML) life cycle. 
Both in the training and deployment (inference) stages of ML models, erroneous data can have devastating effects.
In the training stage, errors in the data can lead to biased ML models~\cite{strongerPoison,Deequ,breck2019data,TFX}, i.e., models that learn wrong decision boundaries. 
In the deployment stage, errors in the inference queries can result in wrong predictions, which in turn can be harmful for critical decision making systems~\cite{breck2019data,certifiedDefense}. 
ML pipelines need reliable data quality assessment during both training and inference to be robust to data errors.

% introduce the problem
We focus on tabular data and seek to develop a simple,  plug-and-play approach to guard against corrupted data (including adversarially corrupted data) during both training and inference in ML pipelines. 
During training, our goal is to identify and filter corrupted examples from the data used to train a model, while during deployment, our goal is to flag erroneous query points to a pre-trained ML model, i.e., points that due to noise will result in incorrect predictions of the ML model. 
This work introduces a unified solution to guard against corrupted data for both the training and deployment stages of ML models.

% Challenges of data verification
Guarding against corrupted data in ML pipelines exhibits many challenges. 
First, detecting corrupted examples in the training data can be a hard exercise that requires developing methods that go beyond standard outlier detection mechanisms~\cite{supervised1}. Data poisoning techniques~\cite{certifiedDefense,strongerPoison,gradientPoison,SVMpoisoning} attack models by adding a small fraction of adversarially crafted poisoned data to the training set. Any reliable mechanism that filters corruptions from a training data set should not only remove easy to detect outliers but also hard to detect poisoned data. 

Second, online-detection of inference queries that yield a model misprediction due to corruption requires not only knowledge of the data quality, but also knowledge of the tolerance of the trained ML model to corruptions. The reason is that not all corruptions will flip the prediction of a trained ML model and different models exhibit different degrees of robustness to corruption. Moreover, adversarial noise may target specific subsets of the data or classes in the ML pipeline~\cite{strongerPoison}. For this reason, online-filtering of corrupted inference queries requires a method that takes both the downstream model and data quality into account.

The above challenges require rethinking the current solutions for identifying errors in data. The majority of outlier detection methods in the statistical literature~\cite{supervised1,isolation,OCSVM} and error detection methods in the database literature~\cite{holodetect,10.1145/3299869.3324956} are not effective against adversarial corruptions~\cite{strongerPoison}. More advanced methods are required to defend against adversarial corruptions~\cite{certifiedDefense}. However, current methods are typically limited to real-valued data~\cite{robustMean} and focus either on training~\cite{sever} or inference~\cite{odds,MWOC} but not both. Finally, recent techniques for data validation in ML pipelines that are deployed in industrial settings~\cite{Deequ,breck2019data} rely on user-specified rule- or schema-based quality assertions evaluated over batches of data and it is unclear if they can support on-the-fly, single point validation, which is required during inference.

We present \emph{\ModelName}, a framework for safeguarding against corrupted data during both the training and deployment stages of ML pipelines. 
\ModelName can be used in an offline manner to validate data that will be used for training but can also be used in an online manner to safeguard against corruptions for on-the-fly queries at inference time.
We empirically demonstrate that \ModelName outperforms both state-of-the-art outlier detection mechanisms such as Robust Variational Autoencoders~\cite{RobustVAE}, and state-of-the-art methods for detection of adversarial corruption attacks during inference~\cite{odds,MWOC}. Our work makes the following technical contributions:

\newparagraph{Self-Attention for Tabular Data} 
\ModelName is built around \NetworkName, a new deep learning-based encoder for mixed-type tabular data. \NetworkName can model mixtures over numerical, categorical, and even text-based entries of limited length (e.g., descriptions). The goal of \NetworkName is to learn the characteristics of the distribution governing the non-corrupted data on which the ML pipeline operates and it is used in \ModelName to distinguish between clean data points and corrupted ones.
The architecture of \NetworkName builds upon the general family of Transformer networks~\cite{attention} and introduces a new multi-head self-attention module~\cite{attention} over tabular data. This module follows a stream-based architecture that is able to capture not only the dependencies between attributes at the schema-level but also the statistical relations between cell values---it follows a schema stream and value stream architecture.
We find that compared to schema-only models, \NetworkName's two-stream architecture is critical for obtaining accurate predictions across diverse data sets.

\newparagraph{Robust Training over Arbitrary Corruptions} 
We show how to learn a \NetworkName model without imposing any extra labeling burden to the user and by operating directly on potentially corrupted data (i.e., we do not not require access to clean data to learn the non-corrupted data distribution). We achieve that by using a robust self-supervised training approach that is robust to corrupted data points (including adversarial points). As with standard self-supervision, the context captured in the data is used as the supervision signal. 
The training procedure for \NetworkName monitors the reconstruction loss of tuples in the input data over early training iterations and uses related statistics to identify suspicious data points.
These points are then excluded from subsequent iterations during training.

\newparagraph{A Plugin to ML Pipelines} We demonstrate how \ModelName can serve as a ``plugin'' that safeguards against corrupted data in different ML pipelines during both training and inference. We evaluate \ModelName over multiple data sets with different distributional characteristics and consider different types and magnitudes of corruption, ranging from simple random noise to adversarial attacks that explicitly aim to harm the performance of downstream ML models. We find that \ModelName provides a reliable mechanism for detecting data corruptions in ML pipelines: \ModelName consistently achieves an area under the receiver operating characteristic curve (AUROC) score of above or close to 80 points for detecting corrupted data across different types of noise and ML models.

\section{Background}\label{sec:background}
\newparagraph{Data Corruption Models}
We consider data corruption due to random, systematic, and adversarial noise. 

\vspace{1pt}\noindent 1. \emph{Random noise} is drawn from some unknown distribution that does not depend on the data. Random noise is not predictable and cannot be replicated in a repeatable manner. While, many ML models are robust to purely random noise during training, high-magnitude random noise can still lead to false predictions, and hence is of interest to our study.

\vspace{1pt}\noindent 2. \emph{Systematic noise} depends on values in the data and leads to repeated errors in data samples. This type of noise biases the distribution of the data. Systematic noise can skew the distribution of the data, and this bias can potentially harm the performance of an ML model depending on the importance of the corrupted features to the downstream prediction task.

\vspace{1pt}\noindent 3. \emph{Adversarial noise} contaminates the data to explicitly mislead ML models and harm their performance. At training time, adversarial noise corrupts the training points to force a model to learn a bad decision boundary; at test time, adversarial noise corrupts the input queries in a manner that will lead to a false prediction by the model. It usually depends on the data and the target model, although some types of adversarial noise may work well across different models.

\newparagraph{Dealing with Corrupted Data in ML}
The most common approach to deal with corrupted data during training is to identify corrupted samples and remove them from the training set.
This process is referred to as \emph{filtering}.
Given a training data set $D$, filtering identifies a set of clean data points $C \subseteq D$ to be used for training.
Common filtering mechanisms rely on outlier detection methods~\cite{isolation,OCSVM,RobustVAE}. In addition, recent filtering methods focus on adversarial corruptions over real-valued data~\cite{RobustVAE,robustMean}. Finally, there are data validation modules for ML platforms~\cite{TFX,Deequ,breck2019data} that rely on user-defined rules and simple statistics to check the quality of data batches. The statistical tests used by these methods are subsumed by outlier detection methods and user-defined quality rules are out of the scope of this work. For inference, apart from outlier detection methods, there are methods that accept or reject inference queries by using statistical tests that compare the query to clean data~\cite{MWOC} or by considering variations in a model's internal data representation~\cite{odds}. We also consider the online detection of inference queries that result in wrong predictions due to corruption.

%Modern ML platforms ~\cite{TFX, Deequ} incorporate data validation modules that rely on rule-based integrity constraints and simple one-column statistics to perform outlier detection. Learning-based outlier detection methods ~\cite{isolation, OCSVM, RobustVAE} leverage ML models to learn the distribution of clean data and detect out-of-distribution samples. There are methods in the adversarial learning literature designed specifically for inference time adversarial data detection ~\cite{odds, MWOC} based on the assumption that adversarial data are statistically different from benign data. 

\newparagraph{Self-Supervision} In self-supervised learning systems~\cite{bert,VL-BERT}, the learning objective is to predict part of the input from the rest of it. A typical approach to self-supervision is to mask a portion of the input, and then let the model reconstruct the masked portion based on the unmasked parts. By self-supervision, a model learns to capture dependencies between different parts of the data. Self-supervised learning is a subset of unsupervised learning in a broad sense since it does not need human supervision.

\newparagraph{Multi-Head Self-Attention} Models with multi-head self-attention mechanism learn representations for structured inputs e.g., a tuple or a text sequence, by capturing the dependencies between different parts of the inputs~\cite{attention}. One part can pay different levels of attention to other parts of the same structured input. For example, consider the text sequence ``the dog wears a white hat'', the token ``wears'' pays more attention to ``hat'' than ``white'' although ``white'' is closer in the sequence. The attention mechanism can also be applied to tuples that consist of different attributes~\cite{AimNet}. Multi-head self-attention takes an ensemble of different attention functions, with each head learning one. 
%In this work, we consider a tuple in the tabular data as the structured input.

We provide a brief review of the multi-head self-attention model~\cite{attention}. Let $x_1, x_2, \dots, x_T$ be the embedding of a structured input with $T$ tokens. Each token $x_i$ is transformed into a query-key-value triplet ($q_i=W_Q x_i$,  $k_i=W_K x_i$, $v_i=W_V x_i$) by three learnable matrices $W_Q$, $W_K$ and $W_V$. The query $q_i$, key $k_i$, and value $v_i$ are real-valued vectors with the same dimension $d$. The output of a single head for the $i^{\text{th}}$ token is $\sum_{j=1}^{T} w_{ij} v_{j}$, a weighted sum of all the values in the sequence, where $w_{ij} = \text{softmax}((q_i^T k_1, q_i^T k_2, \dots, q_i^T k_T) / \sqrt{d})_j$. The attention $x_i$ pays to $x_j$ is determined by the inner product between $q_i$ and $k_j$. Multiple heads share the same mechanism but have different transformation matrices. The outputs of all the heads are concatenated and transformed into the final output by an output matrix $W_O$, which is also learnable.

\section{Overview of \ModelName}\label{sec:framework}
\begin{figure*}[th]
\centering
\includegraphics[width=0.9\textwidth]{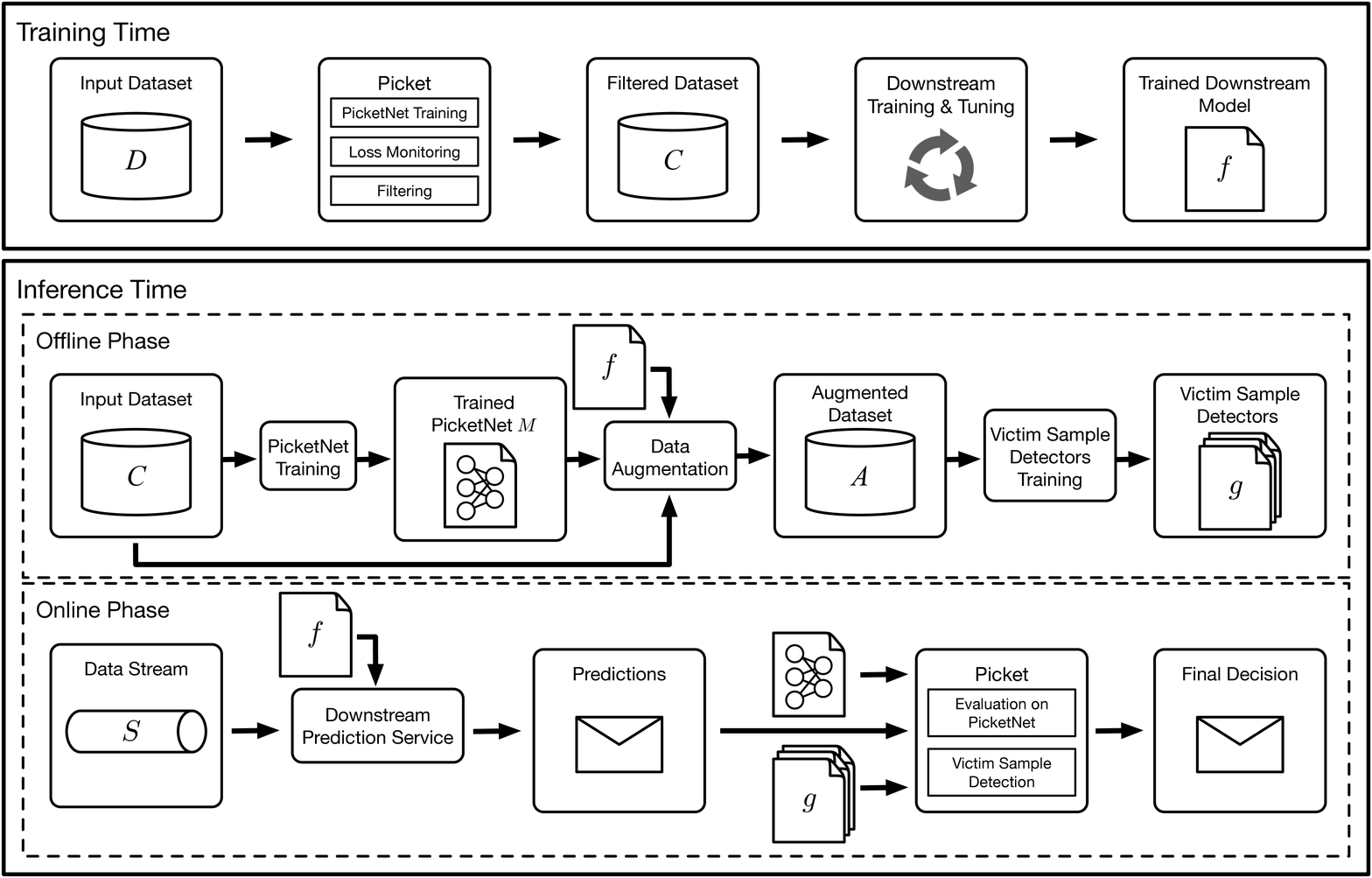}
\caption{The key components of a typical machine learning pipeline with \ModelName.}
\label{fig:pipeline}
\end{figure*}

We review \ModelName's functionalities during the training and inference phases of a ML pipeline. An overview diagram of \ModelName's core components and functionalities is shown in Figure~\ref{fig:pipeline}. \blue{The corresponding pseudo-code is shown in Algorithm~\ref{alg:overview}.} 

\begin{algorithm}
\SetAlgoLined
 \textbf{Training Time:}\\
 \textbf{Input:} dataset $D$, downstream model type and configuration $\mathcal{I}_{\text{config}}$\;
 \textbf{Output:} filtered dataset $C$, trained downstream model $f$\;
 $C = \text{PicketNetTrainingAndEarlyFiltering}(D, )$\;
 $f = \text{DownstreamModelTraining}(C, \mathcal{I}_{\text{config}})$\;
 \hrulefill\\
 \textbf{Inference Time (Offline Phase):}\\
 \textbf{Input:} filtered dataset $C$, trained downstream model $f$\;
 \textbf{Output:} trained \NetworkName $M$, victim sample detectors $g$\;
 $M = \text{PicketNetTraining}(C)$\;
 augmented dataset $A = \text{DataAugmentation}(M, f)$\;
 $g = \text{VictimSampleDetectorTraining}(A)$\;
 \hrulefill\\
 \textbf{Inference Time (Online Phase):}\\
 \textbf{Input:} data stream $S$, trained downstream model $f$, trained \NetworkName $M$, victim sample detectors $g$\;
 \textbf{Output:} final prediction $y_\text{prediction}$\;
 raw prediction $y_\text{raw} = \text{DownstreamPrediction}(S, f)$\;
 $y_\text{prediction} = \text{PicketVictimDetection}(S, y_\text{raw}, M, g)$
 \caption{\blue{Picket in a typical ML pipeline}}
 \label{alg:overview}
\end{algorithm}

\newparagraph{Guarding against Corrupted Data in Training}\\
We consider a tabular data set $D$ with $N$ training examples. Let $x$ be a sample (tuple) in $D$ with $T$ attributes. \blue{These attributes correspond to the features that are used by the downstream model.} %and can also contain the correct label for the data sample (optionally). 
For each $x$ we denote $x^*$ its clean version; if $x$ is not corrupted then $x = x^*$.

We assume that $D$ contains clean and corrupted samples and that the fraction of corrupted samples is always less than half. The goal of \ModelName is to filter out the corrupted samples in $D$ and construct a \emph{clean} set of examples $C \subseteq D$ to be used for training a downstream model. Without loss of generality we assume that \ModelName performs filtering over $D$ once. This process can be repeated for data batches over time. We next describe how we construct $C$ in \ModelName.

\ModelName follows the next steps: First, \ModelName learns a self-supervised \NetworkName model $M$ that captures how data features are distributed for the clean samples. \ModelName does not require human-labelled examples of corrupted or clean data.
During training, \ModelName records the reconstruction loss across training epochs for all points in $D$.
After training of $M$, we analyze the reconstruction loss progression over the first few training epochs to identify points in $D$ that are corrupted (see Section~\ref{sec:robust_ml} for details). 
Set $C$ is constructed by removing these corrupted points from $D$. We also proceed with training $M$ on $C$. The pre-trained \NetworkName model $M$ is used to detect corruptions during inference.

\begin{figure*}[t]
    \centering
    \subfigure[Before Corruption]{\includegraphics[width=0.4\textwidth]{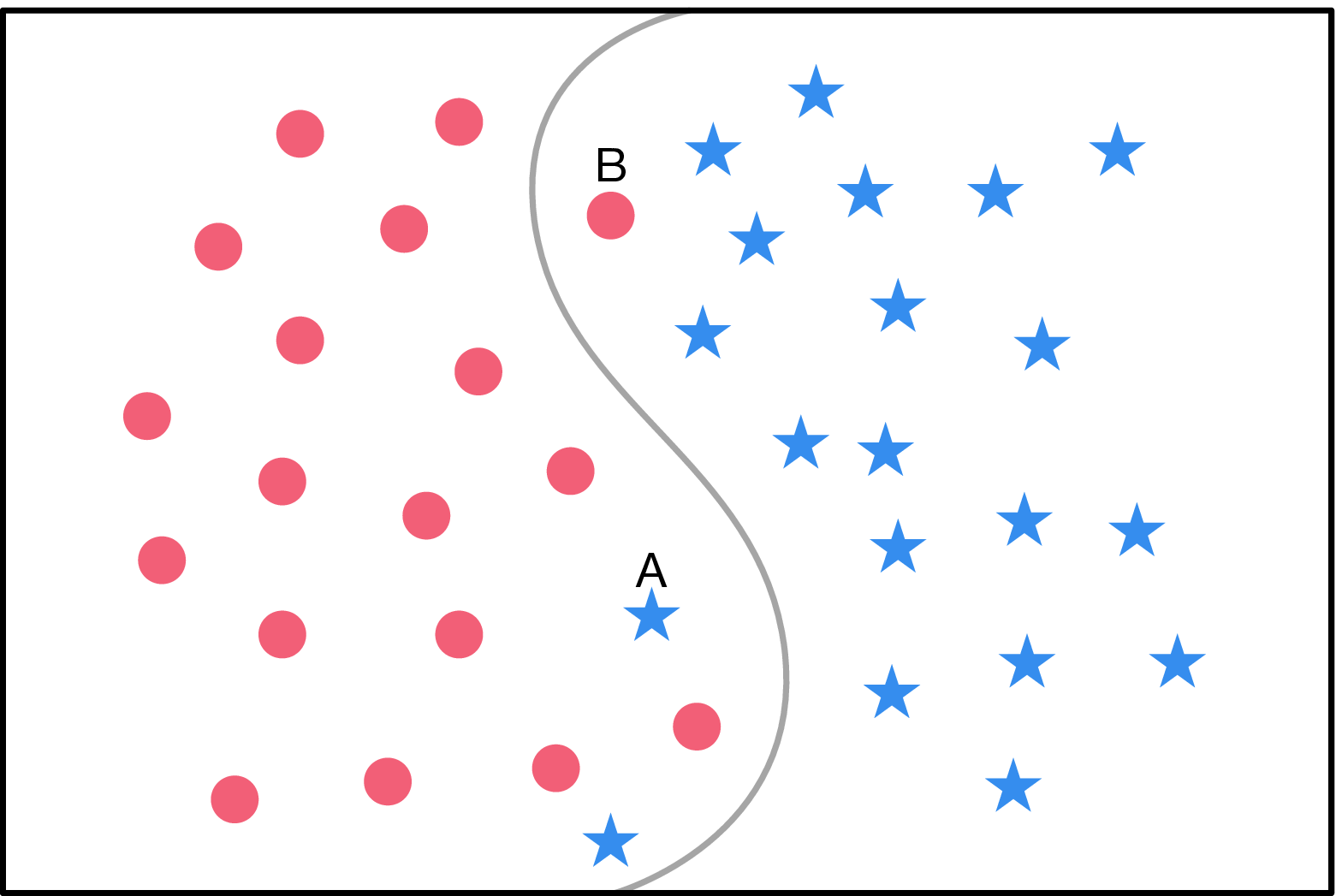}}
    \subfigure[After Corruption]{\includegraphics[width=0.4\textwidth]{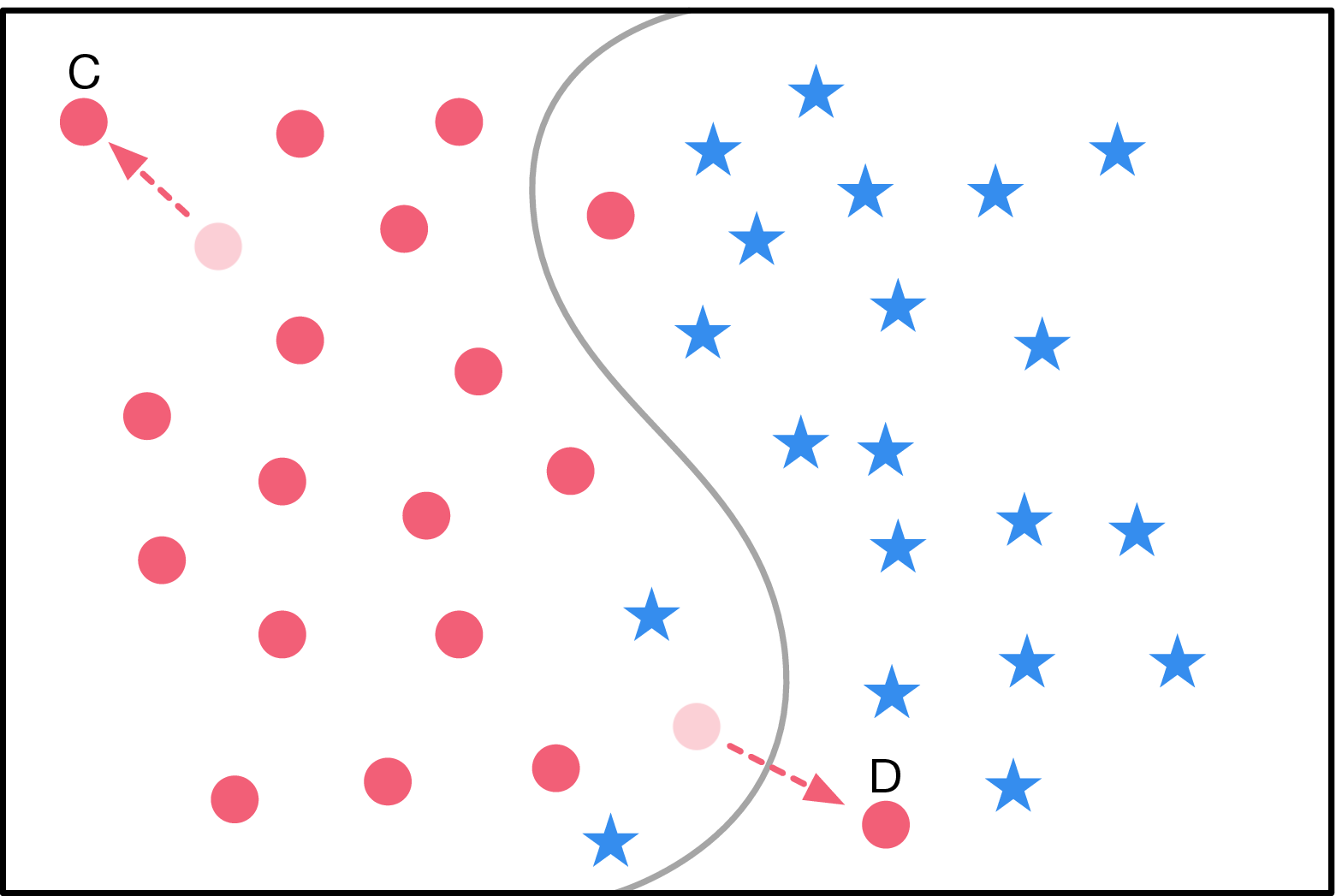}} 
    \caption{An example of non-victim and victim samples. The grey line is the decision boundary of a binary classifier that separates the red circles and the blue stars in the two-dimensional space. (a)Before corruption, some samples (e.g. point A and B) get misclassified, but they are not victim samples because they are clean, and the misclassification is due to the limitation of the model. Those samples should be handled by model analytics, and are out of the scope of our framework. (b) After corruption, two samples are shifted by the noise (point C and D). C is not a victim sample since the noise injected does not affect the correctness of classification. D is a victim sample because it gets misclassified due to noise.}
    \label{fig:victimExample}
    \vspace{-10pt}
\end{figure*}

\newparagraph{Guarding against Corrupted Data in Inference}\\
We consider a trained model $f$ that serves inference queries over data points with the same $T$ attributes as in the training phase of the ML pipeline.
We define a \emph{victim sample} to be a point $x$ such that $f(x^*) = y$ but $f(x) \neq y$, i.e., the input sample is corrupted and it gets misclassified due to corruption. We show an example that illustrates the difference between non-victim and victim samples according to our definition in Figure~\ref{fig:victimExample}.
The goal of \ModelName is to solve the following problem: Given an already-trained classifier $f$, for each sample $x$ that comes on the fly, we want to tell if it is a good sample or it is a victim sample and will be misclassified due to corruption, i.e., we want to detect if $f(x) \neq f(x^*)$. We assume access to data set $C$ and model $M$ output by \ModelName for safeguarding during the training phase of the ML pipeline in hand. We then adopt a two-phase approach, \emph{offline} and \emph{online} phase, to solve the aforementioned problem. 

We now focus on the offline phase. Given the trained model $f$, data set $C$, and model $M$, we learn a victim-sample detector for each class in the prediction task in-hand. Each victim-sample detector is a binary classifier that detects if an input sample $x$ will be misclassified by $f$ due to corruption. The victim-sample detectors operate on an extended feature set: Beyond the original $T$ features of the inference query $x$ we add $T$ additional features corresponding to the reconstruction-loss obtained by masking each feature in turn and applying model $M$ to predict it back.

During the online phase, we use model $M$ and the victim-sample detectors over a stream of incoming inference queries to identify victim samples. \ModelName performs the following: for each incoming point $x$, \ModelName evaluates classifier $f$ on $x$ to obtain an initial prediction $f(x)$. \ModelName also uses $M$ to compute the reconstruction-loss vector for the features of $x$. The extended feature vector containing the original features of $x$ and the reconstruction loss features are given as input to the victim sample detector for the class that corresponds to the prediction $f(x)$. Using this input, the detector identifies if point $x$ corresponds to a victim sample. If the point is not marked as suspicious the final prediction is revealed downstream, otherwise the inference query is flagged.

\section{The \NetworkName Model}\label{sec:picketnet}
% \begin{itemize}
% \color{blue}
%     \item A figure showing the two-stream architecture.
%     \item Describe the two stream attention, and how the model learns the structure both on the schema and the value level.
%     \item The advantage of the two-level structure learning.
% \end{itemize}

\ModelName uses a new two-stream multi-head self-attention model to learn the distribution of tabular data. We refer to this model as \NetworkName.
The term \emph{stream} refers to a path in a neural network that focuses on a specific view of the input data. 
For example, standard attention mechanism is one stream that learns value-based dependencies between the parts of the input data (see Section~\ref{sec:background}).
Combining multiple streams, where each stream focuses on learning a different view of the data, has been shown to achieve state-of-the-art results in natural language processing tasks~\cite{xlnet} and computer vision tasks~\cite{simonyan2014two} but has not been applied on tabular data.
\NetworkName introduces a new two-stream model for tabular data and proposes a robust, self-supervised training procedure for learning this model.

\subsection{Model Architecture}\label{sec:architecture}
\begin{figure*}[t]
 \center
 \includegraphics[width=0.75\textwidth]{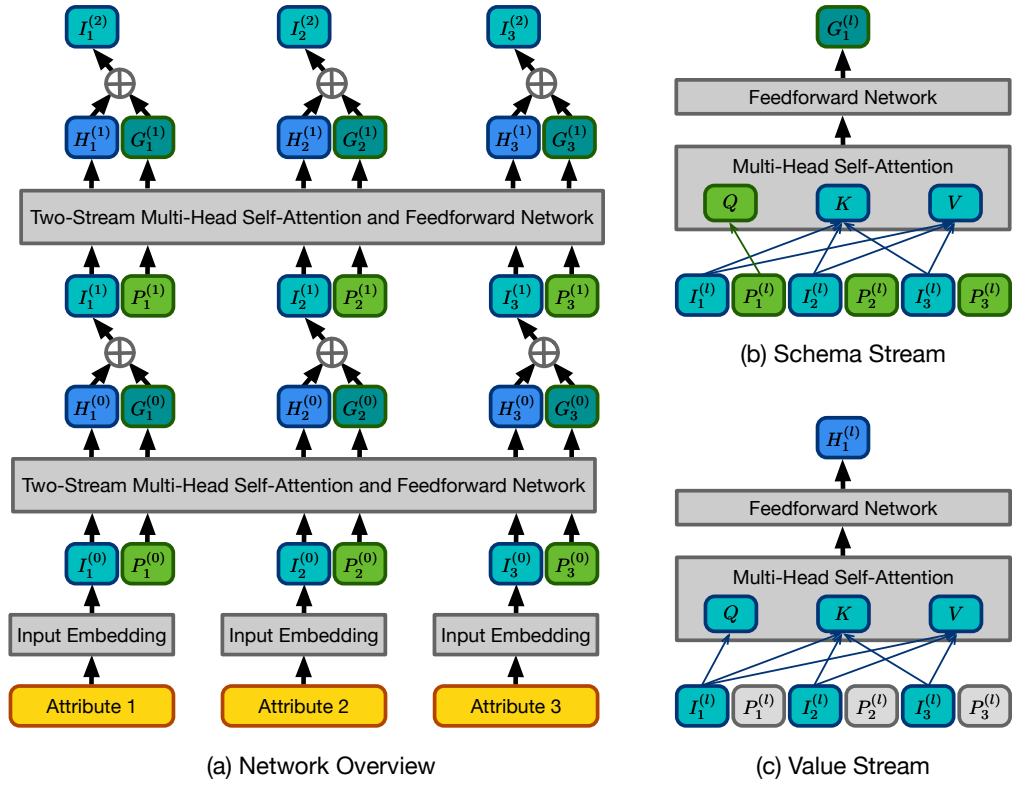}
 \caption{(a) Overview of the two-stream multi-head self-attention network. (b) An illustration of the schema stream for the first attribute.(c) An illustration of the value stream for the first attribute.}
 \label{fig:network}
\end{figure*}

\NetworkName contains two streams: a \emph{schema stream} and a \emph{value stream}.
The schema stream captures schema-level dependencies between attributes of the data, while the value stream captures dependencies between specific data values.
A design overview of \NetworkName is shown in Figure~\ref{fig:network} with details of the two streams.
The input to the network is a mixed-type data tuple $x$ with $T$ attributes $x_1, x_2, \dots, x_T$.

The first level of \ModelName obtains a numerical representation of tuple $x$.
To capture the schema- and value-level information for $x$, we consider two numerical representations for each attribute $i$: 1) a real-valued vector that encodes the information in value $x_i$, denoted by $I_i^{(0)}$, and 2) a real-valued vector that encodes schema-level information of attribute $i$, denoted by $P_i^{(0)}$. For example, a tuple with two attributes is represented as $I_1^{(0)}P_1^{(0)}I_2^{(0)}P_2^{(0)}$. To convert $x_i$ to $I_i^{(0)}$, \NetworkName uses the following process:
The encoding for each attribute value $x_i$ is computed independently. 
We consider 1) categorical, 2) numerical, and 3) textual (short-text) attributes.
For categorical attributes, we use a learnable lookup table to get the embedding for each value in the domain. This lookup table is learned jointly with all other components of \NetworkName. 
\blue{For numerical attributes, we keep the raw value as one dimension and pad the other dimensions with zeros.} 
For text attributes, we train a fastText~\cite{fasttext} model over the corpus of all the texts and apply SIF ~\cite{sif} to aggregate the embedding of all the words in a cell.
The initial embedding vectors $I_i^{(0)}$ are inputs to the value-level stream.

Each vector $P_i^{(0)}$ serves as a \emph{positional encoding} of the attribute associated with index $i$. Positional encodings are used to capture high-level dependencies between attributes. $P_i^{(0)}$ is consistent for attribute $i$ in all examples, i.e., it does not change as the values in different examples vary. Hence, it captures common dependencies at the schema level.
%For example, if there exists a strong dependency between two attributes $i$ and $j$, vectors $P_i^{(0)}$ and $P_j^{(0)}$ should be such that the attention score between $i$ and $j$ is high. 
Each $P_i^{(0)}$ corresponds to a trainable vector that is initialized randomly and is fed as input to the schema stream.

We now describe subsequent layers of our model. These layers consider the two attention streams and form a stack of $n$ self-attention layers. The output of the previous layer serves as the input to the next layer. Self-attention layer $l$ takes the value vector $I_i^{(l)}$ and positional encoding $P_i^{(l)}$ to learn a further representation for attribute $i$ and its value $x_i$. 
After each attention layer, the outputs of the two streams are aggregated and fed as input to the value-level stream of the next layer, while the schema stream still takes as input the positional encoding.
The output of the value stream $H_i^{(l)}$ and that of the schema stream  $G_i^{(l)}$ are computed as: 
\begin{equation*}
\begin{split}
    H_i^{(l)} = \textbf{MHS}(& Q=L_Q(I_i^{(l)}), K=L_K(I_{j=1,\dots,T}^{(l)}),\\ & V=L_V(I_{j=1,\dots,T}^{(l)}))
\end{split}
\end{equation*}
\begin{equation*}
\begin{split}
G_i^{(l)} = \textbf{MHS}(& Q=L_Q(P_i^{(l)}), K=L_K(P_{j=1,\dots,T}^{(l)}), \\ & V=L_V(I_{j=1,\dots,T}^{(l)}))
\end{split}
\end{equation*}
where $\textbf{MHS}$ represents the multi-head attention function followed by a feed-forward network and $L_Q$, $L_K$, $L_V$ are linear transformations that transform the input into query, key, or value vectors by the corresponding weight matrices for $Q$, $K$, and $V$. Finally, $Q,K,V$ are matrices formed by packing the query, key and value vectors from their inputs. 

The difference between the two streams is that the query in the schema stream corresponds to the positional encoding, therefore it learns higher-level dependencies. For the value stream the input to the next level is the sum of the outputs from the two streams: $I_{i}^{(l+1)} = H_i^{(l)}+G_i^{(l)}$; for the schema stream the input to the next level $P_{i}^{(l+1)}$ corresponds to a new positional encoding that does not depend on the previous layers.
If layer $l$ is the last layer, $O_{i} = I_{i}^{(l+1)}$ is the final representation for attribute value $x_i$. 

% \begin{figure}[t]
%  \center
%  \includegraphics[width=0.45\textwidth]{figures/Architecture1.eps}
%  \caption{\NetworkName's two-stream self-attention network.}
%  \label{fig:network1}
% \end{figure}

% \begin{figure}[t]
%  \center
%  \includegraphics[width=0.45\textwidth]{figures/Architecture2.eps}
%  \caption{(a) An illustration of the schema stream for the first attribute. (b) An illustration of the value stream for the first attribute.}
%  \label{fig:network2}
% \end{figure}

%The advantage of our two-stream attention mechanism over one-stream attention models is that it takes both schema-level and value-level dependencies into consideration. For example, consider two attributes \textit{University Name} and \textit{State}. The value stream can learn a mapping from \textit{University Name} to \textit{State}, while the schema stream can learn that \textit{University Name} and \textit{State} are correlated if the name of the university contains the state. The combination of the two streams provides additional robustness to the model.

\subsection{Training Process}\label{sec:training_picketnet}
We learn \NetworkName using the noisy data set $D$ without any human-labeled examples of corrupted or clean data. Training follows a self-supervised learning objective.

\newparagraph{Self-Supervised Training}
For each point in $D$, we mask one of the attributes and then try to reconstruct it based on the values of the other attributes in the same tuple. Other attributes may still contain noisy data or missing values. 
The attributes are masked in turn following an arbitrary order. The training is also multi-task since the reconstruction of each attribute forms one learning task. 

We use different types of losses for the three types of attributes to quantify the quality of reconstruction. Consider a sample $x$ whose original value of attribute $i$ is $x_i$. If $x_i$ is numerical, its a one-dimensional value, and hence, the reconstruction of the input value is a \emph{regression task}: We apply a simple neural network on the output $O_i$ to get an one-dimensional reconstruction $\hat{x}_i$, and use the mean squared error (MSE) loss: $\text{MSE}(x_i, \hat{x}_i) = (x_i - \hat{x}_i)^2$.

For categorical or text-based attributes we use the cross-entropy loss. Consider a tuple $x$ and its attribute $i$. 
For its attribute value $x_i$ let $I_i^{0}(x_i)$ be the base-embedding before passing through the attention layers of \NetworkName, and $O_i(x_{\text{mask}})$ the contextual encoding of value $x_i$ after pushing tuple $x_{\text{mask}}$ (with attribute $i$ masked) through \NetworkName.
Given tuple $x$, we randomly select a set of other values $Z_i$ from the domain of attribute $i$. We consider the training loss associated with identifying $x_i$ as the correct completion value from the set of possible values $\{ x_i\} \cup Z_i$.
To compute the training loss we use the cosine similarity between $O_i(x_{\text{mask}})$ and the input encoding $I_i^{0}(r)$ for each $r \in \{ x_i\} \cup Z_i$, then we apply the softmax function over the similarities and calculate the cross-entropy (CE) loss: 
%If attribute $i$ is categorical or textual, we consider the input embedding $I^i(0)$ of $x_i$ is a vector. We randomly select a set of other values $Z_i$ from the same domain, and the task is to pick $x_i$ from $\{ x_i\} \cup Z_i$. We compute the cosine similarity between $O_i$ and each of the values in $\{ x_i\} \cup Z_i$, Then we apply the softmax function over the similarities and calculate the cross-entropy (CE) loss:
\begin{equation*}
\small
\begin{split}
    &\text{CE}(x, Z_i; i, M) \\ =  &-\log (\frac{\exp(\text{sim}(I_i^{(0)}(x_i), O_i(x_{\text{mask}})))}{\sum_{r \in \{ x_i\} \cup Z_i} \exp(\text{sim}(I_i^{(0)}(r), O_i(x_{\text{mask}})))})
\end{split}
\end{equation*}
where $\text{sim}(a, b)$ is the cosine similarity between $a$ and $b$.

\newparagraph{{Loss-based Filtering to Ensure Robust Training}}\\
The data used to learn \NetworkName can be corrupted, in which case self-supervised learning might lead to a biased model due to the presence of noise. 
To make learning robust to noisy input, we use a loss-based filtering mechanism to detect and ignore corrupted data during training of a \NetworkName model. The process we use follows the next steps:

\vspace{1pt}\noindent 1. Warm-start \NetworkName by training over $D$ for $E_1$ epochs.

\vspace{1pt}\noindent 2. Train \NetworkName over $D$ for $E_2$ epochs and, for each sample in $x \in D$, record the \emph{epoch-wise average loss} $\text{Loss}_i(x)$ for each attribute $i$, $i=1,2,\dots, T$.

\vspace{1pt}\noindent 3. For each sample, aggregate the losses attribute-wise by $\text{Loss}(x) = \sum_{i=1}^{T} \text{Loss}_i(x) / \text{Median}_D(\text{Loss}_i(\cdot))$
%  \begin{equation*}
%  \small
%  \label{eq:loss}
%      \text{Loss}(x) = \sum_{i=1}^{T} \text{Loss}_i(x) / \text{Median}_D(\text{Loss}_i(\cdot))
%  \end{equation*}
 where $\text{Median}_D$ computes the median over all points in $D$.
 
 \vspace{1pt}\noindent 4. Put a sample into set $D'$ if its aggregated loss is less than $\delta_{\text{low}}$ or greater than $\delta_{\text{high}}$, where $\delta_{\text{low}}$ and $\delta_{\text{high}}$ are pre-specified thresholds; $D'$ is the set of samples to be removed.
 
 \vspace{1pt}\noindent 5. Train \NetworkName over $C = D \setminus D'$ until convergence.
 
The thresholds $\delta_{\text{low}}$ and $\delta_{\text{high}}$ control the sensitivity of the detection. In practice, we can set $\delta_{\text{low}}$ and $\delta_{\text{high}}$ based on a relatively clean validation set. A common strategy is setting their values based on the validation set so that the false positive rate (FPR) is under some value (e.g. 5\%). \blue{When a relatively clean validation set is not available, the thresholds can be set based on the histogram of the reconstruction loss. Filtering out abnormal peaks and low density tails in the histogram is a natural strategy, and we validate the effectiveness of it in Section~\ref{sec:microbenchmarks}.}

When we do the attribute-wise aggregation, we normalize the loss of each attribute by dividing with the median of it to bring different types of losses to the same scale. 
The normalized loss characterizes how large the loss is relative to the average level loss in that attribute. We use the median since it is robust against extremely high or low values, while the mean can be  significantly shifted by them.

 \begin{figure}
\centering
\includegraphics[width=0.45\textwidth]{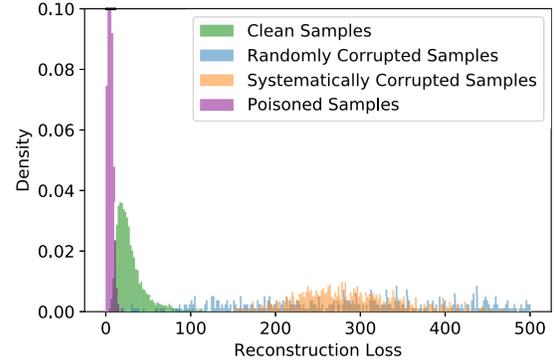}
\caption{Distribution of the reconstruction loss (early in training) for different types of clean and noisy samples.}
\label{fig:lossHist}
\end{figure}

The filtering is two-sided because randomly or systematically corrupted samples and adversarially crafted (poisoned) samples have different behaviors during the early training stage. Outliers with random or systematic noise are internally inconsistent and thus have high reconstruction loss in the early training stage of \NetworkName. However, poisoned samples tend to have unusually low reconstruction loss. The reason is that poisoned data tend to be concentrated on a few locations to be effective and appear normal, as is pointed out by Koh et al.~\cite{strongerPoison}. Such concentration forces deep networks such as \NetworkName to fit quickly and therefore the reconstruction loss in the early stage is lower than that of the clean samples. We confirm this hypothesis experimentally. Figure ~\ref{fig:lossHist} shows the distribution of the reconstruction loss for 1) clean, 2) randomly and systematically corrupted, and 3)poisoned samples for a real-world dataset. The noise used in this illustrative example follows the procedure described in Section~\ref{sec:exp_setup}. The three distributions have notable statistical distances. Hence, we need to remove samples with high loss to capture random or systematic corruptions, and samples with abnormally low loss to defend against poisoning attacks.

% \begin{algorithm}[ht]
% \label{alg:lossbased}
% \SetAlgoLined
% \KwInput{Training set $X$, filtering threshold $\delta_{\text{low}}$, $\delta_{\text{high}}$, number of warm-up epochs $E_1$, number of loss-recording epochs $E_2$}
%  Train \NetworkName over $X$ for $E_1$ epochs to warm up\;
%  Train \NetworkName over $X$ for $E_2$ epochs and record the average loss $\text{Loss}_i$ for attribute $i$ across epochs, $i=1,2,\dots, M$ \;
%  Compute the aggregated loss: $\text{Loss}_{tuple} = \sum_{i=1}^{M} \text{Loss}_i / \text{Median}(\text{Loss}_i)$ \;
%  Samples with aggregated loss less than $\delta_{\text{low}}$ or greater than $\delta_{\text{high}}$ form the set $X'$ to be filtered out \;
%  Train \NetworkName over $X \setminus X'$ until convergence \;
% \caption{Loss-based filtering in the early training stage of \NetworkName.}
% \end{algorithm}

\section{Detecting Data Corruptions}\label{sec:robust_ml}
The reconstruction loss of \NetworkName is the key to training time and inference time detection. We now provide more details on these functions of \ModelName.

\newparagraph{Detecting Corrupted Training Data}
Detection of corrupted training data follows directly from the training procedure of \NetworkName described in the previous section (Section~\ref{sec:training_picketnet}). Given an ML pipeline that aims to learn a model $f$ for a downstream task, we 1) first train a \NetworkName model over the data considered for training and 2) only use the data points that are not filtered during the training of \NetworkName to train the downstream model $f$. This approach allows us to apply \ModelName to any training pipeline regardless of the downstream model. Effectively, the pre-trained \NetworkName is used as an encoder capable to detect outlier points. As we show in Section~\ref{sec:exps}, our approach is effective across different types of ML models. For adversarially poisoned training data, we find that using \ModelName as a filter before training, allows us to train downstream ML models that exhibit similar performance to that of models trained on non-corrupted data.

\newparagraph{Victim Sample Detection for Inference}
We now describe how we construct the victim sample detectors to safeguard against corruptions during inference for a trained classifier $f$ (see Section~\ref{sec:framework}). For each class $y$ in the downstream classification task, we build a detector $g_y$ to identify victim samples, i.e., samples that $f$ will misclassify due to corruption of the feature values. The detectors are binary classifiers. In our experiments, we use logistic regression models with regularization parameter $1.0$ as detectors. 

At inference time, the victim sample detectors are deployed along with the downstream model $f$ and a pre-trained \NetworkName model $M$. Whenever a sample $x$ comes, the downstream model gives the prediction $f(x)$. \blue{The corresponding detector $g_{f(x)}$ takes into account $x$ and the feature-wise reconstruction loss (not aggregated) from $M$ and decides if $x$ should be marked as suspicious.}

We learn the victim-sample detectors by using a data set with artificially corrupted data points. We describe this process below; notice that no human-labeled data is required.  We start from the filtered data $C$ output by \ModelName during the training phase of the ML pipeline. We first apply the already-trained classifier $f$ on all points in $C$ and obtain a subset of points for which $f$ returns the correct prediction, i.e., $f(x) = y$. 
We denote this subset $C_{\text{cor}}$. 
Moreover, we partition $C_{\text{cor}}$ into sets $C_{\text{cor}}^y$, one for each class $y$ of the downstream prediction class.
For each partition, we use the points in $C_{\text{cor}}^y$ to construct artificial victim samples and artificial noisy points for which $f$ returns the correct prediction despite the injection of noise. We discuss the artificial noise we inject in detail in Section~\ref{sec:exp_setup}.
Let ${VS}^y$ and ${NS}_{\text{cor}}^y$ be the set of artificial victim samples and the set of noisy but correctly classified sample generated from $C_{\text{cor}}^y$ respectively.
To construct these two data sets we select a random point $x^*$ from $C_{\text{cor}}^y$ and inject artificial noise to obtain a noisy version $x$; we then evaluate $f(x)$ and if $f(x) = f(x^*) = y$ we assign the generated point $x$ to ${NS}_{\text{cor}}^y$ otherwise we assign it to ${VS}^y$.
We iteratively perform the above process for randomly selected points in $C_{\text{cor}}^y$ until we populate sets ${VS}^y$ and ${NS}_{\text{cor}}^y$ with enough points such that $|C_{\text{cor}}^y| = |{NS}_{\text{cor}}^y| =0.5 \times |{VS}^y|$.
Given these three sets, we construct a new augmented data set $A^y = C_{\text{cor}}^y \cup {NS}_{\text{cor}}^y \cup {VS}^y$.
We extend the feature vector for each point in $x \in A^y$ by concatenating it with the reconstruction loss vector obtained after passing each point through the trained \NetworkName $M$. We also assign to it a positive label (indicative that we will obtain a correct prediction) if it originated from $C_{\text{cor}}^y$ or ${NS}_{\text{cor}}^y$ and a negative label (indicating that we will obtain a wrong prediction) if it originated from ${VS}^y$. 
The output of this procedure is the training data for the victim sample detector $g_y$.
We repeat the above process for each class $y$.

Ideally, the artificial noise that we inject should have the same distribution as that in the real-world case. However, it is impossible to know the exact noise distribution in advance. A practical solution is injecting mixed-type artificial noise to help the detectors learn an approximate boundary between good and victim samples. As mentioned we discuss the artificial noise we consider in Section~\ref{sec:exp_setup}. We validate the effectiveness of mixed-type artificial noise in Section~\ref{sec:microbenchmarks}.

\section{Experiments}\label{sec:exps}
We evaluate how effective \ModelName and a diverse array of competing methods are on detecting different types of corruption in ML pipelines during the training and inference phases. We also provide several micro-benchmarks over different design choices in \ModelName. Finally, we report the runtime and discuss the scalability.

\subsection{Experimental Setup}\label{sec:exp_setup}
\newparagraph{Datasets} We consider six datasets with different mixtures of numerical, categorical, and text-based attributes. These datasets are obtained from the UCI repository~\cite{UCI} and the CleanML benchmark~\cite{cleanml}.
All datasets focus on binary classification tasks. 
The characteristics of these datasets are summarized in Table~\ref{tab:dataset}. A detailed description of the datasets is as follows. 

\begin{itemize}
    \item \textbf{Wine:} The dataset consists of statistics about different types of wine based on physicochemical tests. The task is to predict if the quality of a type of wine is beyond average or not. The features are purely numerical.
    \item \textbf{Adult:} The dataset contains a set of US Census records of adults. The task is to predict if a person makes over \$50,000 per year. The features are a mixture of categorical and numerical attributes. 
    \item \textbf{Marketing:} The dataset comes from a survey on household income consisting of several demographic features. The task is to predict  whether the annual gross income of a household is less than \$25,000. The features are purely categorical.
    \item \textbf{Restaurant:} The dataset contains information of restaurants from Yelp. The task is to predict if the price range of a restaurant is ``one dollar sign'' on Yelp. The features are a mixture of categorical values and textual description,
    \item \textbf{Titanic:} The dataset contains personal and ticket information of passengers. The task is to predict if a passenger survives or not. The features are a mixture of numerical, categorical and textual attributes.
    \item \textbf{HTRU2:} The dataset contains statistics about a set of pulsar candidates collected in a universe survey. The task is to predict if a candidate is a real pulsar or not. The features are purely numeric.
\end{itemize}

The last dataset, i.e., HTRU2, is purely numerical and we use it in the context of adversarial noise. \blue{The datasets above are the ones we use for most of our experiments. In addition, we use Food labeled by~\cite{holodetect} for real noise, and Alarm~\cite{alarm} for the study of scalibility.}
We consider downstream ML pipelines over these datasets that use 80\% of each dataset as the training set, and the rest as test data. To reduce the effect of class imbalance, we undersample the unbalanced datasets where over 70\% of the samples belong to one class. The numerical attributes are normalized to zero mean and unit variance before noise injection.

\begin{table}[t]
\caption{Properties of the datasets in our experiments.}
\scriptsize
\center
\label{tab:dataset}
\begin{tabular}{lcccc}
\toprule
Dataset & Size & \thead{\scriptsize{Numerical} \\ \scriptsize{Attributes}} & \thead{\scriptsize{Categorical} \\ \scriptsize{Attributes}} & \thead{\scriptsize{Textual} \\ \scriptsize{Attributes}}\\
\midrule
Wine    & 4898 & 11 & 0 & 0\\
Adult    & 32561 & 5 & 9 & 0\\
Marketing   & 8993 & 0 & 13 & 0\\
Restaurant    & 12007 & 0 & 3 & 7\\
Titanic    & 891 & 2 & 5 & 3\\
%Wearable Sensor    & 52081 & 7 & 0 & 0\\
HTRU2    & 17898 & 8 & 0 & 0\\
\bottomrule
\end{tabular}
\end{table}

\newparagraph{Noise Models} In our experiments, we consider different types of noise: 1) random, 2) systematic, 3) adversarial noise, and \blue{4) common errors in real-world datasets}. 

Random and systematic noise are model agnostic and only take into account the dataset. For random and systematic noise, we corrupt $\beta$ fraction of the cells in the noisy samples. We now provide a detailed description of the random and systematic noise generation process we consider. 

\vspace{2pt}\noindent\emph{Random Noise}: For a categorical or textual attribute, the value of a corrupted cell is flipped to another value in the domain of that attribute. For a numerical attribute, we add Gaussian noise to the value of a corrupted cell, with zero mean and standard deviation of $\sigma_1$, where $\sigma_1$ is a constant.
    
\vspace{2pt}\noindent\emph{Systematic Noise}: For categorical and textual data, we randomly generate a predefined function $\phi$ which maps the value $x_i^*$ of the cell to be corrupted to another value $x_i$ in the same domain. The mapping function depends on both the original value in that attribute and that in another pre-specified attribute, i.e., $x_i = \phi(x_i^*, x_j^*)$ where $j \neq i$. For a numerical attribute, we add a fixed amount of noise $\sigma_2$ to the value of a corrupted cell, where $\sigma_2$ is a constant.

We consider three settings with respect to the fraction of corrupted cells in the noisy samples  (and the magnitude of error in the case of numerical values) for random and systematic noise, which we refer to as High ($\beta=0.5$, $\sigma_1=\sigma_2=5$), Medium ($\beta=0.3$, $\sigma_1=\sigma_2=3$) and Low ($\beta=0.2$, $\sigma_1=\sigma_2=1$).

 For adversarial attacks, we use methods that take into account specific ML models. Specifically, we use data poisoning techniques at training, and evasion attack methods at inference. For the part of our evaluation that focuses on training time, we generate poisoned samples using the back-gradient method~\cite{gradientPoison}. Since, this type of poisoning is specific to different downstream models we consider different dataset-model combinations in our evaluation. For the part of our evaluation that focuses on inference time, we use the projected gradient descent (PGD) attack ~\cite{PGDAttack}, a popular and effective white-box evasion attack method, to generate adversarial test samples. We use the implementation of PGD attack from ~\cite{art}. The corruption injected by the PGD attack is bounded by an infinity norm of $0.2$. The step size is $0.1$ and the number of iterations is $100$.
 
\blue{For common errors in real-world datasets, we consider missing values that cannot be detected during pre-processing (e.g. 99999 instead of NaN), multiplicative scaling of attributes (e.g. due to accidental changes of units), and typos in textual or categorical attributes. We synthesize this kind of noise as follows:
\begin{enumerate}
    \item If the corrupted cell is numerical, with probability 1/3 it will be 10 times larger, and with the same probability it will be 10 times smaller. Otherwise, the cell will contain a missing value.
    \item If the corrupted cell is categorical or textual, with probability 1/2 one of the character will be replaced by a random character. Otherwise, the cell will contain a missing value.
\end{enumerate}
For this kind of noise, we set the fraction of corrupted cells in the noisy samples as $\beta=0.3$. We also include Food, a dataset that contains real-world errors with manually labeled ground truth~\cite{holodetect}. It has 3 numerical, 6 categorical and 5 textual attributes. Out of its 3000 samples, 30.3\% are corrupted.
}
 
 As discussed in Section~\ref{sec:robust_ml}, we use artificially generated noise to create the training data for learning the victim-sample detectors. We now describe the type of noise we consider. Recall that we consider access to the set of clean sample $C$ and we augment this set with artificially corrupted data. We emphasize that the noise is always different than the noise considered in the training data. Since we assume that the type of noise in the test set is unknown in advance, the artificial noise contains a mixture of different levels of random noise ($(\beta=0.4, \sigma_1=4)$, $(\beta=0.25, \sigma_1=2)$, $(\beta=0.15, \sigma_1=1.5)$). We additionally augment $C$ with samples corrupted by random noise ($\beta=1, \sigma_1=0.25$) and adversarial samples generated by Fast Gradient Sign Method (FGSM)~\cite{Goodfellow2015ExplainingAH}(noise bounded by an infinite norm of $0.1$) to defend against adversarial noise. This corruption is different from the PGD attack considered during inference to ensure that we evaluate against a different noise distribution during online inference.

\newparagraph{Downstream ML Models} We consider the following downstream models: 1) A Logistic regression (LR) model with regularization parameter 1.0; 2) A Support Vector Machine (SVM) with a linear kernel and regularization parameter 1.0; 3) A fully-connected neural network (NN) with 2 hidden layers of size 100. We use a small model with 1 hidden layer of size 10 when we perform poisoning attacks due to the runtime complexity of the attack algorithm. The downstream models we choose cover different optimization objectives (logistic/hinge loss and convex/non-convex optimization objectives) and exhibit different robustness. \blue{Numerical attributes are encoded as their raw values for downstream models. Categorical and textual attributes are encoded in the same way as in Picket.}

%\begin{figure*}[t]
%    \centering
%    \subfigure[Random Noise]{\includegraphics[width=0.4\textwidth]{figures/RandomNoise.eps}}
%    \subfigure[Systematic Noise]{\includegraphics[width=0.4\textwidth]{figures/SystemNoise.eps}}
%    \subfigure[Adversarial Poisoning Noise]{\includegraphics[width=0.58\textwidth]{figures/PoisonNoise.eps}} 
%    \caption{AUROC of outlier detection for random and systematic noise.}
%    \label{fig:AUROC_RS}
%\end{figure*}

\newparagraph{Training-Time Baselines} We compare against three unsupervised outlier detection methods as follows: 1) Isolation Forest (IF)~\cite{isolation}, an approach similar to Random Forests but targeting outlier detection, 2) One-Class SVM (OCSVM)~\cite{OCSVM} with a radial basis function kernel, and 3) Robust Variational Autoencoders (RVAE)~\cite{RobustVAE}, a state-of-the-art generative model used for outlier detection on mixed-type tabular data. \blue{For IF, we use 100 base estimators in the ensemble. For RVAE, we use the default hyperparameter recommended by Eduardo et al.~\cite{RobustVAE}, which has 972,537 parameters. Note that the capacity of the RVAE model used in our experiments is larger than \NetworkName, which has 382,722 parameters. The detailed hyper-parameters we use for \NetworkName is reported in Appendix~\ref{sec:hyperparameters}.}

\newparagraph{Inference-Time Baselines}
We compare against: 1) victim-sample detectors based, 2) na\"ive confidence-based, and 3) adversarial data detection methods.

Methods based on per-class victim sample detectors follow the same strategy as \ModelName but use different features. We consider: 1) \emph{Raw Feature (RF)}, the binary classifiers only use the raw features of the data; 2) \emph{RVAE}, the binary classifiers use only the cell-level probability of being outliers provided by RVAE as features; 3) \emph{RVAE+}, the classifiers use a combination of the features from the two methods above.

We also consider the next na\"ive methods: 1) \emph{Calibrated Confidence Score (CCS)}, which assumes that the predictions of the downstream model have lower confidence for victim samples than clean samples. We calibrate the confidence scores of the downstream models using temperature scaling~\cite{calibration}. 2) \emph{$k$-Nearest Neighbors (KNN)}, which assumes that a victim sample has a different prediction from its neighbors. We use different distances for different types of attributes. For numerical attributes, the distance is $d / 0.05$ if $d \leq 0.05$, where $d$ is the difference between two normalized values; the distance is $1$ if $d > 0.05$. For categorical attributes, we use the Hamming distance and for text attributes the cosine distance. We set $k$ to $10$.

We consider two methods of adversarial sample detection: \emph{The Odds are Odd (TOAO)}~\cite{odds}, which detects adversarial samples based on the change in the distribution of the prediction logit values after the injection of random noise. It adds Gaussian, Bernoulli, and Uniform noise of different magnitude and takes the majority vote of all noise sources. 2) \emph{Model with Outlier Class (MWOC)}~\cite{MWOC}, which assumes that the feature distribution of adversarial samples is different from that of benign samples and adds a new outlier class to the downstream model to characterize the distribution of adversarial samples.

For a fair evaluation of baselines against \ModelName, we also reveal the augmented version of $C$ used to learn the victim-sample detectors in \ModelName to competing methods so that they fine-tune their models to noise (RF, RVAE, RVAE+, MWOC, \ModelName), or use it to find a good threshold (CCS, KNN, TOAO).

\newparagraph{Metrics} For training-time outlier detection, we report the area under the receiver operating characteristic curve (AUROC). We use AUROC since it is an aggregate measure of performance across all possible threshold settings. We also consider the test accuracy of downstream models. For victim sample detection, we report the $F_1$ scores of the classification between correctly classified samples and victim samples.

\newparagraph{Evaluation Protocol} All experiments are repeated five times with different random seeds that control train-test split and noise injection; the mean is reported. \blue{We also perform one-sided paired t-tests when we compare the examined methods. A method is considered significantly better than another one if the p value is less than 0.05. In addition, we provide a cross-validation-based evaluation in Appendix~\ref{sec:cross_validation} that examine the performance of outlier detection on unseen data.} 

\begin{figure}
    \centering
	\includegraphics[width=0.45\textwidth]{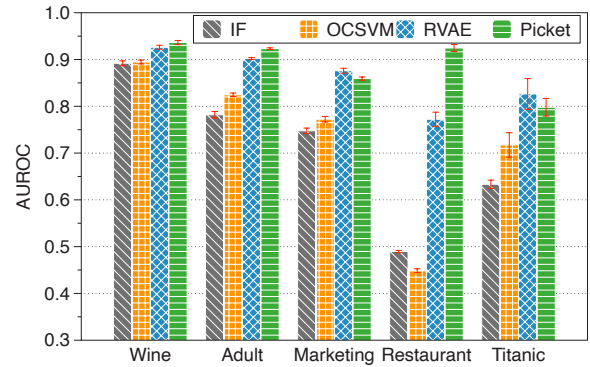}
    \caption{AUROC of outlier detection for random noise. The error bars represent the standard errors. \blue{Picket is significantly better (with p value less than 0.05) than the others on Wine, Adult, Marketing and Restaurant.}}
    \label{fig:AUROC_RS_rand}
\end{figure}
\subsection{Training-Time Evaluation}\label{sec:exp_trtime}
We evaluate the performance of different methods on detecting erroneous points in the training data. 
We then evaluate how these methods affect the performance of downstream models.
\blue{We also provide a study on synthetic datasets in Appendix~\ref{sec:synthetic_data} to see how these outlier detection methods perform when certain aspects of the data and noise vary.}
\begin{figure}
    \centering
	\includegraphics[width=0.45\textwidth]{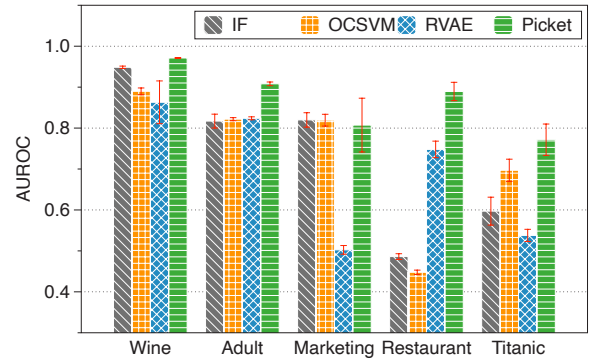}
    \caption{AUROC of outlier detection for systematic noise. The error bars represent the standard errors. \blue{Picket is significantly better (with p value less than 0.05) than the others on Wine, Adult, Restaurant and Titanic.}}
    \label{fig:AUROC_RS_syst}
\end{figure}

\begin{figure}
    \centering
	\includegraphics[width=0.45\textwidth]{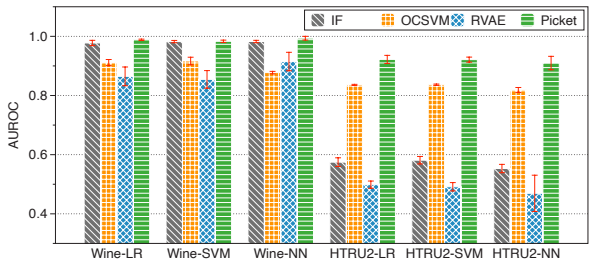}
    \caption{AUROC of outlier detection for poisoned samples. The error bars represent the standard errors. \blue{Picket is significantly better (with p value less than 0.05) than the others on all the combinations except Wine-SVM.}}
    \label{fig:AUROC_RS_Adv}
\end{figure}

\newparagraph{Detecting Corrupted Training Examples}
 Figures~\ref{fig:AUROC_RS_rand},~\ref{fig:AUROC_RS_syst}, and~\ref{fig:AUROC_RS_Adv} show the AUROC obtained by the methods for different types of noise, when 20\% of the samples are corrupted. The results for random and systematic noise correspond to Medium level noise. Results for Low and High levels are reported in Appendix~\ref{sec:train_time_low_high}. For Figure~\ref{fig:AUROC_RS_Adv}, note that the poisoned samples are model-specific and hence we report the dataset model combination on the x-axis. Due to data poisoning being limited to numerical data, we only evaluate on Wine and HTRU.
As shown, \emph{\ModelName is the only approach that consistently achieves an AUROC of close to or more than 0.8 for all datasets and for all noise settings.} Other methods achieve comparable performance in some settings but they are not consistent across diverse settings. IF and OCSVM perform poorly on datasets with textual attributes (Restaurant and Titanic) due to their limited capacity to handle text-based attributes. RVAE works quite well under random noise, but its performance drops a lot when it comes to systematic noise, which shows that it is not robust against noise that introduces bias. In the presence of poisoned data, we find that IF performs well on Wine but poorly on HTRU2, but OCSVM shows the opposite. A possible reason is that the two datasets exhibit different types of correlation between attributes, and the two methods are good at capturing only one of them. RVAE shows poor performance for both datasets. 

\begin{figure}
    \centering
	\includegraphics[width=0.45\textwidth]{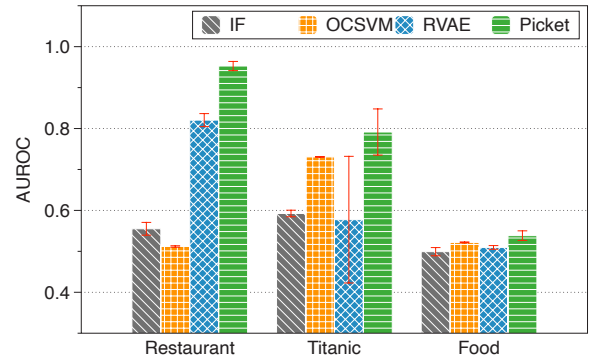}
    \caption{\blue{AUROC of outlier detection for common errors in the real world. The error bars represent the standard errors.} \blue{Picket is significantly better (with p value less than 0.05) than the others on all the datasets.}}
    \label{fig:AUROC_Real}
\end{figure}

\blue{For common errors in the real world, the results are shown in Figure~\ref{fig:AUROC_Real}. We add synthetic errors of this type to Titanic and Restaurant, where 20\% are corrupted. We choose these two because they contain textual attributes for typos. We also report the results on Food with real-world noise. We can see that on Restaurant and Titanic, \ModelName outperforms the others by more than 6 points. On Food, all the methods perform poorly. This is because the noise level in Food is very low, and therefore hard to detect. In fact, the real noise contained in Food does not have a significant effect on the downstream models (as is shown in Table~\ref{tab:real_ds_acc}).}

\blue{We also study how the fraction of corrupted samples affects the performance of detection (see Appendix~\ref{sec:train_time_row_fraction}). We find that \ModelName keeps a relatively consistent performance when the fraction of corrupted samples varies.}

% \begin{figure*}[t]
%     \centering
%     \subfigure[]{\includegraphics[width=0.32\textwidth]{./figures/LR_Random_acc.pdf}}
%     \subfigure[]{\includegraphics[width=0.32\textwidth]{./figures/SVM_Random_acc.pdf}} 
%     \subfigure[]{\includegraphics[width=0.32\textwidth]{./figures/NN_Random_acc.pdf}}
%     \subfigure[]{\includegraphics[width=0.32\textwidth]{./figures/LR_Systematic_acc.pdf}} 
%     \subfigure[]{\includegraphics[width=0.32\textwidth]{./figures/SVM_Systematic_acc.pdf}}
%     \subfigure[]{\includegraphics[width=0.32\textwidth]{./figures/NN_Systematic_acc.pdf}}
%     \caption{Effect on downstream models (random and systematic noise).}
%     \label{fig:effectRS}
% \end{figure*}

\newparagraph{Effect on Downstream Models}
We also study the effect of different filtering methods on the downstream models. For each method, we filter 20\% of the samples with highest outlier scores, and train different downstream models on the resulting training set.
For each dataset, the test set is fixed and contains only clean data.
As reference points, we also include the test accuracy when 1) the training data is clean without corruption (CL), and 2) the training data is corrupted but no filtering (NF) is performed. 
%These baselines correspond to the best-case and worst-case performance. 
\blue{Note that in the CL and NF cases, the sample size is different from the rest since there is no filtering in these two. As a side effect of filtering, the decrease in sample size will also affect the performance of the downstream model. We want to include such an effect in our comparison, so we use CL and NF with no sample filtered out as baselines.}

% \tabcolsep=0.08cm
% \begin{table}[]
% \caption{Test accuracy of downstream models under adversarial poisoning attacks and different filtering methods.}
% \label{tab:poison_ds_acc}
% \footnotesize
% \begin{tabular}{c|c|cccc|ccc}
% \hline
% \multicolumn{1}{c|}{Dataset} & \multicolumn{1}{c|}{\thead{Downstream \\ Model}} & IF & OCSVM & RVAE & \ModelName & CL & NF & PF \\ \hline \hline
% \multirow{3}{*}{Wine}            & LR  & 0.7261 & 0.6976 & 0.7051 & 0.7312 & 0.7349 & 0.6745 & 0.7349   \\                  
%                                 & SVM  & 0.7286 & 0.6933 & 0.7082 & 0.7310 & 0.7386 & 0.6727 & 0.7386  \\ 
%                               & NN  &  &  &  &  &  &  &   \\ \hline
% \multirow{3}{*}{HTRU2}            & LR & 0.8884 & 0.9015 & 0.8811 & 0.9067 & 0.9396 & 0.8799 & 0.9396  \\ 
%                               & SVM   & 0.8884 & 0.8979 & 0.8887 & 0.9232 & 0.9424 & 0.8832 & 0.9424  \\ 
%                               & NN  & 0.8671 & 0.8707 & 0.8643 & 0.9000 & 0.9280 & 0.8646 &  0.9280 \\ \hline
% \end{tabular}
% \end{table}

\tabcolsep=0.08cm
\begin{table}
\center
\caption{Test accuracy of downstream models under adversarial poisoning attacks and different filtering methods. \blue{The numbers are made bold when the corresponding method is significantly better (with p value less than 0.05) than all the others.}}
\label{tab:poison_ds_acc}
\scriptsize
\begin{tabular}{c|c|cccc|cc}
\hline
\multicolumn{1}{c|}{Dataset} & \multicolumn{1}{c|}{\thead{\scriptsize{DM*}}} & IF & OCSVM & RVAE & \ModelName & CL & NF \\ \hline \hline
\multirow{3}{*}{Wine}            & LR  & 0.7261 & 0.6976 & 0.7051 & \textbf{0.7312} & 0.7349 & 0.6745  \\                  
                                & SVM  & 0.7286 & 0.6933 & 0.7082 & 0.7310 & 0.7386 & 0.6727  \\ 
                              & NN  & 0.7210 & 0.6894 & 0.7035 & \textbf{0.7320} & 0.7365 & 0.6722    \\ \hline
\multirow{3}{*}{HTRU2}            & LR & 0.8884 & 0.9015 & 0.8811 & \textbf{0.9067} & 0.9396 & 0.8799  \\ 
                              & SVM   & 0.8884 & 0.8979 & 0.8887 & \textbf{0.9232} & 0.9424 & 0.8832  \\ 
                              & NN  & 0.8671 & 0.8707 & 0.8643 & \textbf{0.9000} & 0.9280 & 0.8646 \\ \hline
\end{tabular}
\begin{tablenotes}
      \item *DM = Downstream Model.
\end{tablenotes}
\end{table}

First, we consider the case of data poisoning since this type of corruption has the most significant effect on the downstream models. We measure the test accuracy of the downstream models when poisoned data are injected into the training stage. The results are shown in Table~\ref{tab:poison_ds_acc}. If we compare CL with NF we see an average drop of six accuracy points if corruptions are ignored and no filtering is applied. We find that all methods reduce the negative impact of the poisoned data and bring up the test accuracy. 
Nevertheless that \emph{\ModelName outperforms all competing baselines and yields test time accuracy improvements of more than three points in some cases.} We see that \ModelName is able to recover most of the accuracy loss for all models in the Wine dataset and comes very close to CL for HRTU2. All other methods exhibit smaller accuracy improvements and do not exhibit consistent behavior across datasets. 
%The next-best method is IF for Wine and OCVSM for HRTU2. 
%These results imply that these methods are also sensitive to intrinsic data characteristics. Overall we find the performance of different methods to be consistent with their outlier detection performance.

We also consider the cases of random and systematic noise, as well as common errors in the real world. 
These types of noise do not directly attack the downstream model. 
Moreover, most ML models are somewhat robust to these types of noise. 
As a result, we expect to see a small gap in the test accuracy between CL and NF, and all methods to perform comparably.
We report the results in these setups in Appendix~\ref{sec:downstream_acc_random_system} for completeness. 

\subsection{Inference-Time Evaluation}\label{sec:inf_time_eval}
We evaluate the different methods on victim sample detection under different types of noise. \blue{The $F_1$ scores under random (Medium level), systematic (Medium level), adversarial noise and common errors in the real world are reported in Table ~\ref{tab:random_f1},~\ref{tab:system_f1},~\ref{tab:adv_f1} and~\ref{tab:real_f1}. Food with real-world noise is not reported since we cannot find enough victim samples from it.} We report results for High and Low noise in Appendix~\ref{sec:test_time_low_high}.

\tabcolsep=0.05cm
\begin{table*}
\center
\caption{$F_1$ scores of victim sample detection at inference time under random noise (Medium level). \blue{The numbers are made bold when the corresponding method is significantly better (with p value less than 0.05) than all the others.}}
\label{tab:random_f1}
\scriptsize
\begin{tabular}{c|c|cccccccc}
\hline
\multicolumn{1}{c|}{Dataset} & \multicolumn{1}{c|}{\thead{\scriptsize{DM*}}} & RF & RVAE & RVAE+ & CCS & KNN & TOAO & MWOC & \ModelName \\ \hline \hline
\multirow{3}{*}{Wine}            & LR  & 0.7690 & 0.7786 & 0.8172 & 0.6667 & 0.6686 & 0.6813 & 0.7150 & 0.8094    \\                  
                                & SVM  & 0.7812 & 0.7859 & 0.8254 & 0.6667 & 0.6750 & 0.6858 & 0.7622 & 0.8223  \\ 
                              & NN  & 0.7125 & 0.7470 & \textbf{0.7833} & 0.5896 & 0.6669 & 0.5107 & 0.6988 & 0.7631 \\ \hline
\multirow{3}{*}{Adult}            & LR & 0.8352 & 0.7403 & 0.8489 & 0.6692 & 0.7866 & 0.2224 & 0.6725 & \textbf{0.8602} \\ 
                              & SVM  & 0.8434 & 0.7416 & 0.8553 & 0.6688 & 0.8060 & 0.4696 & 0.6215 & \textbf{0.8658}  \\ 
                              & NN  & 0.8131 & 0.7127 & 0.8315 & 0.5117 & 0.6891 & 0.3216 & 0.7132 & \textbf{0.8411} \\ \hline
\multirow{3}{*}{Restaurant}            & LR  & 0.7726 & \hspace{0.1cm}--\textsuperscript{\#} & -- & 0.7403 & 0.6456 & 0.6457 & 0.7459 & \textbf{0.8266}   \\  
                              & SVM  & 0.6854 & -- & -- & 0.6796 & 0.6628 & 0.6596 & 0.5580 & \textbf{0.7618}\\ 
                              & NN  & 0.7605 & -- & -- & 0.6994 & 0.6609 & 0.6110 & 0.7025 & \textbf{0.8203} \\ \hline
\multirow{3}{*}{Marketing}            & LR & 0.8366 & 0.6623 & 0.8403 & 0.7567 & 0.7815 & 0.6666 & 0.7996 & \textbf{0.8549}   \\  
                              & SVM  & 0.8461 & 0.6689 & 0.8501 & 0.7527 & 0.7886 & 0.5133 & 0.8109 & \textbf{0.8607}  \\ 
                              & NN  & 0.7931 & 0.6650 & 0.8029 & 0.6588 & 0.7050 & 0.6648 & 0.7265 & 0.8162 \\ \hline
\multirow{3}{*}{Titanic}            & LR  & 0.8257 & -- & -- & 0.6990 & 0.6562 & 0.1409 & 0.7736 & \textbf{0.8424}  \\  
                              & SVM  & 0.8482 & -- & -- & 0.6658 & 0.6436 & 0.4652 & 0.7932 & 0.8528   \\ 
                              & NN  & 0.8393 & -- & -- & 0.6631 & 0.6387 & 0.2575 & 0.7566 & 0.8483\\ \hline
\end{tabular}
\begin{tablenotes}
      \item *DM = Downstream Model. \textsuperscript{\#}RVAE is not applicable to text attributes.
\end{tablenotes}
\end{table*}
From the tables, we can see that \ModelName has the best performance in most cases. By comparing RF and our method, we show that \emph{the reconstruction loss features provided by \NetworkName are good signals to help identify victim samples}. Such signals are better than those provided by RVAE since our method outperforms RVAE+ most of the time. TOAO performs consistently poorly since the assumption it relies on does not hold for the downstream models and datasets we consider. It works for image classification with complex convolutional neural networks under adversarial settings since adding random noise to images could eliminate the effect of adversarial noise. However, for tabular datasets and models which are not that complex, especially when the noise is not adversarial, adding random noise does not make a big difference. Another method from the adversarial learning literature (MWOC) works well in some cases even if the noise is not adversarial.  
\tabcolsep=0.05cm
\begin{table*}
\center
\caption{$F_1$ scores of victim sample detection at inference time under systematic noise (Medium level). \blue{The numbers are made bold when the corresponding method is significantly better (with p value less than 0.05) than all the others.}}
\label{tab:system_f1}
\scriptsize
\begin{tabular}{c|c|cccccccc}
\hline
\multicolumn{1}{c|}{Dataset} & \multicolumn{1}{c|}{\thead{\scriptsize{DM*}}} & RF & RVAE & RVAE+ & CCS & KNN & TOAO & MWOC & \ModelName \\ \hline \hline
\multirow{3}{*}{Wine}            & LR  & 0.6883 & 0.4987 & 0.6619 & 0.6667 & 0.6499 & 0.3152 & \textbf{0.7937} & 0.7046   \\                
                                & SVM  & 0.6785 & 0.5056 & 0.6630 & 0.6667 & 0.6325 & 0.3399 & \textbf{0.7957} & 0.6973 \\ 
                              & NN  & 0.6760 & 0.6134 & 0.5689 & 0.6865 & 0.6659 & 0.3765 & \textbf{0.7190} & 0.6034 \\ \hline
\multirow{3}{*}{Adult}            & LR & 0.8281 & 0.6960 & 0.8342 & 0.6695 & 0.7488 & 0.1864 & 0.7430 & \textbf{0.8501}  \\ 
                              & SVM & 0.8414 & 0.6729 & 0.8428 & 0.6694 & 0.7900 & 0.3617 & 0.6646 & \textbf{0.8643}  \\ 
                              & NN  & 0.8108 & 0.6534 & 0.8245 & 0.5439 & 0.6808 & 0.2195 & 0.7850 & \textbf{0.8336} \\ \hline
\multirow{3}{*}{Restaurant}   & LR  & 0.7773 & \hspace{0.1cm}--\textsuperscript{\#} & -- & 0.7419 & 0.6524 & 0.6496 & 0.7487 & \textbf{0.8255}  \\  
                              & SVM & 0.7275 & -- & -- & 0.7093 & 0.6475 & 0.6356 & 0.6125 & \textbf{0.7845} \\ 
                              & NN & 0.7628 & -- & -- & 0.7010 & 0.6579 & 0.6051 & 0.7003 & \textbf{0.8126}  \\ \hline
\multirow{3}{*}{Marketing}            & LR  & 0.8358 & 0.6504 & 0.8403 & 0.7623 & 0.7770 & 0.6090 & 0.8068 & \textbf{0.8514}  \\  
                              & SVM & 0.8501 & 0.6575 & 0.8552 & 0.7716 & 0.7817 & 0.6185 & 0.8208 & \textbf{0.8638}  \\ 
                              & NN  & 0.8036 & 0.6355 & 0.8098 & 0.6649 & 0.7074 & 0.6635 & 0.7035 & 0.8118  \\ \hline
\multirow{3}{*}{Titanic}            & LR  & 0.8376 & -- & -- & 0.7349 & 0.6493 & 0.4076 & 0.7901 & 0.8438 \\  
                              & SVM  & 0.8224 & -- & -- & 0.6674 & 0.6387 & 0.5592 & 0.7593 & 0.8412    \\ 
                              & NN & 0.8112 & -- & -- & 0.6660 & 0.6333 & 0.3139 & 0.7462 & 0.8159 \\ \hline
\end{tabular}
\begin{tablenotes}
      \item *DM = Downstream Model. \textsuperscript{\#}RVAE is not applicable to text attributes.
\end{tablenotes}
\end{table*}

% \subsection{Summary of Micro-Benchmarks}\label{sec:summary_microbenchmarks}
% We perform a series of micro-benchmarks to evaluate different design decisions related to \ModelName. The detailed experiments are presented in Appendix~\ref{sec:microbenchmarks}. Here we want to highlight our finding that the two-stream attention of \ModelName is crucial in obtaining good performance. Specifically, we evaluate the two-stream self-attention against a version of \NetworkName that uses a value-only or a schema-only attention. Our results demonstrate that there are datasets where schema-only outperforms value-only and vice-verse. However, using both streams yields improved performance that exceeds both single-stream options. In Appendix~\ref{sec:microbenchmarks}, we also present additional results on the robustness of \NetworkName's training and the effect of different types of artificial noise in training the victim-sample detectors. 

\tabcolsep=0.05cm
\begin{table*}
\center
\caption{$F_1$ scores of victim sample detection at inference time under adversarial noise. \blue{The numbers are made bold when the corresponding method is significantly better (with p value less than 0.05) than all the others.}}
\label{tab:adv_f1}
\scriptsize
\begin{tabular}{c|c|cccccccc}
\hline
\multicolumn{1}{c|}{Dataset} & \multicolumn{1}{c|}{\thead{\scriptsize{DM*}}} & RF & RVAE & RVAE+ & CCS & KNN & TOAO & MWOC & \ModelName \\ \hline \hline
\multirow{3}{*}{Wine}            & LR  & 0.7899 & 0.6758 & 0.7905 & 0.8233 & 0.6660 & 0.5030 & 0.8287 & 0.8197   \\                  
                                & SVM  & 0.7951 & 0.6791 & 0.8004 & 0.8119 & 0.6660 & 0.5743 & 0.8324 & 0.8291 \\ 
                              & NN  & 0.7400 & 0.6922 & 0.7347 & 0.6815 & 0.6663 & 0.6620 & 0.3980 & 0.7442 \\ \hline
\multirow{3}{*}{HTRU2}            & LR  & 0.8727 & 0.0160 & 0.8699 & 0.6667 & 0.6654 & 0.5123 & 0.8389 & 0.8757 \\ 
                              & SVM  & 0.9409 & 0.3436 & 0.9399 & 0.6667 & 0.6623 & 0.6456 & 0.2211 & \textbf{0.9438}   \\ 
                              & NN  & 0.9103 & 0.3007 & 0.9164 & 0.7258 & 0.6656 & 0.2873 & 0.7726 & 0.9201  \\ \hline
\end{tabular}
\begin{tablenotes}
      \item *DM = Downstream Model.
\end{tablenotes}
\end{table*}

\tabcolsep=0.05cm
\begin{table*}
\center
\caption{\blue{$F_1$ scores of victim sample detection at inference time under common errors in the real world.} \blue{The numbers are made bold when the corresponding method is significantly better (with p value less than 0.05) than all the others.}}
\label{tab:real_f1}
\scriptsize
\begin{tabular}{c|c|cccccccc}
\hline
\multicolumn{1}{c|}{Dataset} & \multicolumn{1}{c|}{\thead{\scriptsize{DM*}}} & RF & RVAE & RVAE+ & CCS & KNN & TOAO & MWOC & \ModelName \\ \hline \hline
\multirow{3}{*}{Restaurant} & LR & 0.7335 & \hspace{0.1cm}--\textsuperscript{\#} & -- & 0.7420 & 0.6527 & 0.5003 & 0.7330 & 0.7445          
   \\  
                              & SVM  & 0.6948 & -- & -- & 0.7168 & 0.6415 & 0.6104 & 0.6189 & 0.6928 \\ 
                              & NN  & 0.7716 & -- & -- & 0.6818 & 0.6633 & 0.5470 & 0.6762 & 0.7713 \\ \hline
\multirow{3}{*}{Titanic}            & LR  & 0.5633 & -- & -- & 0.3350 & 0.6792 & 0.5934 & 0.4740 & \textbf{0.8905}  \\  
                              & SVM  & 0.6304 & -- & -- & 0.4412 & 0.6798 & 0.4374 & 0.5706 & \textbf{0.8651}     \\ 
                              & NN  & 0.6100 & -- & -- & 0.4140 & 0.6816 & 0.6855 & 0.8093 & 0.8205 \\ \hline
\end{tabular}
\begin{tablenotes}
      \item *DM = Downstream Model. \textsuperscript{\#}RVAE is not applicable to text attributes.
\end{tablenotes}
\end{table*}

\subsection{Micro-Benchmarks}\label{sec:microbenchmarks}
We perform a series of micro-benchmarks to evaluate different design decisions related to \ModelName.

\begin{figure}
\centering
\includegraphics[width=0.4\textwidth]{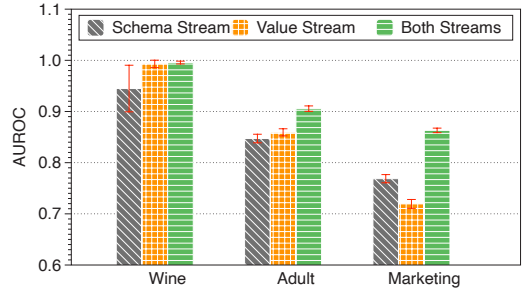}
\caption{\blue{Outlier detection under different stream settings. The error bars represent the standard errors.}}
\label{fig:streamEffect}
\end{figure}

\newparagraph{Effectiveness of the Two-Stream Self-Attention}\\
% \label{sec:twoStreamEffect}
We perform an ablation study to validate the effectiveness of the two-stream self-attention. We evaluate the performance of outlier detection with only one stream and with both. The results are depicted in Figure ~\ref{fig:streamEffect}. In the case of one stream, we simply let the output of self-attention layer $l$ be either $H_i^{(l)}$ for the value stream, or $G_i^{(l)}$ for the schema stream instead of $H_i^{(l)}+G_i^{(l)}$, where $i$ is the index of the attribute. \blue{For fair comparison, we expand the dimension of all the vectors involved in the computation of multi-head self-attention functions and feed-forward networks by a factor of $\sqrt{2}$ in the one-stream cases, so that the network capacity (number of parameters) remains the same after the pruning of one stream. }We use three setups: Wine with poisoning attack on NN, Adult with systematic noise (Medium level), and Marketing with random noise (Medium level).

From Figure ~\ref{fig:streamEffect}, we see that for Adult and Marketing, \NetworkName with two streams outperforms both one-stream options. For Wine, the value stream itself works fine, but a combination of the two streams does not impair the performance of the model. Neither of the two one-stream options demonstrates obvious superiority over the other one, since there are cases that the value stream performs better than the schema stream, and cases that the opposite happens. 

\begin{figure}
\centering
\includegraphics[width=0.4\textwidth]{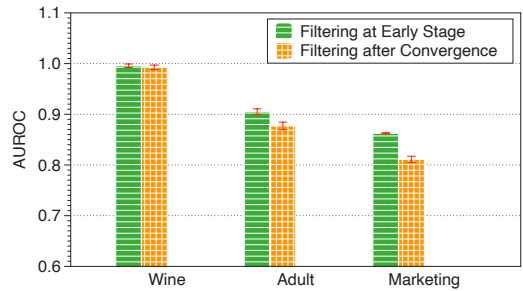}
\caption{Early vs. after-convergence filtering. The error bars represent the standard errors.}
\label{fig:earlyEffect}
\end{figure}

\newparagraph{Effectiveness of the Early Filtering Mechanism}\\ We validate the effectiveness of early filtering by comparing the performance of outlier detection at the early stage of \NetworkName's training to that after convergence. The results are shown in Figure ~\ref{fig:earlyEffect}. We use the setup from the previous micro-benchmark.

Figure ~\ref{fig:earlyEffect} shows that filtering at early stages consistently outperforms filtering after convergence. The reason is that in the early stage of training, the model is less likely to overfit to the input, and therefore the reconstruction loss of the outliers differs from that of the clean samples more.

\begin{figure*}[t]
    \centering
    \subfigure[Wine-LR]{\includegraphics[width=0.32\textwidth]{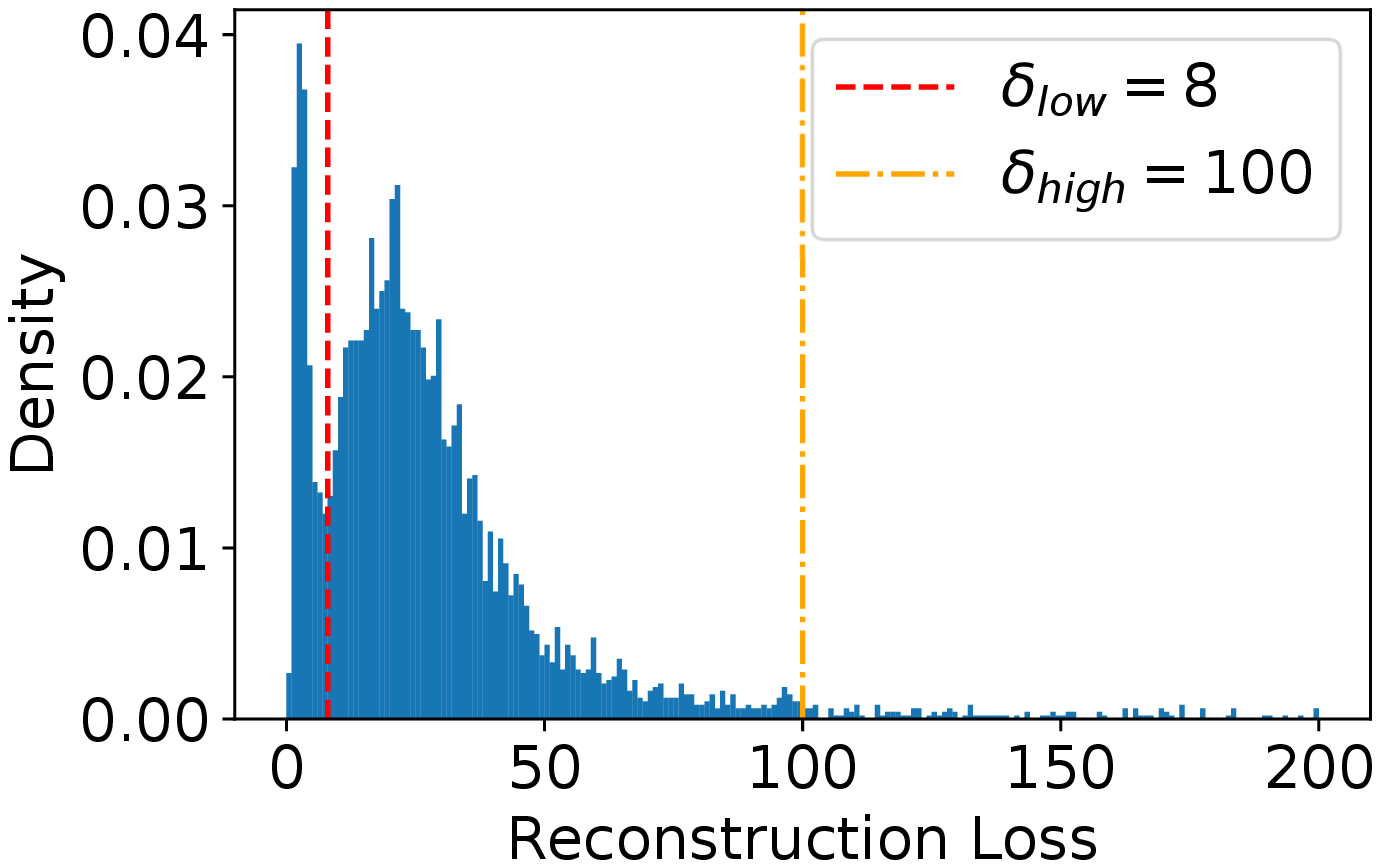}}
    \subfigure[Wine-SVM]{\includegraphics[width=0.32\textwidth]{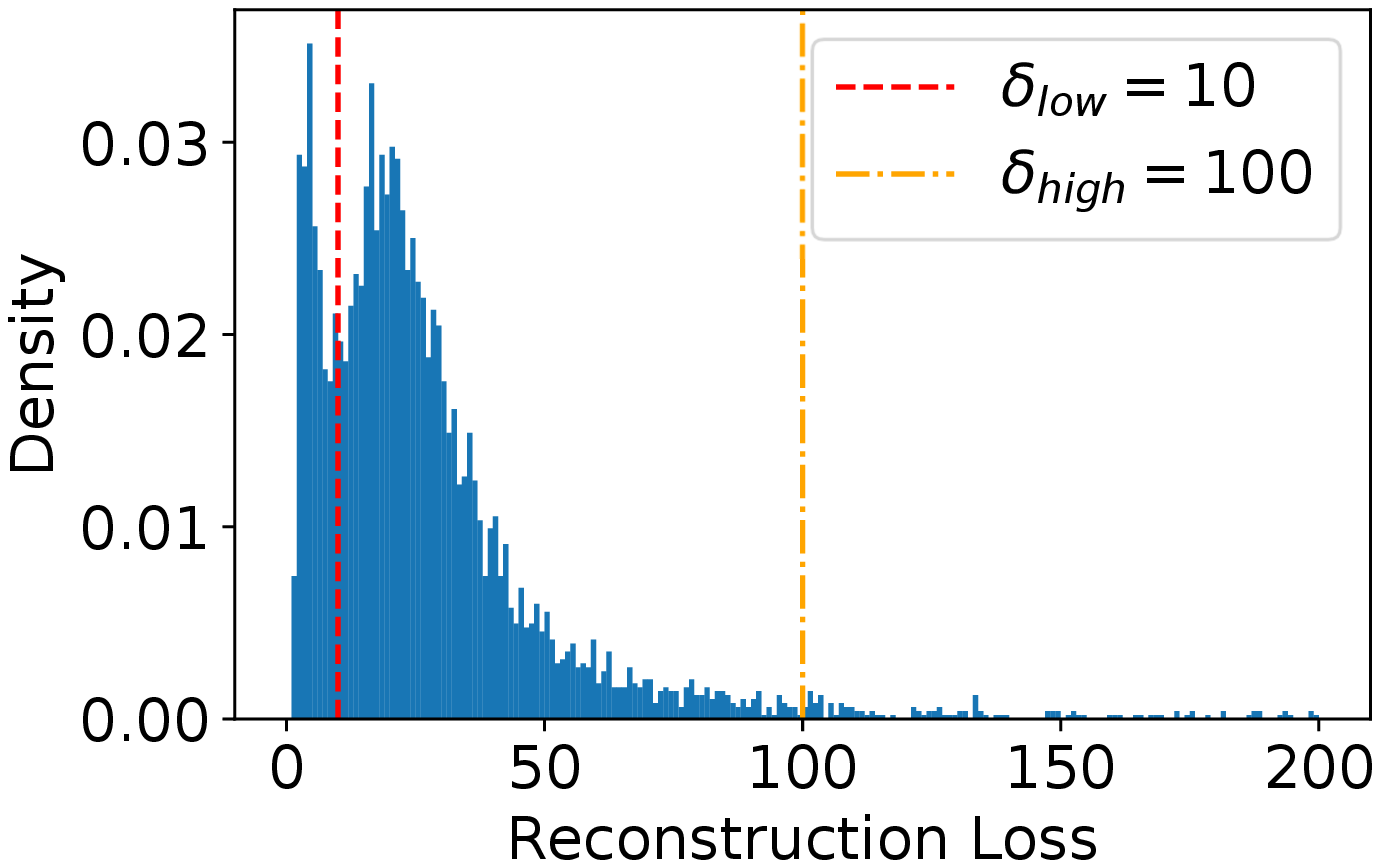}}
    \subfigure[Wine-NN]{\includegraphics[width=0.32\textwidth]{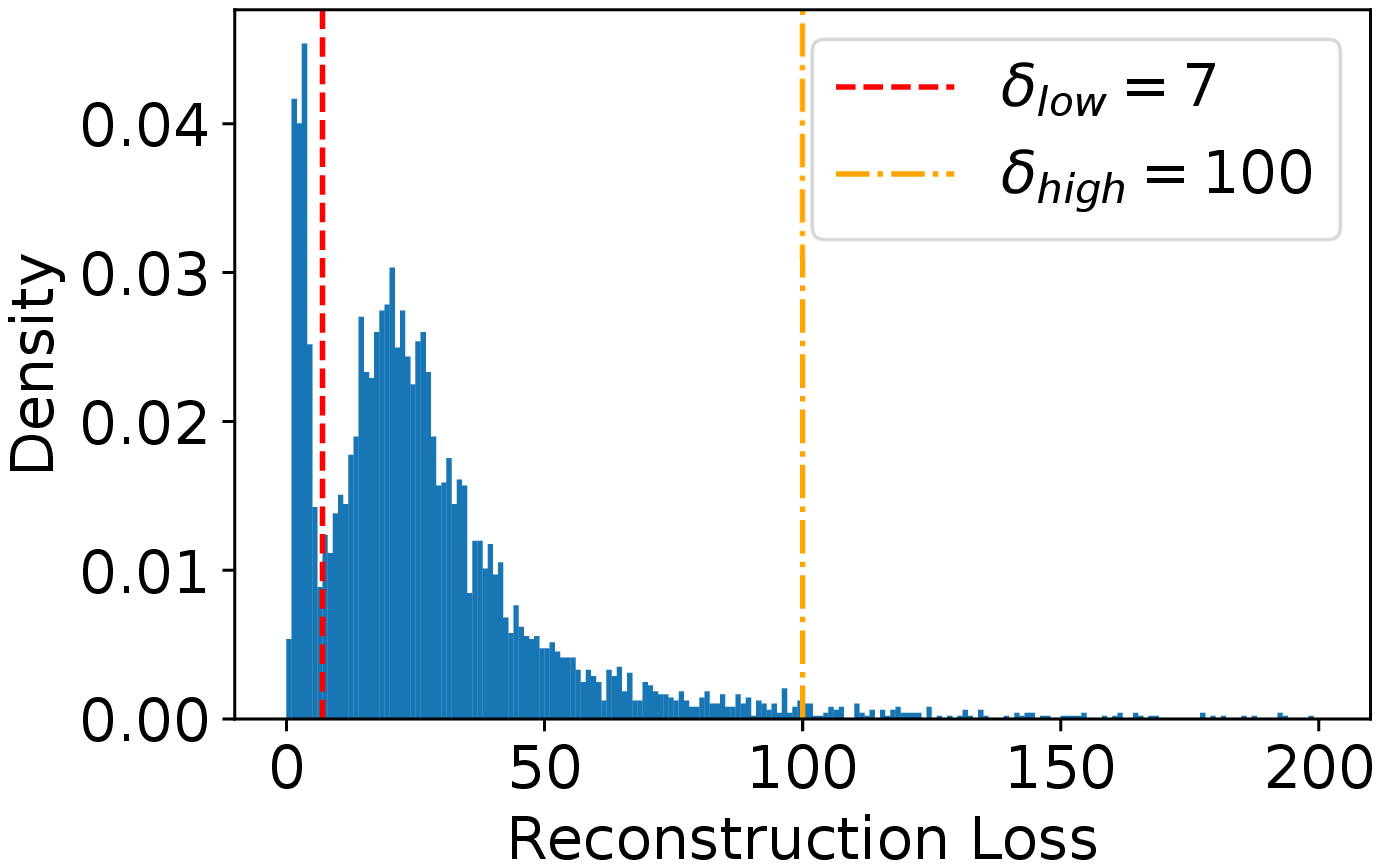}}
    \subfigure[HTRU2-LR]{\includegraphics[width=0.32\textwidth]{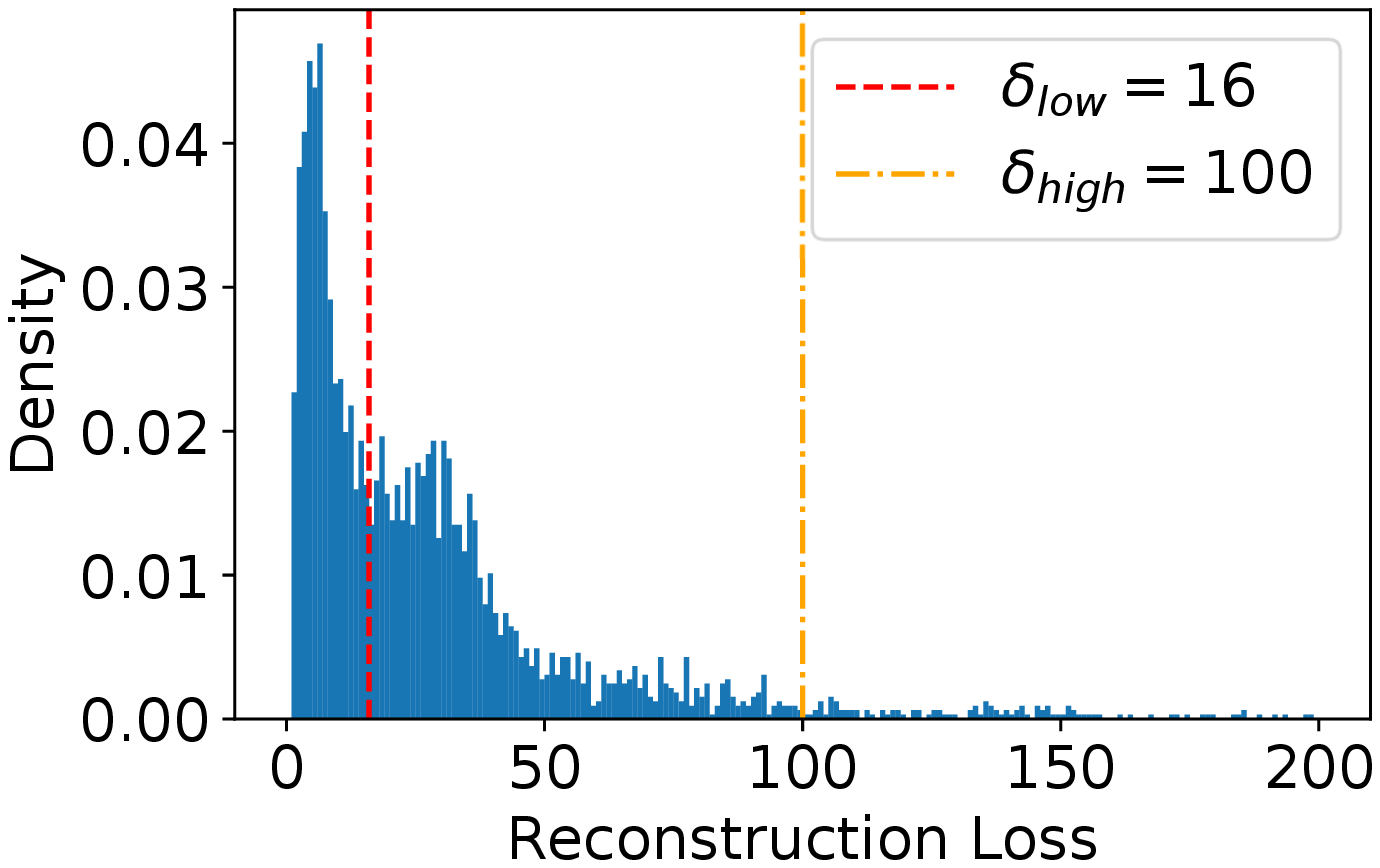}}
    \subfigure[HTRU2-SVM]{\includegraphics[width=0.32\textwidth]{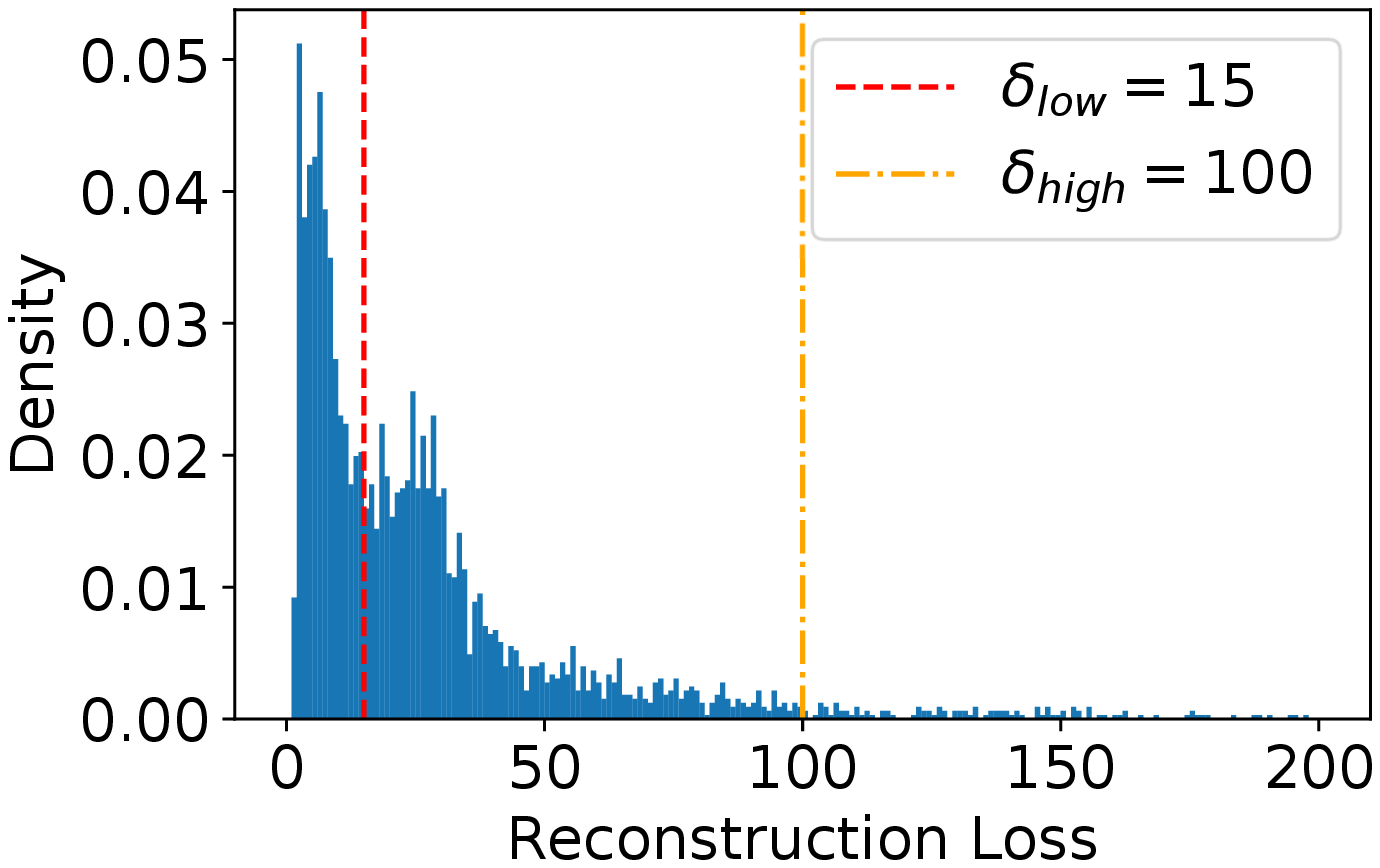}}
    \subfigure[HTRU2-NN]{\includegraphics[width=0.32\textwidth]{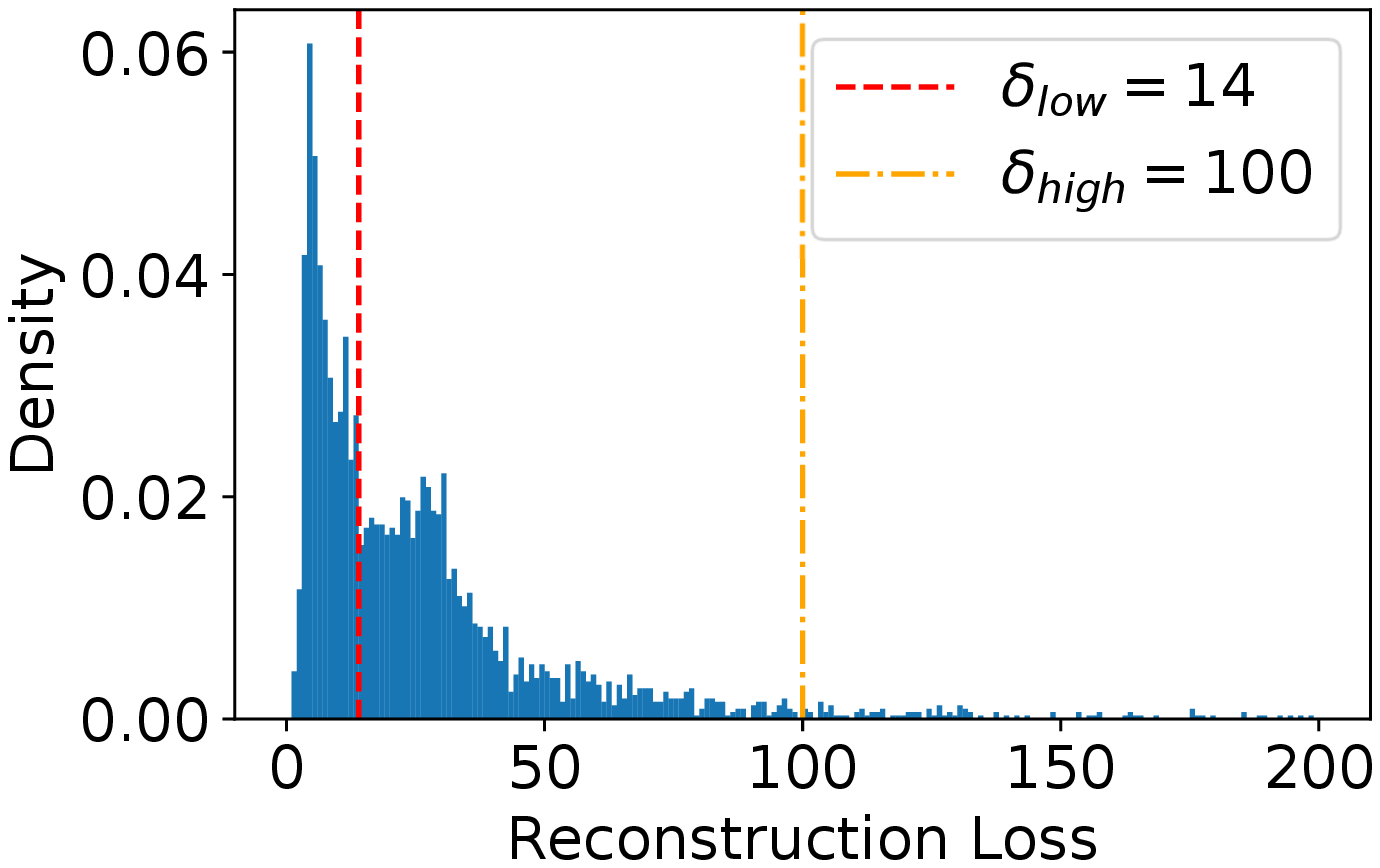}}
    \caption{\blue{Histograms of the reconstruction loss under different dataset-model combinations and the thresholds $\delta_\text{low}$, $\delta_\text{high}$.}}
    \label{fig:hist_and_thres}
    \vspace{-10pt}
\end{figure*}

\newparagraph{Histogram-Based Threshold Selection}
\blue{We validate the effectiveness of the histogram-based threshold selection strategy mentioned in Section~\ref{sec:training_picketnet}. To better illustrate how it affects the downstream accuracy, we use Wine and HTRU2 with poisoning attacks (20\% of the samples are poisoned) where corruption has a significant effect on the downstream models. For each dataset and downstream model combination, we plot the histogram of the Picket reconstruction loss in Figure~\ref{fig:hist_and_thres}, and select the thresholds $\delta_\text{low}$ and $\delta_\text{high}$ accordingly so that the abnormal peaks and low-density tails are filtered out. We report the downstream accuracy after filtering with this strategy (Picket-Hist) in Table~\ref{tab:hist_and_thres}. Same as Section ~\ref{sec:exp_trtime}, we also report the downstream accuracy under CL and NF as reference points. The results show that Picket-Hist gets very close to CL where the data is clean, and much better than NF where no filtering is applied, which verifies the effectiveness of this threshold-selection strategy.}

\tabcolsep=0.08cm
\begin{table}
\center
\caption{\blue{Test accuracy of downstream models after filtering based on the histogram of reconstruction loss (Picket-Hist).}}
\label{tab:hist_and_thres}
\scriptsize
\begin{tabular}{c|c|ccc}
\hline
\multicolumn{1}{c|}{Dataset} & \multicolumn{1}{c|}{\thead{\scriptsize{Downstream Model}}} & Picket-Hist & CL & NF \\ \hline \hline
\multirow{3}{*}{Wine}            & LR  & 0.7500 & 0.7551 &  0.6846  \\                  
                                & SVM  & 0.7459 & 0.7530 & 0.6836\\ 
                              & NN  & 0.7133 & 0.7204 &  0.6561   \\ \hline
\multirow{3}{*}{HTRU2}            & LR & 0.9344 & 0.9435 & 0.8810  \\ 
                              & SVM   & 0.9375 & 0.9435 &  0.8856 \\ 
                              & NN  & 0.9207 & 0.9207 & 0.8720 \\ \hline
\end{tabular}
\end{table}

\newparagraph{Effectiveness of Per-Class Victim Sample Detectors}
We compare the performance of our per-class detectors against a unified detector and a score-based detector. The unified detector uses one single logistic regression model over the same features to distinguish between good and victim samples regardless of the downstream predictions. The score-based detector follows the logic of the training time outlier detector, i.e., it aggregates the reconstruction losses attribute-wise, and considers samples with high loss as victims. We perform the comparison on three datasets with all of the three downstream models: Wine with adversarial noise, Adult with systematic noise (Medium level) and Marketing with random noise (Medium level).

The result is shown in Table ~\ref{tab:perclass}. Per-Class Detectors outperform the other two, which validates the effectiveness of having one detector per-class. The unified detector performs poorly because the victim samples in one class differ from those in the other statistically, \blue{in which case one class may suffer from corruption in one group of attributes, while the other class may suffer from that in another group of attributes.} The score-based detector does not work well since it only has access to the noise level of the samples but does not consider the connection between corruptions and the downstream prediction. 

\tabcolsep=0.07cm
\begin{table}[]
\center
\caption{A comparison between the per-class detectors, the unified detector, and the score-based detector on inference time victim sample detection.}
\label{tab:perclass}
\scriptsize
\begin{tabular}{c|c|cccccccc}
\hline
\multicolumn{1}{c|}{Dataset} & \multicolumn{1}{c|}{\thead{\scriptsize{Downstream}\\\scriptsize{Model}}} & \thead{\scriptsize{Per-Class}\\\scriptsize{Detectors}} & \thead{\scriptsize{Unified}\\\scriptsize{Detector}} & \thead{\scriptsize{Score-based}\\\scriptsize{Detector}} \\ \hline \hline
\multirow{3}{*}{Wine}            & LR  & 0.8188 & 0.7023 & 0.6885 \\                  
                                & SVM & 0.8287 & 0.7152 & 0.7261 \\ 
                              & NN  & 0.7444 & 0.4027 & 0.6594 \\ \hline
\multirow{3}{*}{Adult}            & LR & 0.8489 & 0.6710 & 0.7197 \\ 
                              & SVM  & 0.8634 & 0.6983 & 0.7297   \\ 
                              & NN  & 0.8336 & 0.6785 & 0.7225  \\ \hline
\multirow{3}{*}{Marketing}  & LR  & 0.8553 & 0.7740 & 0.7343 \\ 
                            & SVM  & 0.8618 & 0.7774 & 0.7361  \\ 
                            & NN & 0.8152 & 0.7370 &  0.7174 \\ \hline
\end{tabular}
\end{table}

\newparagraph{Effectiveness of Mixed Artificial Noise}
We validate the effectiveness of our artificial noise setting (Mixed) by comparing it to the setting where the artificial noise is generated in the same way as the test time noise (Exact). The results are shown in Table~\ref{tab:artificial_noise}. We use the same datasets and test time noise as the previous micro-benchmark. We find that with mixed artificial noise, the performance of Picket is comparable to the setting where the exact noise distribution is known under random (see Marketing) and systematic noise (see Adult). Under adversarial noise (see Wine), Exact is better than Mixed but the gap is not excessively large.

\tabcolsep=0.07cm
\begin{table}[]
\center
\caption{F1 scores of Picket on victim sample detection under different artificial noise settings.}
\label{tab:artificial_noise}
\scriptsize
\begin{tabular}{c|c|cccccccc}
\hline
\multicolumn{1}{c|}{Dataset} & \multicolumn{1}{c|}{\thead{\scriptsize{Downstream}\\\scriptsize{Model}}} & \thead{\scriptsize{Mixed}} & \thead{\scriptsize{Exact}} \\ \hline \hline
\multirow{3}{*}{Wine}            & LR  & 0.8197 & 0.8646    \\                  
                                & SVM & 0.8291 & 0.8812  \\ 
                              & NN  & 0.7442  & 0.7790 \\ \hline
\multirow{3}{*}{Adult}            & LR & 0.8501 & 0.8372 \\ 
                              & SVM & 0.8643 & 0.8562 \\ 
                              & NN & 0.8336 & 0.8157 \\ \hline
\multirow{3}{*}{Marketing}   & LR & 0.8549 & 0.8544 \\ 
                              & SVM & 0.8607 & 0.8592   \\ 
                              & NN & 0.8162 & 0.8120 \\ \hline
\end{tabular}
\end{table}

\subsection{Fairness of Outlier Detection}
\blue{
We compute the equality of opportunity between majority and minority groups to check the fairness of outlier detection. Specifically, the opportunity $\gamma_\mathcal{G}$ for each group $\mathcal{G}$ is defined as the fraction of clean examples in that group that are kept after filtering:
\begin{equation*}
    \gamma_\mathcal{G} = N_{\mathcal{G}}^{\text{kept}} / N_{\mathcal{G}}^{\text{clean}}
\end{equation*}
where $N_{\mathcal{G}}^{\text{clean}}$ is the number of clean examples in group $\mathcal{G}$, and $N_{\mathcal{G}}^{\text{kept}}$ is the number of clean examples in $\mathcal{G}$ that are not filtered out. We report the difference of opportunity $\Delta \gamma = \gamma_{\mathcal{G}_m} - \gamma_{\mathcal{G}_M}$, where $\mathcal{G}_M$ is the majority group and $\mathcal{G}_m$ is the minority. $\Delta \gamma$ closer to $0$ indicates better fairness.}

\blue{We choose two demographic datasets, Adult and Marketing, to verify the fairness of the outlier detection methods. For each dataset, we pick one sensitive attribute at a time, and divide its value domain into majority and minority groups as follows:
\begin{enumerate}
    \item Sort the values by their frequency in descending order.
    \item Add values in order to the majority group until it covers more than 80\% of the examples.
    \item Add the rest of the values into the minority group.
\end{enumerate}
 We inject random and systematic noise of medium magnitude to 20\% of the examples, filter out 20\%, and report the difference of opportunity for each dataset-attribute combination in Table~\ref{tab:oppo_diff}. We can see that for most of the sensitive attributes, the difference of opportunity is less than $0.05$ if the data are filtered by \ModelName. However, for certain attributes (e.g. Marketing-Marital and Marketing-Language), the difference is quite large, which shows potential risk of unfairness. The other models also show bias towards the majority group for certain attributes. We defer the improvement of fairness as a future direction to explore.
}

\tabcolsep=0.07cm
\begin{table*}[]
\center
\caption{\blue{Difference of opportunity when 20\% of the examples are filtered out.}}
\label{tab:oppo_diff}
\scriptsize
\begin{tabular}{c|c|cccc}
\hline
\multicolumn{1}{c|}{\thead{\scriptsize{Noise}\\\scriptsize{Type}}} & \multicolumn{1}{c|}{Dataset-Attribute} & \thead{\scriptsize{IF}} & \thead{\scriptsize{OCSVM}} & \thead{\scriptsize{RVAE}} & \thead{\scriptsize{Picket}}\\ \hline \hline
\multirow{3}{*}{Random}            & Marketing-Marital & -0.0469 & -0.0821 & \textbf{-0.0196} & -0.1400 \\
& Marketing-Age & 0.0720 & -0.0071 & \textbf{0.0019} & 0.0216 \\
& Marketing-Education & -0.0436 & -0.0521 & -0.0192 & \textbf{0.0174} \\ 
& Marketing-Live & -0.1131 & -0.1242 & -0.0488 & \textbf{-0.0242} \\ 
& Marketing-Dual & -0.0226 & -0.0737 & -0.0357 & \textbf{-0.0089} \\ 
& Marketing-Hometype & -0.0865 & -0.1581 & -0.0634 & \textbf{-0.0458} \\
& Marketing-Ethnic & -0.1762 & -0.2258 & -0.0760 & \textbf{-0.0610} \\ 
& Marketing-Language & -0.5625 & -0.4739 & \textbf{-0.0753} & -0.3739 \\
& Adult-Workclass & -0.1290 & -0.0259 & -0.0277 & \textbf{-0.0042} \\ 
& Adult-Marital-status & -0.3111 & -0.0676 & \textbf{-0.0013} & -0.0706 \\  
& Adult-Relationship & -0.2545 & \textbf{0.0027} & 0.0081 & -0.0188 \\ 
& Adult-Race & -0.4452 & -0.0326 & \textbf{-0.0259} & -0.0515 \\  \hline
\multirow{3}{*}{Systematic}           & Marketing-Marital & -0.0541 & -0.1178 & \textbf{-0.0418} & -0.2031 \\ 
& Marketing-Age & 0.0902 & \textbf{-0.0051} & 0.0097 & 0.0270 \\ 
& Marketing-Education & -0.0366 & -0.0509 & -0.0164 & \textbf{0.0142} \\
& Marketing-Live & -0.0781 & -0.1333 & -0.0640 & \textbf{-0.0090} \\ 
& Marketing-Dual & -0.0275 & -0.0757 & \textbf{-0.0224} & -0.0244 \\
& Marketing-Hometype & -0.0995 & -0.1892 & -0.1182 & \textbf{-0.0528} \\
& Marketing-Ethnic & -0.1919 & -0.2465 & -0.1388 & \textbf{-0.0777} \\  
& Marketing-Language & -0.5981 & -0.5397 & \textbf{-0.1555} & -0.4791 \\
& Adult-Workclass & -0.1819 & -0.0079 & -0.0287 & \textbf{-0.0012} \\  
& Adult-Marital-status & -0.3158 & -0.0690 & \textbf{0.0026} & -0.2519 \\
& Adult-Relationship & -0.2444 & -0.0148 & \textbf{0.0067} & -0.0635 \\  
& Adult-Race & -0.4124 & -0.0560 & \textbf{-0.0088} & -0.0719 \\  
 \hline
\end{tabular}
\end{table*}

\subsection{Runtime and Scalability}
We report the training time of \NetworkName for each dataset in Table ~\ref{tab:runtime}. The device we use is a single NVIDIA Tesla V100-PCIE GPU with 32GB memory. Note that the current runtime has not been fully optimized.

\begin{figure}
\centering
\includegraphics[width=0.32\textwidth]{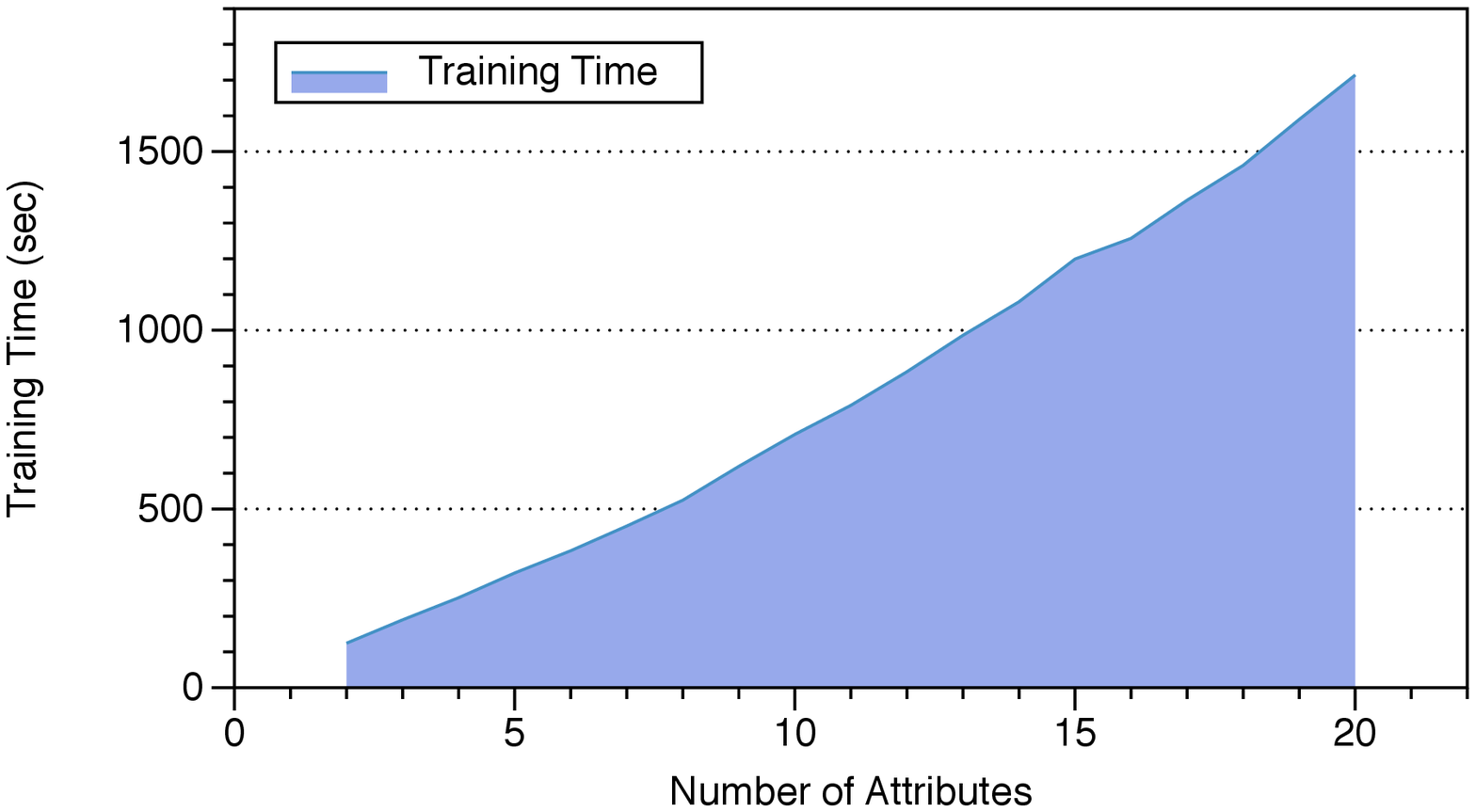}
\caption{Attribute-wise scalibility of \NetworkName}
\label{fig:runtime}
\end{figure}

We also study the attribute-wise scalibilty of \NetworkName using synthetic datasets. The datasets have a different number of attributes ranging from 2 to 20 with a increase step of one, while the other settings are the same (the dimension of $I_i^{(l)}$ and $P_i^{(l)}$ is fixed to 8). We report the training time of 100 epochs in Figure ~\ref{fig:runtime}. The growth of the runtime is roughly quadratic as the number of attributes increases. This is expected since the dependencies between one attribute and all the others yield quadratic complexity. When the number of attributes is excessively large, we can apply simple methods like computing the correlations between attributes to split the attributes into groups, where only the attributes within the same group exhibit correlations. Then, we can apply \NetworkName to learn the structure for each of the groups. \blue{We evaluate the effectiveness of this strategy on the Alarm dataset~\cite{alarm} which contains 36 attributes and 1000 records. The functional dependencies in Alarm is known. We split the attributes into three groups based on the functional dependencies. Each group contains 12 of them. We run Picket outlier detection on the three groups independently, and then aggregate the reconstruction loss across groups. We inject random and systematic noise of medium magnitude to 20\% of the records, and report the AUROC of outlier detection in Table~\ref{tab:alarm}. The results show that \ModelName provides high-quality outlier detection under the aforementioned strategy.}

\tabcolsep=0.08cm
\begin{table}
\center
\caption{ \blue{AUROC scores of outlier detection on the Alarm dataset. The attributes are split into three groups for Picket.}}
\label{tab:alarm}
\scriptsize
\begin{tabular}{c|cccc}
\hline
Noise Type &  IF & OCSVM & RVAE & \ModelName  \\ \hline \hline
Random  &   0.8848 &  0.8835 &  0.9357 & 0.9579   \\      
Systematic  & 0.7410 & 0.7283 & 0.7957 & 0.7967  \\\hline
\end{tabular}
\end{table}

We report the inference time overhead (runtime of \NetworkName loss computing and victim sample detectors) as long as the runtime of downstream prediction of each dataset in Table~\ref{tab:runtime_test}, when the data come in batches of 100. We can see that the overhead of PicketNet loss computing dominates the runtime, but it is still no more than a few seconds. As the downstream model becomes more complex, the relative overhead introduced by \ModelName would be reduced.

% \begin{table}[t]
% \caption{Training time of \NetworkName for each dataset.}
% \scriptsize
% \center
% \label{tab:runtime}
% \begin{tabular}{l|cccccc}
% \hline
% Dataset & Wine & Adult & Restaurant & Marketing & Titanic & HTRU2 \\
% \hline
% \thead{\scriptsize{Training}\\\scriptsize{Time (sec)}}   &1953 &8256 & 3794&4581 & 1693 & 189\\
% \hline
% \end{tabular}
% \end{table}

\begin{table}[t]
    \scriptsize
    \caption{Training time of \NetworkName for each dataset.}
    \vspace{7pt}
    \label{tab:runtime}
    \centering
    \begin{tabular}{lcccccc}
        \toprule
        Dataset & Wine & Adult & Restaurant & Marketing & Titanic & HTRU2 \\
        \midrule
            \thead{\scriptsize{Training}\\\scriptsize{Time (sec)}}   &1953 &8256 & 3794&4581 & 1693 & 189\\
        \bottomrule
    \end{tabular}
\end{table}

% \begin{table*}[t]
% \caption{Inference overhead of Picket and runtime of downstream prediction.}
% \scriptsize
% \center
% \label{tab:runtime_test}
% \begin{tabular}{l|cccccc}
% \hline
% \thead{\scriptsize{Dataset}} & Wine & Adult & Restaurant & Marketing & Titanic & HTRU2 \\
% \hline
% \thead{\scriptsize{\NetworkName Loss}\\\scriptsize{Computing (sec)}}   & 0.1512 & 2.5125 & 1.5345 & 2.0128& 1.5402 & 0.0316\\
% \hline
% \thead{\scriptsize{Victim Sample}\\ \scriptsize{Detectors (sec)}}   & 0.0006 & 0.0009 & 0.0012 & 0.0009& 0.0008 & 0.0004\\
% \hline
% \thead{\scriptsize{Downstream} \\\scriptsize{Prediction (LR) (sec)}}   & 0.0003 & 0.0010 & 0.0016 & 0.0011& 0.0006 & 0.0003\\
% \hline
% \thead{\scriptsize{Downstream} \\\scriptsize{Prediction (SVM) (sec)}}   & 0.0003 & 0.0021 & 0.0012 &0.0012 & 0.0006 & 0.0003\\
% \hline
% \thead{\scriptsize{Downstream} \\\scriptsize{Prediction (NN) (sec)}}   & 0.0004 & 0.0021 & 0.0030 &0.0014 & 0.0005 & 0.0004\\
% \hline
% \end{tabular}
% \end{table*}

\begin{table*}
    \small
    \caption{\blue{Inference overhead of Picket and runtime of downstream prediction.}}
    \vspace{7pt}
    \label{tab:runtime_test}
    \centering
    \begin{tabular}{l|cccccc}
        \toprule
        \thead{Dataset} & Wine & Adult & Restaurant & Marketing & Titanic & HTRU2 \\
        \midrule
            \thead{\NetworkName Loss\\Computing (sec)}   & 0.1512 & 0.4303 & 0.3003 & 0.4048 & 0.2892 & 0.0316\\
            \hline
            \thead{Victim Sample\\ Detectors (sec)}   & 0.0006 & 0.0009 & 0.0012 & 0.0009& 0.0008 & 0.0004\\
            \hline
            \thead{Downstream \\Prediction (LR) (sec)}   & 0.0003 & 0.0010 & 0.0016 & 0.0011& 0.0006 & 0.0003\\
            \hline
            \thead{Downstream \\Prediction (SVM) (sec)}   & 0.0003 & 0.0021 & 0.0012 &0.0012 & 0.0006 & 0.0003\\
            \hline
            \thead{Downstream \\Prediction (NN) (sec)}   & 0.0004 & 0.0021 & 0.0030 &0.0014 & 0.0005 & 0.0004\\
        \bottomrule
    \end{tabular}
\end{table*}

\section{Related Work}
\newparagraph{Data Validation Systems for ML}
TFX ~\cite{TFX,breck2019data} and Deequ ~\cite{Deequ} propose data validation modules that rely on user-defined constraints and simple anomaly detection. CleanML~\cite{cleanml} studies how the quality of training data affects the performance of downstream models. These works focus on simple constraints such as data types, value ranges, and one-column statistics and ignore the structure of the data. NaCL ~\cite{ijcai2019-377} and CPClean ~\cite{karla2020nearest} propose algorithms to deal with missing entries, and the effect of missing entries are analyzed theoretically in~\cite{liu2020robust}. These works are orthogonal to ours since they only consider missing entries.

\newparagraph{Learning Dependencies with Attention Mechanisms}
Attention mechanisms have been widely used in the field of natural language processing to learn the dependencies between tokens~\cite{attention,xlnet}. Recently, AimNet~\cite{AimNet} demonstrates that attention mechanisms are also effective in learning the dependencies between attributes in structured tabular data. AimNet employs the attention techniques to impute the missing values in tabular data and achieve state-of-the-art performance. AimNet is rather simplistic and it only captures schema-level dependencies. Furthermore, AimNet requires clean training data and does not employ any robust-training mechanism to tolerate noise.

\newparagraph{Outlier Detection Methods} Outlier detection for tabular data has been studied for years, and many rule-based methods have been proposed ~\cite{rulebased1,rulebased2,rulebased3}. Learning-based outlier detection has become popular recently and focuses on semi-supervised or unsupervised approaches. Semi-supervised methods such as the ones proposed in ~\cite{holodetect,supervised1,10.1145/3299869.3324956} still need human in the loop to explicitly label some data. Isolation Forest~\cite{isolation} and One-Class SVM~\cite{OCSVM} are simple unsupervised methods that are widely used. Autoencoder-based outlier detection methods~\cite{autoencoderOD1,autoencoderOD2,RobustVAE} are most relevant to our work since they also rely on the reconstruction of the input, and among them RVAE ~\cite{RobustVAE} works best for mixed-type tabular data. 

\newparagraph{Adversarial Attacks and Defenses} Training time attacks ~\cite{strongerPoison,gradientPoison,SVMpoisoning} add poisoned samples to corrupt the target model. Filtering-based defenses ~\cite{certifiedDefense,sever} remove suspicious samples during training based on training statistics. Inference time attacks ~\cite{PGDAttack,CWAttack,deepfool} add small perturbation to test samples to fool the classifier. Efforts have been made to improve the robustness of the model by training data augmentation ~\cite{aug1,aug2} or making modifications to the model ~\cite{kWinner,rethinking,ensembleDiv}. Those works focus on robustness from the model perspective without assessment of data quality. Hence, they are orthogonal to ours. Another group of defenses trying to detect adversarial samples at inference time are more directly related to our work. Roth et al. ~\cite{odds} and Hu et al. ~\cite{weaknessToS} add random noise to input samples and detect suspicious ones based on the changes in the logit values. Grosse et al. ~\cite{MWOC} assume that adversarial samples have different distributions from benign samples and add another class to the classifier to detect them.

\section{Conclusion}\label{sec:conclusion}
We introduced \ModelName, a first-of-its-kind system that safeguards against data corruptions for machine learning pipelines over tabular data either during training or deployment.
%For the training stage, \ModelName identifies erroneous training examples that if used for learning will lead to a biased model, while for the deployment stage, \ModelName flags corrupted query points to a trained machine learning model that due to noise will result in incorrect predictions.
To design \ModelName, we introduced \NetworkName, a novel self-supervised deep learning model that corresponds to a Transformer network for tabular data. \ModelName is designed as a plugin that can increase the robustness of any machine learning pipeline.

\bibliographystyle{ACM-Reference-Format}
\bibliography{main}

\clearpage
\appendix
\section{Appendix}

\subsection{Hyper-parameters of \NetworkName}\label{sec:hyperparameters}
\NetworkName is not sensitive to hyper-parameters in most cases. The default hyper-parameters we use in the experiments is shown in Table ~\ref{tab:hyper}. For purely numerical datasets, we reduce the dimension of $I_i^{(l)}$ and $P_i^{(l)}$ to 8, and for HTRU2, we reduce the number of self-attention layers to 1. In the other datasets, we always use the default hyper-parameters. We use the Adam optimizer ~\cite{adam} with $\beta_1 = 0.9$, $\beta_2 = 0.98$ and $\epsilon = 10^{-9}$ for training. The learning rate $lr = 0.5 d^{-0.5} \min(s^{-0.5}, 300^{-1.5} s)$, where $d$ is the dimension of $I_i^{(l)}$ and $P_i^{(l)}$, $s$ is the index of the training step. $lr$ increases in the first few steps and then decreases. Typically, \NetworkName takes 100 to 500 epochs to converge, depending on the datasets. 

% \begin{table}[]
% \scriptsize
% \center
% \caption{Default hyper-parameters for \NetworkName.}
% \label{tab:hyper}
% \begin{tabular}{l|c}
% \hline
% Hyper-Parameter & Value \\ \hline\hline
% Number of Self-Attention Layers & 6  \\ \hline
% Number of Attention Heads  &  2  \\ \hline
% Dimension of $I_i^{(l)}$ and $P_i^{(l)}$  &  64 \\ \hline
% Number of Hidden Layers in Each Feedforward Network & 1 \\ \hline
% Dimension of the Hidden Layers in Feedforward Networks & 64 \\ \hline
% Dropout & 0.1 \\ \hline
% Size of the Negative Sample Set $Z_i$ & 4 \\ \hline
% Warm-up Epochs $E_1$ for Loss-Based Filtering & 50 \\ \hline
% Loss Recording Epochs $E_2$ & 20\\ \hline
% \end{tabular}
% \end{table}

\begin{table*}
    \small
    \caption{Default hyper-parameters for \NetworkName.}
    \vspace{7pt}
    \label{tab:hyper}
    \centering
    \begin{tabular}{l c}
        \toprule
        Hyper-Parameter & Value \\
        \midrule
            Number of Self-Attention Layers & 6  \\
            Number of Attention Heads  &  2  \\
            Dimension of $I_i^{(l)}$ and $P_i^{(l)}$  &  64 \\
            Number of Hidden Layers in Each Feedforward Network & 1 \\
            Dimension of the Hidden Layers in Feedforward Networks & 64 \\
            Dropout & 0.1 \\
            Size of the Negative Sample Set $Z_i$ & 4 \\
            Warm-up Epochs $E_1$ for Loss-Based Filtering & 50 \\
            Loss Recording Epochs $E_2$ & 20\\
        \bottomrule
    \end{tabular}
\end{table*}

\subsection{Outlier Detection on Synthetic Data}\label{sec:synthetic_data}
\blue{We evaluate the performance of outlier detection on synthetic datasets to understand the effects of several aspects about the data and noise, including the strength of structure, data dimension, noise level and magnitude of extreme outliers. Here the term structure means dependencies or correlations between attributes.}

\blue{We generate synthetic datasets as follows. Each synthetic data point $x = [x_1, x_2, \dots, x_T]^T$ is generated by $x=Az$, where $z \in \mathbb{R}^R$ and $A \in \mathbb{R}^{T \times R}$. Each entry of $z$ is sampled from the standard Gaussian distribution, and each entry of $A$ is sampled uniformly from $-1$ to $1$. Unless otherwise specified, we inject random noise with $\beta=0.2$ and $\sigma_1=1$ to 20\% of the samples by default.}

\begin{figure}[ht]
\centering
\includegraphics[width=0.45\textwidth]{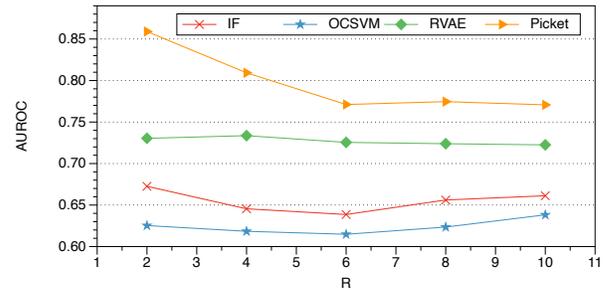}
\caption{Training time outlier detection over synthetic datasets that exhibit different strength of structure.}
\label{fig:structureStrength}
\end{figure}

\newparagraph{Effect of Structure}
\blue{By performing outlier detection over synthetic datasets that exhibit different strength of structure, we show that the advantage of Picket over the other outlier detection methods is its ability to capture the structure of the data. We fix $T=10$ and vary $R$ to change the strength of structure. Smaller $R$ indicates stronger structure and more redundancy across attributes.  The results are shown in Figure~\ref{fig:structureStrength}. Picket performs better when the structure is strong, while the performance of the other methods is not affected by the strength of structure, which indicates that Picket is able to capture the structure of the data and benefit from it.}

\begin{figure}[ht]
\centering
\includegraphics[width=0.45\textwidth]{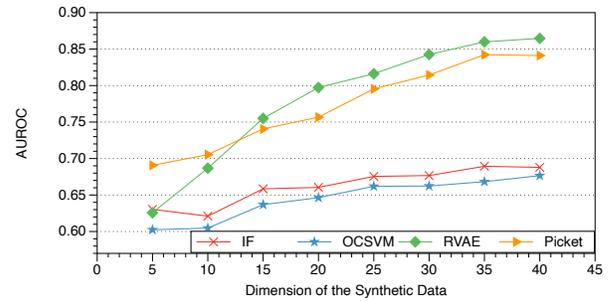}
\caption{\blue{Training time outlier detection over synthetic datasets that have different dimensions.}}
\label{fig:dimSynthetic}
\end{figure}

\newparagraph{Effect of Data Dimension}
\blue{We vary the the data dimension $T$ to study how it affects the performance. The hidden dimension $R$ is set to $T$ so that the attributes are independent. The results are shown in Figure~\ref{fig:dimSynthetic}. The performance of all methods increases as the data dimension gets larger. The reason is that there are more corrupted cells in corrupted samples when the dimension increases, making them easier to be detected. Note that RVAE performs quite well in this setting, which is not surprising since it is built exactly on the assumption that the data come from Gaussian distributions.}

\begin{figure}[ht]
\centering
\includegraphics[width=0.45\textwidth]{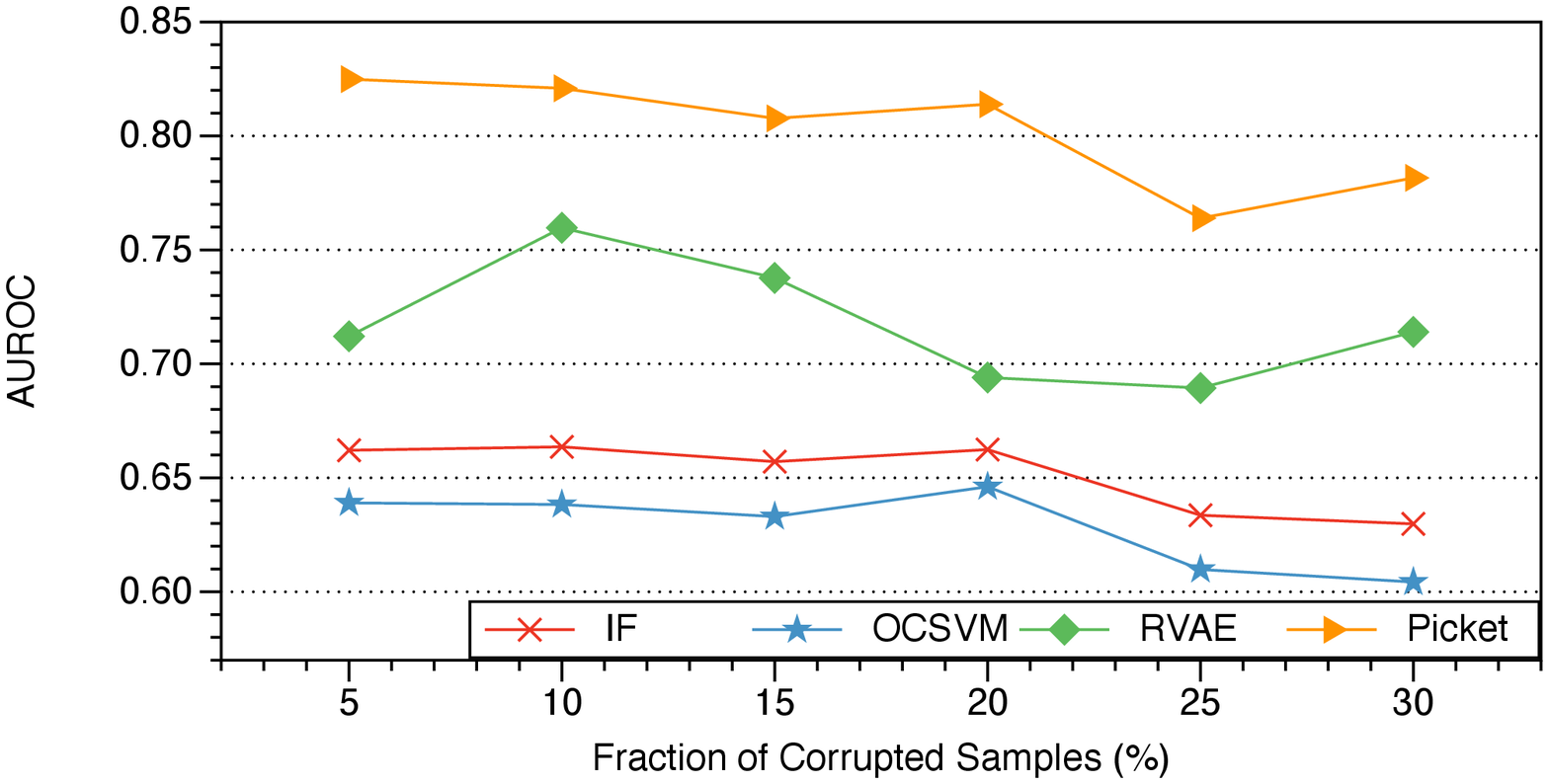}
\caption{\blue{Training time outlier detection over synthetic datasets under different fractions of corrupted samples.}}
\label{fig:RowEpsSynthetic}
\end{figure}

\begin{figure}[ht]
\centering
\includegraphics[width=0.45\textwidth]{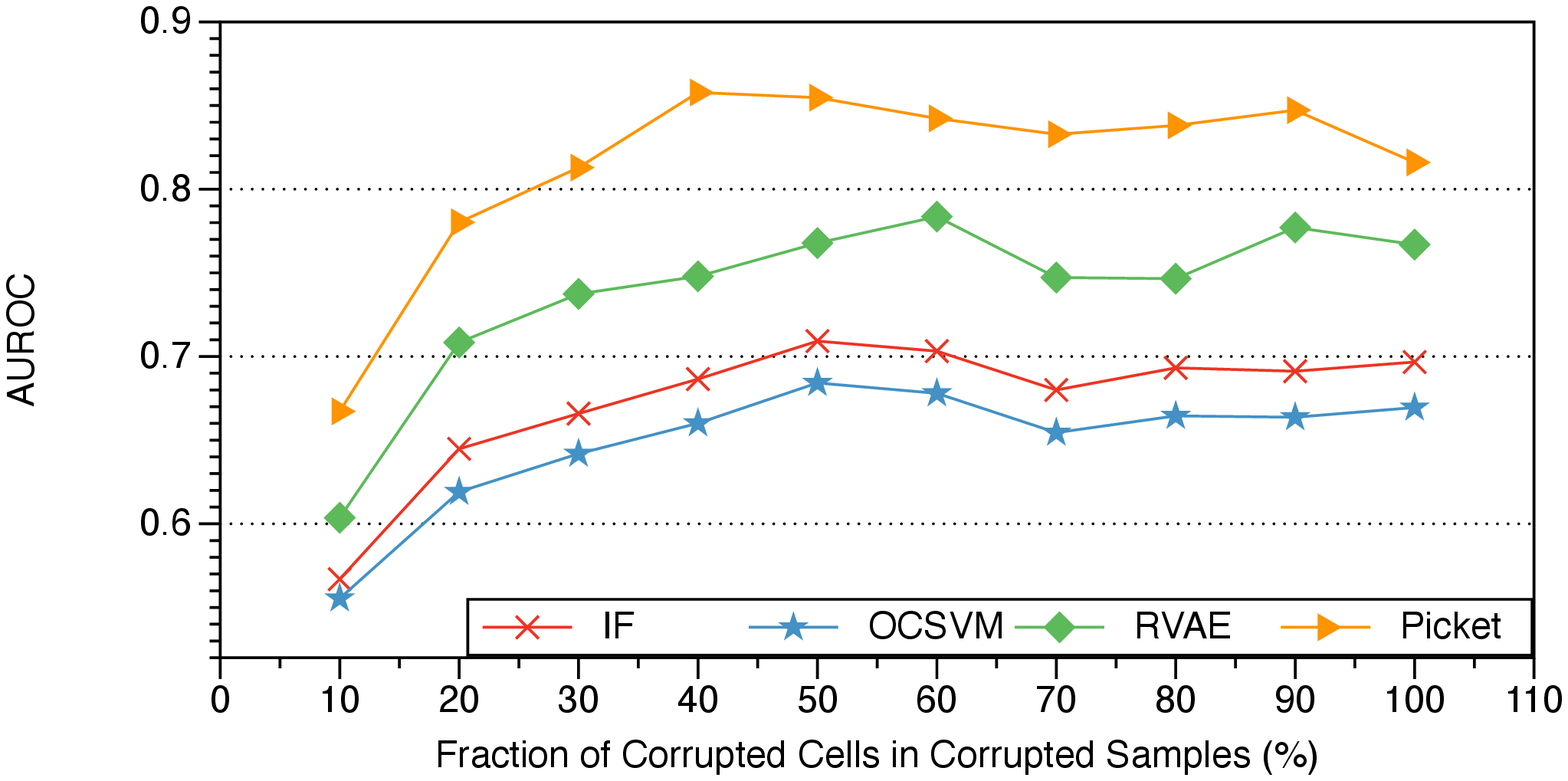}
\caption{\blue{Training time outlier detection over synthetic datasets under different fractions of corrupted cells in corrupted samples.}}
\label{fig:ColEpsSynthetic}
\end{figure}

\begin{figure}[ht]
\centering
\includegraphics[width=0.45\textwidth]{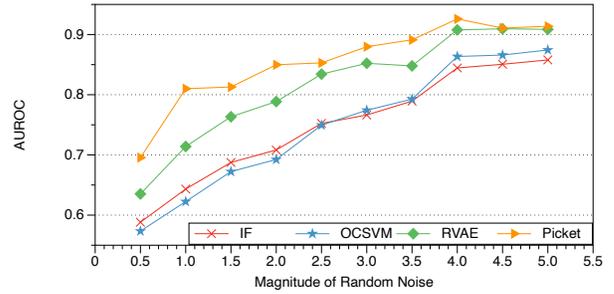}
\caption{\blue{Training time outlier detection over synthetic datasets under different noise magnitudes.}}
\label{fig:MagnitudeSynthetic}
\end{figure}

\newparagraph{Effect of Noise Level}
\blue{We study the effect of noise level, including the fraction of corrupted samples, the fraction of corrupted cells in corrupted samples ($\beta$) and the magnitude of the random noise ($\sigma_1$). Each time we vary one of the factors and fix the others. The data dimension $T$ is fixed to $10$, and $R$ is fixed to $5$. As is shown in Figure~\ref{fig:RowEpsSynthetic}, when we vary the fraction of corrupted samples, the performance of all methods keeps stable. Figure~\ref{fig:ColEpsSynthetic} and~\ref{fig:MagnitudeSynthetic} show that the performance of all methods increases as we increase the fraction of corrupted cells in corrupted samples or the magnitude of the random noise. These results show that the corruption level of the corrupted samples have a more significant effect on the outlier detection performance than the fraction of corrupted samples. }

\begin{figure}[ht]
\centering
\includegraphics[width=0.45\textwidth]{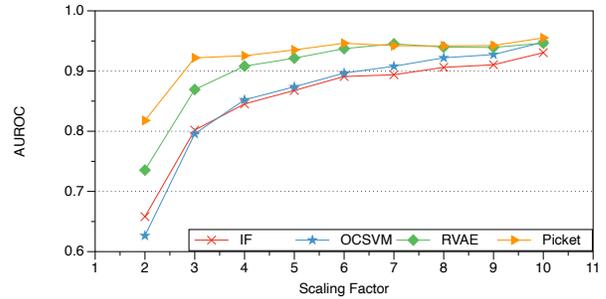}
\caption{\blue{Training time outlier detection over synthetic datasets under different levels of extreme values.}}
\label{fig:ExtremeEpsSynthetic}
\end{figure}

\newparagraph{Effect of Extreme Outliers}
\blue{We study how the models behave under extreme outliers with different magnitude. We corrupt 20\% of the samples, and among those samples 20\% of the cells are multiplied by a scaling factor. We vary the value of the scaling factor and report the detection performance in Figure~\ref{fig:ExtremeEpsSynthetic}. As the scaling factor gets larger, the performance of all methods increases. This is expected since more extreme values deviate more from the normal distribution.}

\subsection{Outlier Detection with Cross-Validation}\label{sec:cross_validation}
\blue{We evaluate the ability to detect outliers for unseen data during training using cross-validation. We use 5 iterations of 2-fold cross-validation with a modified t-test recommended by Dietterich and Thomas~\cite{dietterich1998approximate}. Specifically, in each iteration, we randomly split the dataset into two folds. Then we use one fold to train the outlier detection models, and the other to validate their performance. The results are reported in Table~\ref{tab:cross_validation}. The results shows that Picket achieves the best performance among the examined methods on all dataset-noise combinations for unseen data at training time. In some cases, Picket is significantly better than all competing methods.}

% Cross-Validation
\tabcolsep=0.08cm
\begin{table}
\center
\caption{\blue{AUROC of outlier detection from cross-validation. The numbers are made bold when the corresponding method is significantly better (p value less than 0.05) than all the others.}}
\label{tab:cross_validation}
\scriptsize
\begin{tabular}{c|c|cccc}
\hline
\multicolumn{1}{c|}{Dataset} & \multicolumn{1}{c|}{\thead{\scriptsize{Noise Type}}} & IF & OCSVM & RVAE & \ModelName\\ \hline \hline
\multirow{5}{*}{Wine}           & Random-Medium & 0.8876 & 0.8886 & 0.9170 & 0.9252\\
                                & Systematic-Medium & 0.9537 & 0.8971 & 0.9123 & 0.9669 \\
                                & Poison-LR & 0.9756 & 0.9054 & 0.9061 & 0.9781   \\                  
                                & Poison-SVM   & 0.9761 & 0.9047 & 0.9025 & 0.9787  \\ 
                              & Poison-NN  & 0.9877 & 0.8696 & 0.9356 & 0.9921 \\ \hline
\multirow{2}{*}{Adult}            & Random-Medium & 0.7800 & 0.8260 & 0.9019 & \textbf{0.9240}\\
                                & Systematic-Medium & 0.8048 & 0.8217 & 0.8530 & \textbf{0.9180} \\ \hline
\multirow{3}{*}{Restaurant}            & Random-Medium & 0.4814 & 0.4431 & 0.6985 & \textbf{0.9281} \\
                                & Systematic-Medium & 0.4805 & 0.4449 & 0.6596 & \textbf{0.8778} \\
                                & Real* & 0.5514 & 0.5116 & 0.4558 & \textbf{0.8978} \\\hline
\multirow{2}{*}{Marketing}            & Random-Medium & 0.7539 & 0.7804 & 0.8688 & 0.8646\\
                                & Systematic-Medium & 0.6746 & 0.6632 & 0.7787 & 0.7810\\ \hline
\multirow{3}{*}{Titanic}            & Random-Medium & 0.6014 & 0.6933 & 0.5819 & \textbf{0.7709} \\
                                & Systematic-Medium & 0.5811 & 0.7037 & 0.5557 & 0.7691 \\
                                & Real & 0.5851 & 0.6472 & 0.5000 & 0.7314 \\\hline
\multirow{1}{*}{Food}            & Real & 0.5094 & 0.5210 & 0.5180 & 0.5506\\\hline
\end{tabular}
\begin{tablenotes}
      \item *Real is short for common errors in the real world. 
\end{tablenotes}
\end{table}

\subsection{Performance of Training Time Outlier Detection under Different Fraction of Corrupted Samples}\label{sec:train_time_row_fraction}
\begin{figure*}[t]
    \centering
    \subfigure[Wine under Poisoning Attack]{\includegraphics[width=0.45\textwidth]{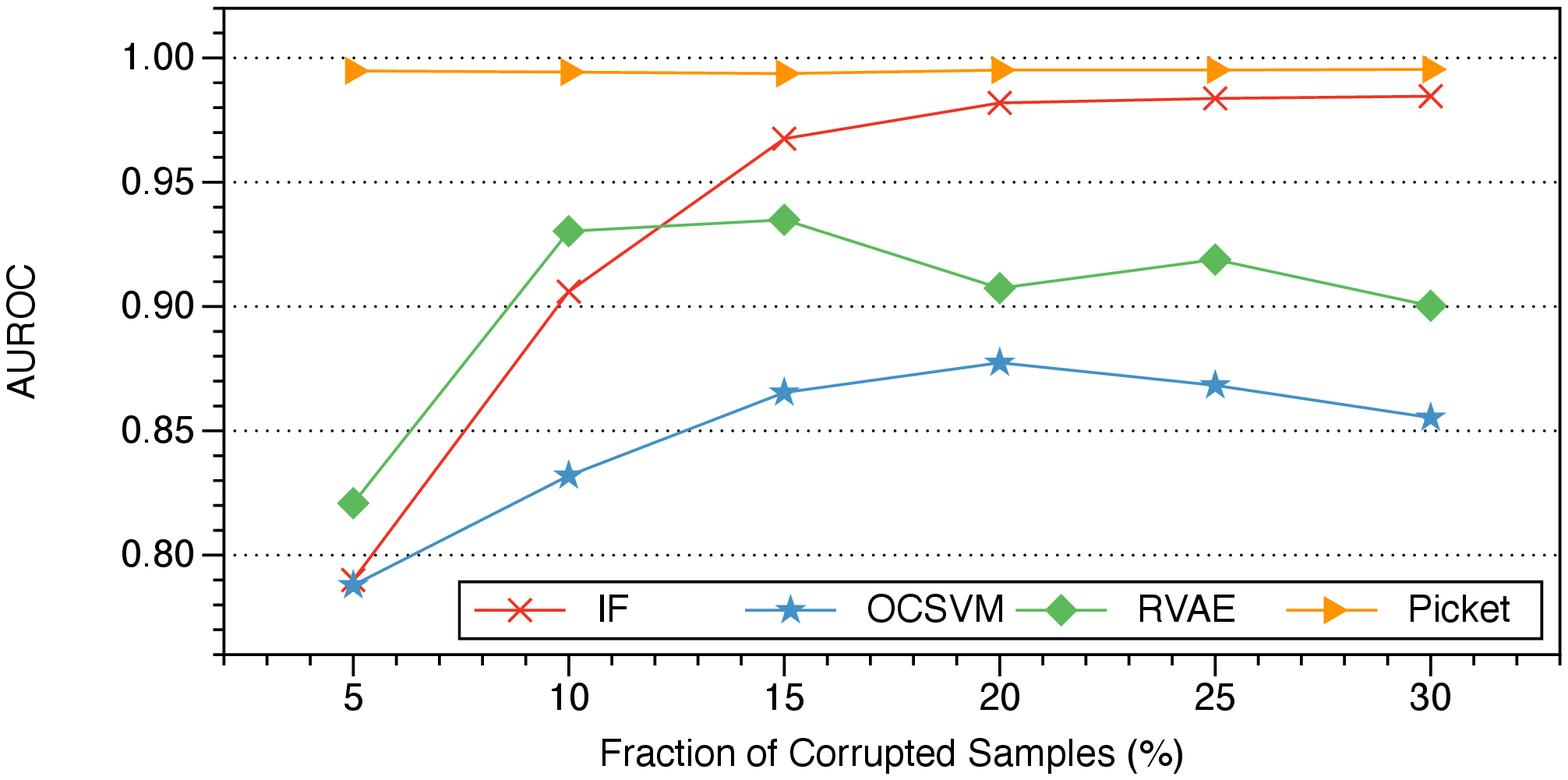}}
    \subfigure[Adult under Systematic Noise (Medium)]{\includegraphics[width=0.45\textwidth]{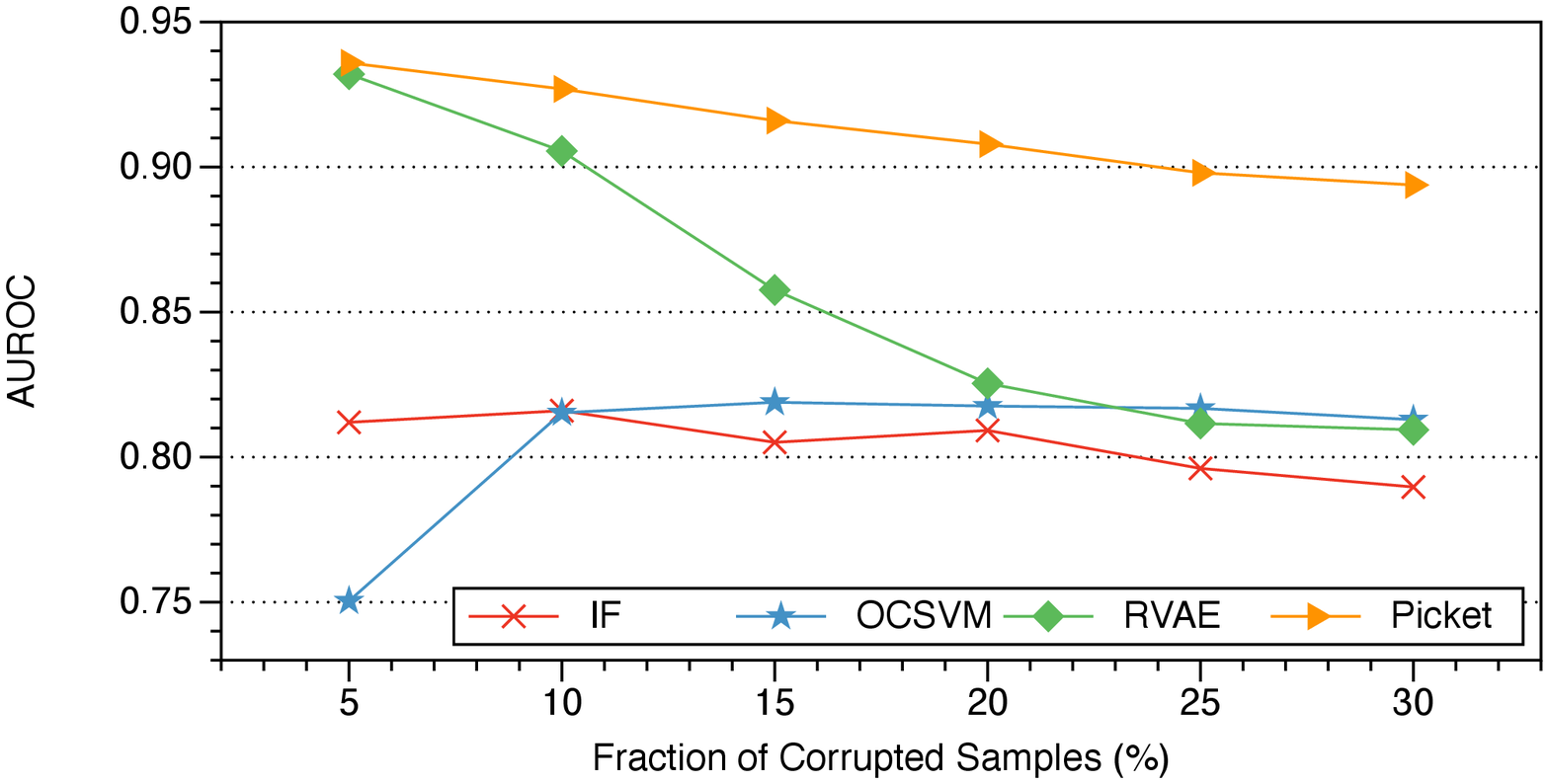}}
    \subfigure[Marketing under Random Noise (Medium)]{\includegraphics[width=0.45\textwidth]{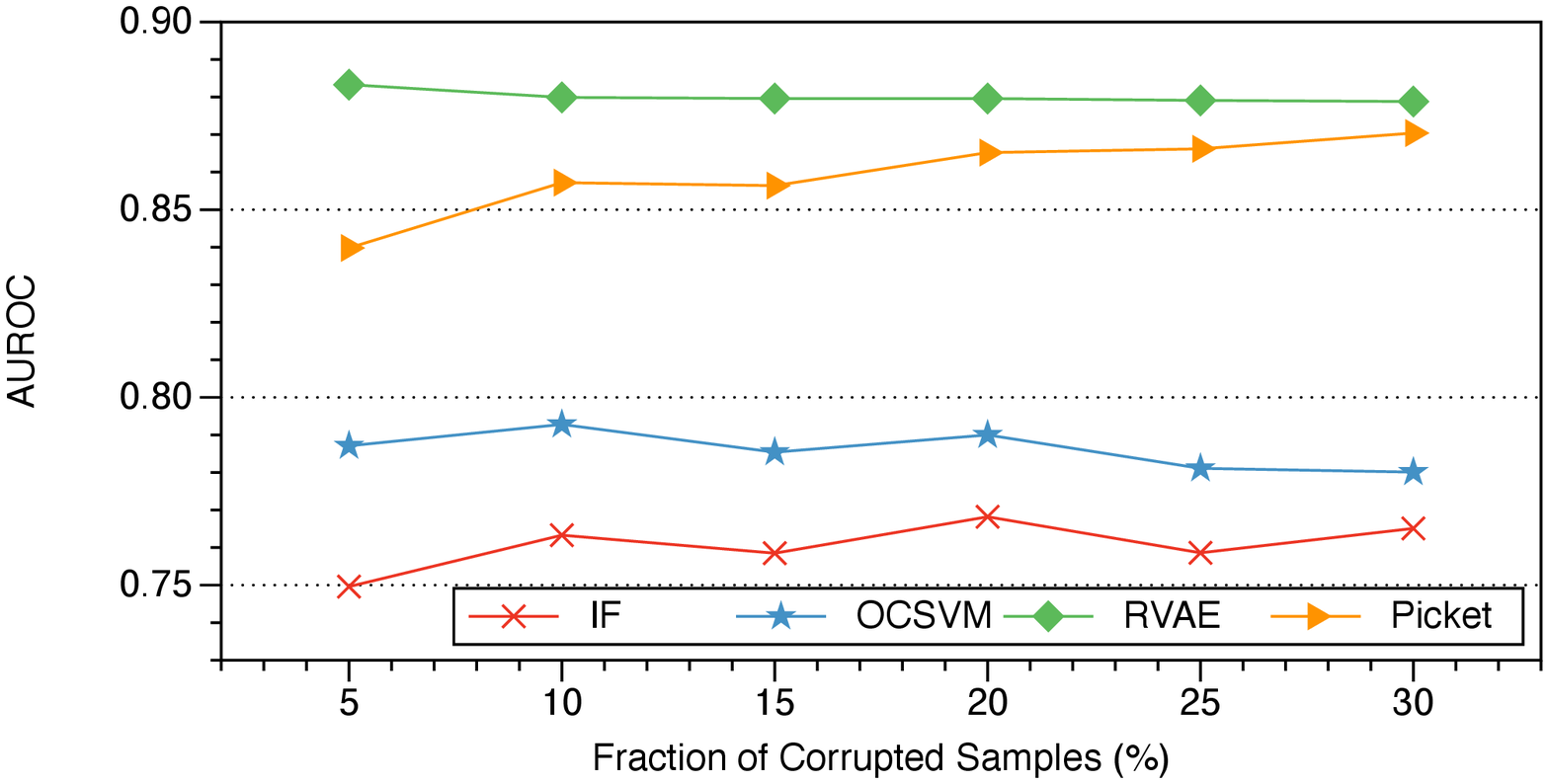}}
    \caption{\blue{AUROC of outlier detection under different fractions of corrupted samples.}}
    \label{fig:AUROC_row_eps}
    \vspace{-10pt}
\end{figure*}

\blue{We vary the fraction of corrupted samples, and report the corresponding AUROC of training time outlier detection in Figure~\ref{fig:AUROC_row_eps}. The datasets we use are Wine with poisoning attack on NN, Adult with systematic noise, and Marketing with random noise. The random and systematic noise is in the Medium level.}

\blue{From the results we can see that the detection performance could either increase or decrease as the fraction of corrupted samples grows, depending on the type of noise and detection method. One one hand, the outliers are easier to detect when they form a larger cluster; one the other hand, more corrupted samples may mislead the learning of the clean distribution. Nevertheless, Picket keeps a relatively consistent performance with either large or small fraction of corrupted samples, while other methods may have a large gap when the fraction varies.}

\subsection{Performance of Training Time Outlier Detection under Low/High Level Random/Systematic Noise}
\label{sec:train_time_low_high}

\begin{figure}[htbp]
    \centering
	\includegraphics[width=0.35\textwidth]{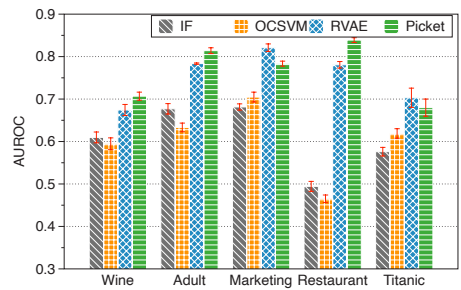}
    \caption{AUROC of outlier detection for random noise (Low level). The error bars represent the standard errors.}
    \label{fig:AUROC_RS_rand_low}
\end{figure}

\begin{figure}[htbp]
    \centering
	\includegraphics[width=0.35\textwidth]{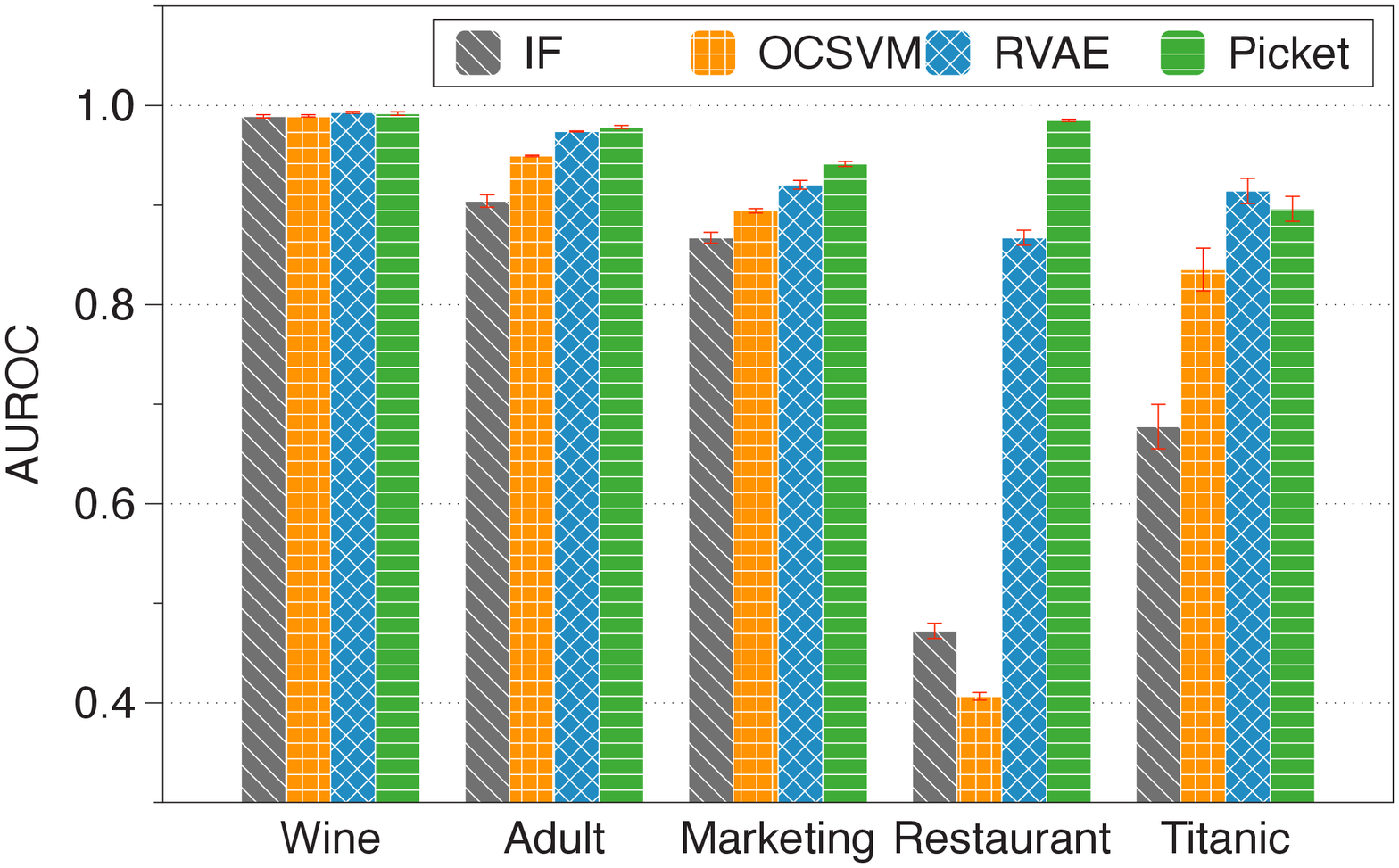}
    \caption{AUROC of outlier detection for random noise (High level). The error bars represent the standard errors.}
    \label{fig:AUROC_RS_rand_high}
\end{figure}

\begin{figure}[htbp]
    \centering
	\includegraphics[width=0.35\textwidth]{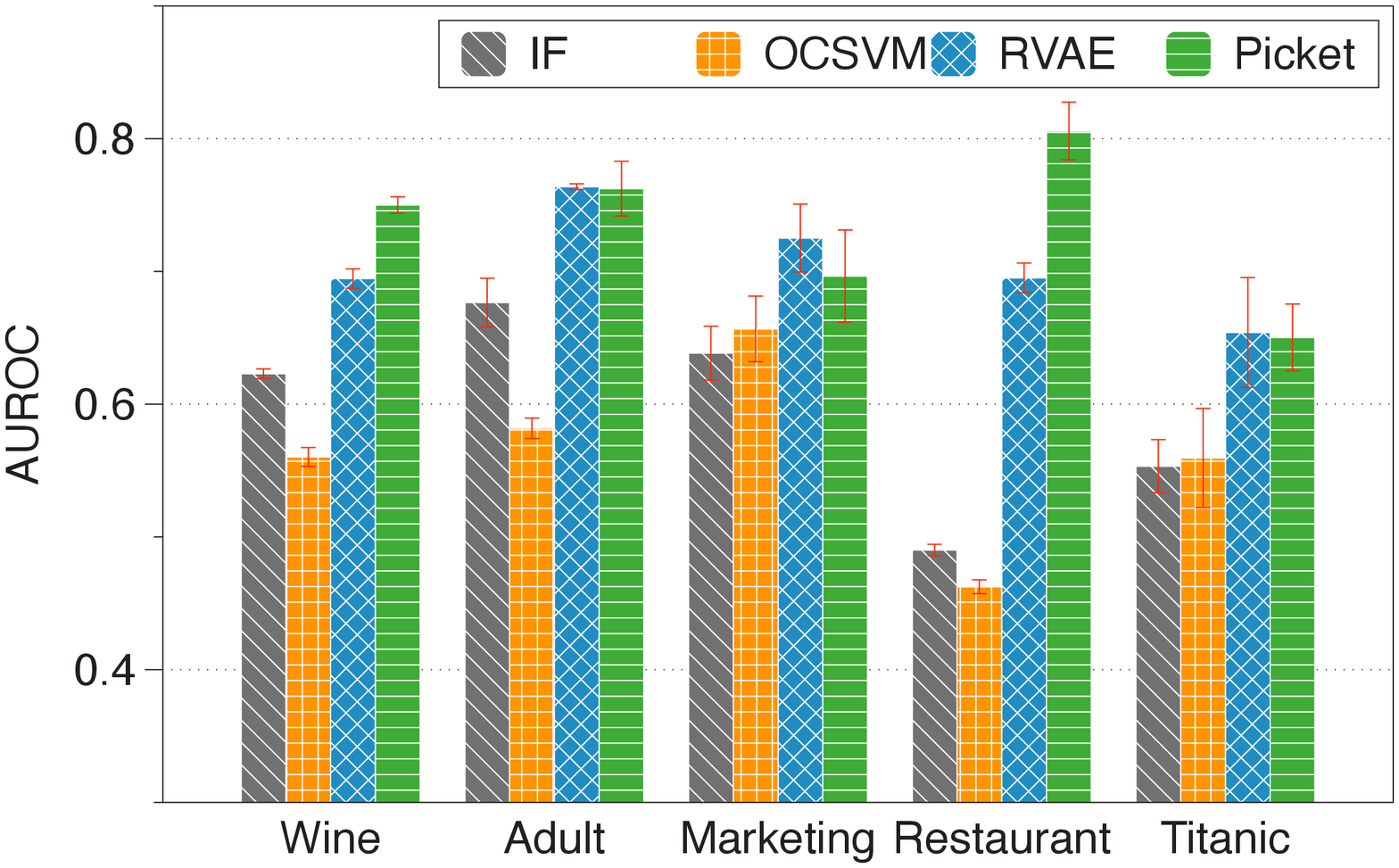}
    \caption{AUROC of outlier detection for systematic noise (Low level). The error bars represent the standard errors.}
    \label{fig:AUROC_RS_system_low}
\end{figure}

\begin{figure}[htbp]
    \centering
	\includegraphics[width=0.35\textwidth]{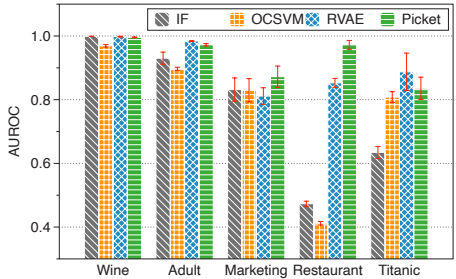}
    \caption{AUROC of outlier detection for systematic noise (High level). The error bars represent the standard errors.}
    \label{fig:AUROC_RS_system_high}
\end{figure}

We depict the AUROC of training time outlier detection under low/high level random/systematic noise in Figure~\ref{fig:AUROC_RS_rand_low}, ~\ref{fig:AUROC_RS_rand_high}, ~\ref{fig:AUROC_RS_system_low}, ~\ref{fig:AUROC_RS_system_high}, when 20\% of the samples are corrupted. The observation is quite similar to the case of medium level noise. The performance of \ModelName is quite good and consistent across different datasets and noise settings. RF and OCSVM perform poorly on the datasets that contain textual attributes. RVAE is competitive in some cases but fails in the others. Note that low level noise is much harder to detect than high level noise. The reason is that samples with high level noise tend to deviate a lot from the clean distribution, while samples with low level noise look quite similar to the clean ones and may not be detectable in some cases. However, low level noise will not affect the downstream model as much as high level noise, unless it is adversarially crafted.     

\subsection{Accuracy of Downstream Models under Random/Systematic Noise with Different Filtering Methods}
\label{sec:downstream_acc_random_system}
We also study how the accuracy of the downstream models changes when we apply different filtering methods under random and systematic noise. We first focus on random noise. The results are shown in Tables~\ref{tab:random_ds_acc_low}, ~\ref{tab:random_ds_acc},~\ref{tab:random_ds_acc_high}.
As expected, in the presence of random noise, the performance of the downstream models drops in some cases and remains roughly the same in the other cases if we look at CL and NF. In the cases when the downstream accuracy drops, we can see that filtering helps most of the time. 

% To better understand the performance of different methods, we consider the average changes of accuracy compared to the two reference points across all datasets and models. The aggregated result is shown in Table ~\ref{tab:aggregated_acc}. 
% Among all the methods, our model has the most positive effect on the downstream accuracy since the improvement over doing nothing is the largest, and the accuracy is closest to clean data. This is expected since our method is the most powerful in detecting outliers. 

\begin{figure}
\centering
\includegraphics[width=0.45\textwidth]{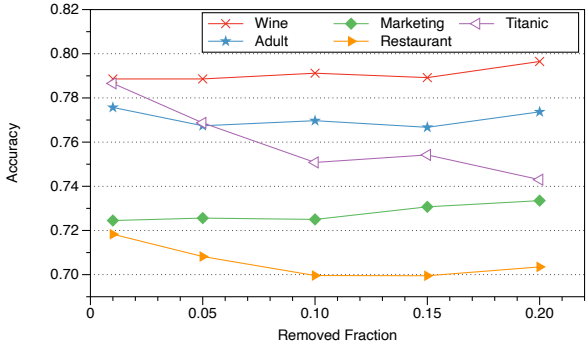}
\caption{Changes in test accuracy of a neural network when filtering different fraction of the points; random noise (Medium level).}
\label{fig:fractionAcc}
\end{figure}

If we compare the performance of \ModelName and NF in Table~\ref{tab:random_ds_acc} for Neural Networks, we see that for Adult, Titanic, and Restaurant \ModelName exhibits slightly worse test accuracy. These results are attributed to the selected thresholds for filtering in \ModelName (see Section~\ref{sec:robust_ml}). In Figure~\ref{fig:fractionAcc}, we show the test accuracy of the downstream neural network for different levels of the \ModelName threshold. We can see that for some datasets, random noise serves as regularization and improves the performance of the downstream model. Therefore, we need to tune the threshold to achieve the best performance.

% Random Low
\tabcolsep=0.08cm
\begin{table}
\center
\caption{Test accuracy of downstream models under random noise (Low level) and different filtering methods.}
\label{tab:random_ds_acc_low}
\scriptsize
\begin{tabular}{c|c|cccc|cc}
\hline
\multicolumn{1}{c|}{Dataset} & \multicolumn{1}{c|}{\thead{\scriptsize{DM*}}} & IF & OCSVM & RVAE & \ModelName & CL & NF\\ \hline \hline
\multirow{3}{*}{Wine}            & LR  & 0.7429 & \textbf{0.7435} & 0.7427 & 0.7429 & 0.7457 & 0.7443   \\                  
                                & SVM  & 0.7447 & 0.7437 & \textbf{0.7486} & 0.7465 & 0.7465 & 0.7453 \\ 
                              & NN  & 0.7857 & 0.7800 & 0.7849 & \textbf{0.7941} & 0.8051 & 0.7922  \\ \hline
\multirow{3}{*}{Adult}            & LR & 0.8207 & 0.8211 & 0.8127 & \textbf{0.8233} & 0.8240 & 0.8190  \\ 
                              & SVM  & 0.8181 & 0.8196 & 0.8075 & \textbf{0.8212} & 0.8238 & 0.8187   \\ 
                              & NN  & \textbf{0.7818} & 0.7800 & 0.7803 & 0.7816 & 0.7909 & 0.7836  \\ \hline
\multirow{3}{*}{Restaurant}            & LR  & 0.7318 & 0.7347 & \textbf{0.7361} & 0.7352 & 0.7375 & 0.7378 \\  
                              & SVM & 0.6922 & 0.7078 & \textbf{0.7123} & 0.6972 & 0.7116 & 0.7060\\ 
                              & NN  & 0.7128 & 0.6982 & 0.7099 & \textbf{0.7135} & 0.7306 & 0.7182 \\ \hline
\multirow{3}{*}{Marketing}            & LR  & 0.7622 & 0.7661 & 0.7642 & \textbf{0.7663} & 0.7672 & 0.7691\\  
                              & SVM & 0.7649 & 0.7668 & 0.7655 & \textbf{0.7678} & 0.7681 & 0.7708  \\ 
                              & NN & \textbf{0.7362} & 0.7282 & 0.7302 & 0.7265 & 0.7261 & 0.7300\\ \hline
\multirow{3}{*}{Titanic}            & LR & 0.7810 & 0.7777 & 0.7832 & \textbf{0.7844} & 0.7877 & 0.7821 \\  
                              & SVM  & 0.7799 & 0.7866 & 0.7788 & \textbf{0.7877} & 0.7888 & 0.7888 \\ 
                              & NN & \textbf{0.7654} & 0.7542 & 0.7531 & \textbf{0.7654} & 0.7743 & 0.7709 \\ \hline
\end{tabular}
\begin{tablenotes}
      \item *DM = Downstream Model.
\end{tablenotes}
\end{table}

\tabcolsep=0.08cm
\begin{table}
\center
\caption{Test accuracy of downstream models under random noise (Medium level) and different filtering methods.}
\label{tab:random_ds_acc}
\scriptsize
\begin{tabular}{c|c|cccc|cc}
\hline
\multicolumn{1}{c|}{Dataset} & \multicolumn{1}{c|}{\thead{\scriptsize{DM*}}} & IF & OCSVM & RVAE & \ModelName & CL & NF\\ \hline \hline
\multirow{3}{*}{Wine}            & LR  & \textbf{0.7410} & 0.7396 & \textbf{0.7410} & 0.7398 & 0.7457 & 0.7280   \\                  
                                & SVM  & 0.7441 & \textbf{0.7457} & 0.7443 & 0.7431 & 0.7467 & 0.7259 \\ 
                              & NN  & 0.7743 & 0.7776 & \textbf{0.7816} & 0.7776 & 0.7973 & 0.7761  \\ \hline
\multirow{3}{*}{Adult}            & LR & 0.8140 & 0.8220 & \textbf{0.8233} & 0.8224 & 0.8240 & 0.8111  \\ 
                              & SVM  & 0.8109 & 0.8200 & \textbf{0.8219} & 0.8207 & 0.8238 & 0.8082   \\ 
                              & NN  & \textbf{0.7856} & 0.7795 & 0.7830 & 0.7850 & 0.7934 & 0.7883  \\ \hline
\multirow{3}{*}{Restaurant}            & LR  & 0.7342 & 0.7321 & 0.7313 & \textbf{0.7366} & 0.7375 & 0.7349 \\  
                              & SVM & \textbf{0.7111} & 0.7083 & 0.6898 & 0.6858 & 0.7185 & 0.6872\\ 
                              & NN & 0.7059 & 0.7064 & 0.7062 & \textbf{0.7157} & 0.7298 & 0.7210 \\ \hline
\multirow{3}{*}{Marketing}            & LR & 0.7645 & 0.7624 & 0.7642 & \textbf{0.7656} & 0.7672 & 0.7665 \\  
                              & SVM & 0.7654 & 0.7639 & 0.7654 & \textbf{0.7665} & 0.7681 & 0.7669  \\ 
                              & NN & 0.7267 & \textbf{0.7360} & 0.7301 & 0.7344 & 0.7311 & 0.7310\\ \hline
\multirow{3}{*}{Titanic}            & LR & 0.7799 & 0.7821 & 0.7777 & \textbf{0.7877} & 0.7877 & 0.7754 \\  
                              & SVM & 0.7810 & 0.7765 & 0.7788 & \textbf{0.7933} & 0.7888 & 0.7821  \\ 
                              & NN & 0.7575 & 0.7665 & 0.7408 & \textbf{0.7765} & 0.7944 & 0.7844\\ \hline
\end{tabular}
\begin{tablenotes}
      \item *DM = Downstream Model.
\end{tablenotes}
\end{table}

% Random High
\tabcolsep=0.08cm
\begin{table}
\center
\caption{Test accuracy of downstream models under random noise (High level) and different filtering methods.}
\label{tab:random_ds_acc_high}
\scriptsize
\begin{tabular}{c|c|cccc|cc}
\hline
\multicolumn{1}{c|}{Dataset} & \multicolumn{1}{c|}{\thead{\scriptsize{DM*}}} & IF & OCSVM & RVAE & \ModelName & CL & NF\\ \hline \hline
\multirow{3}{*}{Wine}            & LR  & 0.7410 & 0.7406 & 0.7398 & \textbf{0.7418} & 0.7457 & 0.6861   \\                  
                                & SVM & 0.7441 & 0.7414 & 0.7427 & \textbf{0.7453} & 0.7469 & 0.6806    \\ 
                              & NN  & 0.7865 & 0.7839 & \textbf{0.7896} & 0.7806 & 0.7941 & 0.7780  \\ \hline
\multirow{3}{*}{Adult}            & LR & 0.8047 & 0.8196 & 0.8218 & \textbf{0.8224} & 0.8240 & 0.8002  \\ 
                              & SVM  & 0.8024 & 0.8196 & \textbf{0.8207} & 0.8205 & 0.8238 & 0.7971   \\ 
                              & NN  & 0.7853 & 0.7763 & \textbf{0.7867} & 0.7861 & 0.7982 & 0.7863  \\ \hline
\multirow{3}{*}{Restaurant}            & LR & \textbf{0.7380} & 0.7369 & 0.7335 & 0.7327 & 0.7375 & 0.7416  \\  
                              & SVM & \textbf{0.7161} & 0.7060 & 0.7154 & 0.7126 & 0.7053 & 0.6872\\ 
                              & NN  & 0.7147 & 0.7172 & 0.7155 & \textbf{0.7206} & 0.7251 & 0.7247\\ \hline
\multirow{3}{*}{Marketing}            & LR & 0.7653 & 0.7649 & 0.7641 & \textbf{0.7668} & 0.7672 & 0.7671 \\  
                              & SVM  & 0.7660 & 0.7660 & 0.7659 & \textbf{0.7699} & 0.7681 & 0.7686 \\ 
                              & NN & 0.7255 & 0.7265 & \textbf{0.7284} & 0.7271 & 0.7245 & 0.7295\\ \hline
\multirow{3}{*}{Titanic}            & LR & \textbf{0.7877} & 0.7777 & 0.7799 & 0.7799 & 0.7877 & 0.7877 \\  
                              & SVM & \textbf{0.7922} & 0.7810 & 0.7855 & 0.7799 & 0.7888 & 0.7844  \\ 
                              & NN & 0.7609 & 0.7687 & 0.7709 & \textbf{0.7765} & 0.7866 & 0.7832\\ \hline
\end{tabular}
\begin{tablenotes}
      \item *DM = Downstream Model.
\end{tablenotes}
\end{table}

% \begin{table}
% \scriptsize
% \center
% \caption{Average test time accuracy change after filtering with different methods against CL and NF for random noise (Medium level).}
% \label{tab:aggregated_acc}
% \begin{tabular}{c|cccc}
% \hline
% Reference Point & IF & OCSVM & RVAE & \ModelName \\ \hline\hline
% CL  & -0.0105 & -0.0092 & -0.0116 & -0.0069               \\ \hline
% NF & 0.0007 & 0.0019 & -0.0005 & 0.0042               \\ \hline
% \end{tabular}
% \end{table}

We then turn our attention to systematic noise. The results are shown in Table~\ref{tab:system_ds_acc_low}, ~\ref{tab:system_ds_acc},~\ref{tab:system_ds_acc_high}. \ModelName performs the best in most cases, but still the numbers are quite close.
\blue{Under common errors in the real world, CL and NF are also quite close, and filtering does not help.}

% System Low
\tabcolsep=0.08cm
\begin{table}
\center
\caption{Test accuracy of downstream models under systematic noise (Low level) and different filtering methods.}
\label{tab:system_ds_acc_low}
\scriptsize
\begin{tabular}{c|c|cccc|cc}
\hline
\multicolumn{1}{c|}{Dataset} & \multicolumn{1}{c|}{\thead{\scriptsize{DM*}}} & IF & OCSVM & RVAE & \ModelName & CL & NF\\ \hline \hline
\multirow{3}{*}{Wine}            & LR  & 0.7418 & 0.7424 & \textbf{0.7478} & 0.7473 & 0.7457 & 0.7408   \\                  
                                & SVM  & 0.7422 & 0.7453 & \textbf{0.7498} & 0.7492 & 0.7473 & 0.7484  \\ 
                              & NN  & 0.7876 & 0.7890 & 0.7882 & \textbf{0.7976} & 0.8045 & 0.7939  \\ \hline
\multirow{3}{*}{Adult}            & LR & \textbf{0.8224} & 0.8205 & 0.8209 & 0.8189 & 0.8240 & 0.8200  \\ 
                              & SVM  & \textbf{0.8203} & 0.8196 & 0.8165 & 0.8170 & 0.8238 & 0.8186   \\ 
                              & NN  & \textbf{0.7816} & 0.7746 & 0.7748 & 0.7779 & 0.7955 & 0.7815  \\ \hline
\multirow{3}{*}{Restaurant}            & LR  & 0.7336 & 0.7339 & \textbf{0.7359} & 0.7336 & 0.7375 & 0.7356 \\  
                              & SVM & 0.7063 & 0.6863 & 0.7035 & \textbf{0.7082} & 0.7108 & 0.7047 \\ 
                              & NN  & 0.7113 & 0.7072 & 0.7079 & \textbf{0.7160} & 0.7301 & 0.7201 \\ \hline
\multirow{3}{*}{Marketing}            & LR  & 0.7639 & 0.7630 & 0.7616 & \textbf{0.7644} & 0.7672 & 0.7668\\  
                              & SVM  & 0.7658 & 0.7634 & 0.7614 & \textbf{0.7683} & 0.7681 & 0.7676 \\ 
                              & NN & 0.7316 & 0.7305 & \textbf{0.7329} & 0.7312 & 0.7324 & 0.7325\\ \hline
\multirow{3}{*}{Titanic}            & LR  & 0.7866 & 0.7888 & 0.7799 & \textbf{0.7989} & 0.7877 & 0.7821\\  
                              & SVM  & 0.7899 & 0.7866 & 0.7754 & \textbf{0.8022} & 0.7888 & 0.7911 \\ 
                              & NN & 0.7575 & 0.7520 & 0.7564 & \textbf{0.7598} & 0.7944 & 0.8011\\ \hline
\end{tabular}
\begin{tablenotes}
      \item *DM = Downstream Model.
\end{tablenotes}
\end{table}

\tabcolsep=0.08cm
\begin{table}[]
\center
\caption{Test accuracy of downstream models under systematic noise (Medium level) and different filtering methods.}
\label{tab:system_ds_acc}
\scriptsize
\begin{tabular}{c|c|cccc|cc}
\hline
\multicolumn{1}{c|}{Dataset} & \multicolumn{1}{c|}{\thead{\scriptsize{DM*}}} & IF & OCSVM & RVAE & \ModelName & CL & NF\\ \hline \hline
\multirow{3}{*}{Wine}            & LR & 0.7414 & 0.7388 & 0.7435 & \textbf{0.7445} & 0.7457 & 0.7316 \\                  
                                & SVM & 0.7441 & 0.7384 & 0.7459 & \textbf{0.7463} & 0.7461 & 0.7316 \\ 
                              & NN & \textbf{0.7959} & 0.7933 & 0.7918 & 0.7953 & 0.8000 & 0.7855 \\ \hline
\multirow{3}{*}{Adult}            & LR & 0.8136 & 0.8156 & \textbf{0.8207} & 0.8171 & 0.8240 & 0.8098\\ 
                              & SVM  & 0.8103 & 0.8142 & \textbf{0.8178} & 0.8159 & 0.8238 & 0.8080\\ 
                              & NN & 0.7822 & 0.7839 & \textbf{0.7843} & 0.7837 & 0.7931 & 0.7869\\ \hline
\multirow{3}{*}{Restaurant}            & LR  & 0.7305 & 0.7315 & 0.7351 & \textbf{0.7383} & 0.7375 & 0.7372 \\  
                              & SVM & 0.7070 & 0.7008 & \textbf{0.7107} & 0.7077 & 0.7136 & 0.6964\\ 
                              & NN  & 0.7198 & 0.7154 & 0.7175 & \textbf{0.7228} & 0.7346 & 0.7215\\ \hline
\multirow{3}{*}{Marketing}            & LR  & 0.7642 & 0.7640 & 0.7660 & \textbf{0.7673} & 0.7672 & 0.7664 \\  
                              & SVM & 0.7670 & 0.7658 & 0.7655 & \textbf{0.7686} & 0.7681 & 0.7686 \\ 
                              & NN  & 0.7272 & \textbf{0.7311} & 0.7251 & 0.7281 & 0.7277 & 0.7295 \\ \hline
\multirow{3}{*}{Titanic}            & LR & \textbf{0.7877} & 0.7821 & 0.7799 & 0.7866 & 0.7877 & 0.7877 \\  
                              & SVM  & 0.7922 & 0.7777 & 0.7821 & \textbf{0.8022} & 0.7888 & 0.7911 \\ 
                              & NN & 0.7464 & 0.7508 & 0.7464 & \textbf{0.7553} & 0.7866 & 0.7777 \\ \hline
\end{tabular}
\begin{tablenotes}
      \item *DM = Downstream Model.
\end{tablenotes}
\end{table}

% System High
\tabcolsep=0.08cm
\begin{table}
\center
\caption{Test accuracy of downstream models under systematic noise (High level) and different filtering methods.}
\label{tab:system_ds_acc_high}
\scriptsize
\begin{tabular}{c|c|cccc|cc}
\hline
\multicolumn{1}{c|}{Dataset} & \multicolumn{1}{c|}{\thead{\scriptsize{DM*}}} & IF & OCSVM & RVAE & \ModelName & CL & NF\\ \hline \hline
\multirow{3}{*}{Wine}            & LR  & 0.7437 & 0.7359 & 0.7443 & \textbf{0.7447} & 0.7457 & 0.7100   \\                  
                                & SVM  & 0.7457 & 0.7365 & \textbf{0.7476} & 0.7455 & 0.7467 & 0.7041  \\ 
                              & NN  & 0.7961 & 0.7961 & 0.7990 & \textbf{0.8008} & 0.7992 & 0.8031  \\ \hline
\multirow{3}{*}{Adult}            & LR  & 0.8071 & 0.8055 & \textbf{0.8193} & 0.8079 & 0.8240 & 0.8011  \\ 
                              & SVM  & 0.8039 & 0.8038 & \textbf{0.8175} & 0.8060 & 0.8238 & 0.8002   \\ 
                              & NN  & \textbf{0.7843} & 0.7800 & 0.7834 & 0.7822 & 0.7961 & 0.7885   \\ \hline
\multirow{3}{*}{Restaurant}            & LR & 0.7329 & 0.7332 & 0.7346 & \textbf{0.7371} & 0.7375 & 0.7361  \\  
                              & SVM & 0.7155 & 0.7051 & 0.7041 & \textbf{0.7187} & 0.6726 & 0.6925\\ 
                              & NN & 0.7100 & 0.7032 & \textbf{0.7132} & 0.7111 & 0.7232 & 0.7124 \\ \hline
\multirow{3}{*}{Marketing}            & LR & 0.7653 & \textbf{0.7655} & 0.7638 & 0.7636 & 0.7672 & 0.7656 \\  
                              & SVM & 0.7656 & \textbf{0.7661} & 0.7646 & 0.7640 & 0.7681 & 0.7678   \\ 
                              & NN & 0.7292 & \textbf{0.7304} & 0.7256 & 0.7258 & 0.7303 & 0.7294 \\ \hline
\multirow{3}{*}{Titanic}            & LR & 0.7777 & 0.7788 & \textbf{0.7821} & 0.7799 & 0.7877 & 0.7877 \\  
                              & SVM  & 0.7799 & \textbf{0.7855} & \textbf{0.7855} & 0.7799 & 0.7888 & 0.7866 \\ 
                              & NN & 0.7553 & 0.7598 & \textbf{0.7654} & 0.7441 & 0.7855 & 0.7832\\ \hline
\end{tabular}
\begin{tablenotes}
      \item *DM = Downstream Model.
\end{tablenotes}
\end{table}

% Real
\tabcolsep=0.08cm
\begin{table}
\center
\caption{\blue{Test accuracy of downstream models under common errors in the real world and different filtering methods.}}
\label{tab:real_ds_acc}
\scriptsize
\begin{tabular}{c|c|cccc|cc}
\hline
\multicolumn{1}{c|}{Dataset} & \multicolumn{1}{c|}{\thead{\scriptsize{DM*}}} & IF & OCSVM & RVAE & \ModelName & CL & NF\\ \hline \hline
\multirow{3}{*}{Restaurant}            & LR  & \textbf{0.7388} & 0.7351 & 0.7328 & 0.7351 & 0.7404 & 0.7395    \\  
                              & SVM & 0.7028 & 0.6937 & 0.6922 & \textbf{0.7072} & 0.6959 & 0.7112  \\ 
                              & NN & 0.7187 & 0.7176 & \textbf{0.7204} & 0.7137 & 0.7118 & 0.7215   \\ \hline
\multirow{3}{*}{Titanic}            & LR & 0.7464 & 0.7497 & \textbf{0.7732} & 0.7475 & 0.7799 & 0.7609   \\  
                              & SVM  & 0.7363 & 0.7363 & \textbf{0.7609} & 0.7520 & 0.7542 & 0.7598  \\ 
                              & NN  & 0.7274 & 0.7251 & 0.7285 & \textbf{0.7318} & 0.7095 & 0.7207  \\ \hline
\multirow{3}{*}{Food}            & LR &  0.6628 & 0.6960 & 0.6917 & \textbf{0.6978} & 0.7163 & 0.6868   \\  
                              & SVM  & 0.6529 & \textbf{0.6849} & 0.6720 & 0.6794 & 0.7095 & 0.7108   \\ 
                              & NN & 0.6505 & 0.6443 & 0.6431 & \textbf{0.6560} & 0.6609 & 0.6597   \\ \hline
\end{tabular}
\begin{tablenotes}
      \item *DM = Downstream Model.
\end{tablenotes}
\end{table}

\subsection{Test Time Victim Sample Detection under Low/High Level Random/Systematic Noise}
\label{sec:test_time_low_high}
In Table~\ref{tab:random_f1_low},~\ref{tab:random_f1_high},~\ref{tab:system_f1_low},~\ref{tab:system_f1_high}, we show the F1 scores of victim sample detection under low/high level random/systematic noise. The artificial noise setting is the same as described in Section~\ref{sec:inf_time_eval}. We can see that \ModelName outperforms all the other methods in most cases. MWOC performs quite well for the Wine dataset, but it fails completely under high random noise (the F1 score is 0.33). Similar to the case of medium noise, we observe that the reconstruction loss from \NetworkName provides extra signals that improve the detection of victim samples (see the comparison between RF and \ModelName).

%% Test time F1 scores
% Random Low
\tabcolsep=0.05cm
\begin{table}[h]
\center
\caption{$F_1$ scores of victim sample detection at inference time under random noise (Low level).}
\label{tab:random_f1_low}
\scriptsize
\begin{tabular}{c|c|cccccccc}
\hline
\multicolumn{1}{c|}{Dataset} & \multicolumn{1}{c|}{\thead{\scriptsize{DM*}}} & RF & RVAE & RVAE+ & CCS & KNN & TOAO & MWOC & \ModelName \\ \hline \hline
\multirow{3}{*}{Wine}            & LR  & 0.7408 & 0.6910 & 0.7523 & 0.6667 & 0.6626 & 0.4971 & \textbf{0.8084} & 0.7824    \\                  
                                & SVM  & 0.7440 & 0.6918 & 0.7558 & 0.6667 & 0.6638 & 0.6016 & \textbf{0.8004} & 0.7828   \\ 
                              & NN & 0.6882 & 0.6318 & 0.6456 & 0.6770 & 0.6656 & 0.5231 & \textbf{0.7202} & 0.6713  \\ \hline
\multirow{3}{*}{Adult}            & LR & 0.8393 & 0.6563 & 0.8486 & 0.6696 & 0.7834 & 0.1968 & 0.7902 & \textbf{0.8685} \\ 
                              & SVM  & 0.8456 & 0.6743 & 0.8535 & 0.6691 & 0.8131 & 0.4602 & 0.7114 & \textbf{0.8714}  \\ 
                              & NN & 0.8017 & 0.5429 & 0.8052 & 0.6635 & 0.6806 & 0.1900 & 0.7965 & \textbf{0.8267}  \\ \hline
\multirow{3}{*}{Restaurant}            & LR  & 0.7870 & \hspace{0.1cm}--\textsuperscript{\#} & -- & 0.7586 & 0.6702 & 0.6441 & 0.7649 & \textbf{0.8328}   \\  
                              & SVM & 0.6370 & -- & -- & 0.6895 & 0.6351 & 0.6634 & 0.5538 & \textbf{0.7123} \\ 
                              & NN & 0.7609 & -- & -- & 0.7066 & 0.6643 & 0.6071 & 0.7075 & \textbf{0.8119}   \\ \hline
\multirow{3}{*}{Marketing}            & LR & 0.8503 & 0.6340 & 0.8565 & 0.7771 & 0.7913 & 0.6630 & 0.8227 & \textbf{0.8662}\\  
                              & SVM & 0.8590 & 0.6324 & 0.8635 & 0.7789 & 0.8034 & 0.6636 & 0.7748 & \textbf{0.8720}  \\ 
                              & NN & 0.7917 & 0.6197 & 0.7986 & 0.6809 & 0.7134 & 0.6665 & 0.7128 & \textbf{0.8125} \\ \hline
\multirow{3}{*}{Titanic}            & LR  & 0.8281 & -- & -- & 0.7060 & 0.6487 & 0.4377 & 0.7917 & \textbf{0.8451}    \\  
                              & SVM  & 0.8547 & -- & -- & 0.6750 & 0.6544 & 0.6489 & 0.7738 & \textbf{0.8731}      \\ 
                              & NN  & 0.8343 & -- & -- & 0.6678 & 0.6432 & 0.1717 & 0.7798 & \textbf{0.8544}   \\ \hline
\end{tabular}
\begin{tablenotes}
      \item *DM is short for Downstream Model. \textsuperscript{\#}RVAE is not applicable to textual attributes.
\end{tablenotes}
\end{table}

% Random High
\tabcolsep=0.05cm
\begin{table}[h]
\center
\caption{$F_1$ scores of victim sample detection at inference time under random noise (High level).}
\label{tab:random_f1_high}
\scriptsize
\begin{tabular}{c|c|cccccccc}
\hline
\multicolumn{1}{c|}{Dataset} & \multicolumn{1}{c|}{\thead{\scriptsize{DM*}}} & RF & RVAE & RVAE+ & CCS & KNN & TOAO & MWOC & \ModelName \\ \hline \hline
\multirow{3}{*}{Wine}            & LR  & 0.7525 & 0.7867 & 0.7950 & 0.6657 & 0.6727 & 0.5901 & 0.5860 & \textbf{0.8059}    \\                  
                                & SVM & 0.7496 & 0.7898 & 0.7984 & 0.6633 & 0.6815 & 0.7256 & 0.7295 & \textbf{0.8030}   \\ 
                              & NN  & 0.6805 & 0.7697 & \textbf{0.7887} & 0.4560 & 0.6668 & 0.5752 & 0.3301 & 0.7803 \\ \hline
\multirow{3}{*}{Adult}            & LR & 0.7969 & 0.7725 & 0.8149 & 0.6570 & 0.7593 & 0.2408 & 0.5033 & \textbf{0.8273} \\ 
                              & SVM  & 0.8035 & 0.7765 & 0.8201 & 0.6580 & 0.7700 & 0.4737 & 0.4909 & \textbf{0.8312}    \\ 
                              & NN  & 0.7952 & 0.7781 & 0.8124 & 0.3089 & 0.6988 & 0.4284 & 0.4234 & \textbf{0.8214} \\ \hline
\multirow{3}{*}{Restaurant}            & LR  & 0.7457 & \hspace{0.1cm}--\textsuperscript{\#} & -- & 0.7075 & 0.6506 & 0.6504 & 0.7111 & \textbf{0.8137}   \\  
                              & SVM & 0.6948 & -- & -- & 0.6704 & 0.6553 & 0.6567 & 0.5964 & \textbf{0.7824} \\ 
                              & NN & 0.7437 & -- & -- & 0.6788 & 0.6642 & 0.6119 & 0.6852 & \textbf{0.8135}  \\ \hline
\multirow{3}{*}{Marketing}            & LR & 0.8118 & 0.7044 & 0.8146 & 0.7052 & 0.7566 & 0.6645 & 0.7590 & \textbf{0.8244}   \\  
                              & SVM & 0.8111 & 0.7022 & 0.8156 & 0.6994 & 0.7527 & 0.6652 & 0.7486 & \textbf{0.8247}   \\ 
                              & NN & 0.7934 & 0.7068 & 0.7999 & 0.6085 & 0.7042 & 0.6630 & 0.7042 & \textbf{0.8038} \\ \hline
\multirow{3}{*}{Titanic}            & LR  & 0.8134 & -- & -- & 0.6437 & 0.6457 & 0.4383 & 0.7153 & \textbf{0.8227} \\  
                              & SVM  & \textbf{0.8113} & -- & -- & 0.6533 & 0.6354 & 0.6444 & 0.6815 & 0.8105    \\ 
                              & NN  & 0.7993 & -- & -- & 0.6516 & 0.6328 & 0.2824 & 0.6505 & \textbf{0.8058}\\ \hline
\end{tabular}
\begin{tablenotes}
      \item *DM is short for Downstream Model. \textsuperscript{\#}RVAE is not applicable to textual attributes.
\end{tablenotes}
\end{table}

% System Low
\tabcolsep=0.05cm
\begin{table}[h]
\center
\caption{$F_1$ scores of victim sample detection at inference time under Systematic noise (Low level).}
\label{tab:system_f1_low}
\scriptsize
\begin{tabular}{c|c|cccccccc}
\hline
\multicolumn{1}{c|}{Dataset} & \multicolumn{1}{c|}{\thead{\scriptsize{DM*}}} & RF & RVAE & RVAE+ & CCS & KNN & TOAO & MWOC & \ModelName \\ \hline \hline
\multirow{3}{*}{Wine}            & LR  & 0.6826 & 0.5225 & 0.6632 & 0.6667 & 0.6474 & 0.4203 & \textbf{0.8063} & 0.7039    \\                  
                                & SVM  & 0.6658 & 0.5252 & 0.6566 & 0.6667 & 0.6328 & 0.4835 & \textbf{0.7933} & 0.6915   \\ 
                              & NN  & 0.6741 & 0.6010 & 0.5601 & 0.6856 & 0.6661 & 0.4980 & \textbf{0.6985} & 0.6058 \\ \hline
\multirow{3}{*}{Adult}            & LR & 0.8146 & 0.6291 & 0.8176 & 0.6696 & 0.7463 & 0.1842 & 0.7412 & \textbf{0.8317} \\ 
                              & SVM  & 0.8360 & 0.6277 & 0.8418 & 0.6694 & 0.7952 & 0.3382 & 0.6374 & \textbf{0.8589}   \\ 
                              & NN & 0.8100 & 0.5607 & 0.8208 & 0.6026 & 0.6763 & 0.1878 & 0.7740 & \textbf{0.8262}    \\ \hline
\multirow{3}{*}{Restaurant}            & LR  & 0.7951 & \hspace{0.1cm}--\textsuperscript{\#} & -- & 0.7725 & 0.6274 & 0.6460 & 0.7770 & \textbf{0.8269}   \\  
                              & SVM & 0.7080 & -- & -- & 0.6524 & 0.6585 & 0.6488 & 0.5976 & \textbf{0.7321}   \\ 
                              & NN  & 0.7633 & -- & -- & 0.7143 & 0.6588 & 0.6080 & 0.7043 & \textbf{0.7897}   \\ \hline
\multirow{3}{*}{Marketing}            & LR & 0.8540 & 0.6090 & 0.8606 & 0.7855 & 0.7923 & 0.6615 & 0.8274 & \textbf{0.8724}   \\  
                              & SVM  & 0.8597 & 0.6214 & 0.8590 & 0.7939 & 0.7936 & 0.6629 & 0.7828 & \textbf{0.8676} \\ 
                              & NN & 0.7892 & 0.5557 & 0.7899 & 0.6864 & 0.7142 & 0.6658 & 0.6819 & \textbf{0.7972} \\ \hline
\multirow{3}{*}{Titanic}            & LR & 0.8064 & -- & -- & 0.7235 & 0.6409 & 0.3751 & 0.7684 & \textbf{0.8300}  \\  
                              & SVM  & 0.8563 & -- & -- & 0.6778 & 0.6361 & 0.6498 & 0.7867 & \textbf{0.8656}    \\ 
                              & NN & 0.8314 & -- & -- & 0.6679 & 0.6462 & 0.1507 & 0.7667 & \textbf{0.8434} \\ \hline
\end{tabular}
\begin{tablenotes}
      \item *DM is short for Downstream Model. \textsuperscript{\#}RVAE is not applicable to textual attributes.
\end{tablenotes}
\end{table}

% System High
\tabcolsep=0.05cm
\begin{table}[h]
\center
\caption{$F_1$ scores of victim sample detection at inference time under Systematic noise (High level).}
\label{tab:system_f1_high}
\scriptsize
\begin{tabular}{c|c|cccccccc}
\hline
\multicolumn{1}{c|}{Dataset} & \multicolumn{1}{c|}{\thead{\scriptsize{DM*}}} & RF & RVAE & RVAE+ & CCS & KNN & TOAO & MWOC & \ModelName \\ \hline \hline
\multirow{3}{*}{Wine}            & LR  & 0.6866 & 0.3982 & 0.6697 & 0.6667 & 0.6440 & 0.3612 & \textbf{0.7826} & 0.6918    \\                  
                                & SVM  & 0.6784 & 0.4293 & 0.6712 & 0.6667 & 0.6175 & 0.4102 & \textbf{0.7688} & 0.6878  \\ 
                              & NN  & 0.6701 & 0.6127 & 0.5913 & 0.6876 & 0.6656 & 0.5009 & \textbf{0.7536} & 0.5967 \\ \hline
\multirow{3}{*}{Adult}            & LR  & 0.8100 & 0.7619 & 0.8120 & 0.6699 & 0.7234 & 0.1846 & 0.7431 & \textbf{0.8370}\\ 
                              & SVM  & 0.8156 & 0.7507 & 0.8174 & 0.6694 & 0.7463 & 0.3736 & 0.6833 & \textbf{0.8313}  \\ 
                              & NN  & 0.8086 & 0.7341 & 0.8186 & 0.4264 & 0.6883 & 0.2859 & 0.7701 & \textbf{0.8285} \\ \hline
\multirow{3}{*}{Restaurant}            & LR  & 0.7552 & \hspace{0.1cm}--\textsuperscript{\#} & -- & 0.7156 & 0.6475 & 0.6525 & 0.7221 & \textbf{0.8136}   \\  
                              & SVM & 0.7017 & -- & -- & 0.6693 & 0.6626 & 0.6594 & 0.5877 & \textbf{0.7705} \\ 
                              & NN & 0.7523 & -- & -- & 0.6853 & 0.6667 & 0.6123 & 0.7003 & \textbf{0.8149}  \\ \hline
\multirow{3}{*}{Marketing}            & LR  & 0.8232 & 0.6981 & 0.8285 & 0.7423 & 0.7620 & 0.6634 & 0.7864 & \textbf{0.8406}  \\  
                              & SVM  & 0.8361 & 0.6703 & 0.8387 & 0.7138 & 0.7701 & 0.6653 & 0.7433 & \textbf{0.8483} \\ 
                              & NN  & 0.7896 & 0.6960 & 0.7991 & 0.6413 & 0.7066 & 0.6623 & 0.7176 & \textbf{0.8092} \\ \hline
\multirow{3}{*}{Titanic}            & LR & 0.8255 & -- & -- & 0.6843 & 0.6298 & 0.4501 & 0.7830 & \textbf{0.8270}  \\  
                              & SVM  & 0.7945 & -- & -- & 0.6517 & 0.6120 & 0.6686 & 0.6815 & \textbf{0.7972}    \\ 
                              & NN & 0.8240 & -- & -- & 0.6665 & 0.6349 & 0.2243 & 0.7519 & \textbf{0.8347}  \\ \hline
\end{tabular}
\begin{tablenotes}
      \item *DM is short for Downstream Model. \textsuperscript{\#}RVAE is not applicable to textual attributes.
\end{tablenotes}
\end{table}
%\clearpage

\clearpage

\end{document}